%% file: ms.tex
\documentclass[10pt,journal,compsoc]{IEEEtran}

\usepackage{ifpdf}

\ifCLASSOPTIONcompsoc
  \usepackage[nocompress]{cite}
\else
  \usepackage{cite}
\fi

\ifCLASSINFOpdf
   \usepackage[pdftex]{graphicx}

\else

   \usepackage[dvips]{graphicx}
\fi

\usepackage{amsmath,amssymb}

\usepackage{algorithmic}
\newsavebox{\ieeealgbox}

\usepackage{array}

\ifCLASSOPTIONcompsoc
  \usepackage[caption=false,font=footnotesize,labelfont=sf,textfont=sf]{subfig}
\else
  \usepackage[caption=false,font=footnotesize]{subfig}
\fi

\usepackage{fixltx2e}
\usepackage{stfloats}

\ifCLASSOPTIONcaptionsoff
  \usepackage[nomarkers]{endfloat}
 \let\MYoriglatexcaption\caption
 \renewcommand{\caption}[2][\relax]{\MYoriglatexcaption[#2]{#2}}
\fi

\newcommand{\removelatexerror}{\let \@latex@error \@gobble}

\usepackage{url}

\usepackage{amsfonts,amsthm,amsmath}
\usepackage{multirow}

\newcommand{\normm}[1]{\ensuremath{\|{#1}\|}}
\newcommand{\comment}[1]{}
\newcommand{\shortcomment}[1]{}
\newcommand{\later}[1]{$\ast\ast\ast$}

\input{tool/math_commands}

\input{tool/my_math_commands}

\begin{document}
%\title{Nonlinear Projection using Denoising Autoencoders}

\title{GradDiv: Adversarial Robustness of Randomized Neural Networks via Gradient Diversity Regularization}

\author{Sungyoon~Lee,
        Hoki~Kim,
        and~Jaewook~Lee% <-this % stops a space
\IEEEcompsocitemizethanks{\IEEEcompsocthanksitem S. Lee is with the Department of Mathematical Sciences, Seoul National University,
599 Gwanak-ro, Gwanak-gu, Seoul 151-744, South Korea. \protect
\IEEEcompsocthanksitem H. Kim and J. Lee are with the Department
of Industrial Engineering, Seoul National University,
599 Gwanak-ro, Gwanak-gu, Seoul 151-744, South Korea.}
% goman1934@snu.ac.kr, ghrl9613@snu.ac.kr, 

\thanks{Please address all correspondences to Dr. Jaewook Lee, Department of Industrial Engineering, Seoul National University, 1 Gwanak-ro, Gwanak-gu, Seoul 151-744, South Korea. {\it
E-mail address:} jaewook@snu.ac.kr.}}

\IEEEtitleabstractindextext{%
\begin{abstract}
Deep learning is vulnerable to adversarial examples. Many defenses based on randomized neural networks have been proposed to solve the problem, but fail to achieve robustness against attacks using proxy gradients such as the Expectation over Transformation (EOT) attack. We investigate the effect of the adversarial attacks using proxy gradients on randomized neural networks and demonstrate that it highly relies on the directional distribution of the loss gradients of the randomized neural network. We show in particular that proxy gradients are less effective when the gradients are more scattered. To this end, we propose Gradient Diversity (GradDiv) regularizations that minimize the concentration of the gradients to build a robust randomized neural network. Our experiments on MNIST, CIFAR10, and STL10 show that our proposed GradDiv regularizations improve the adversarial robustness of randomized neural networks against a variety of state-of-the-art attack methods. Moreover, our method efficiently reduces the transferability among sample models of randomized neural networks.
 %in the white/black-box setting.
\end{abstract}

% \begin{IEEEkeywords}
% Adversarial machine learning, Directional Analysis, Determinantal Point Process
% \end{IEEEkeywords}
}
%Bayesian deep learning, Image classification,

\maketitle

\IEEEdisplaynontitleabstractindextext

% \fi
\IEEEpeerreviewmaketitle

\input{text_revised}

%%%%%%%%%%%%%%%%%%%%%%%%%%%%%%%%%%%%%%%%%%%%%%%%%%%%%%%%%%%%%%%%%%%%%%%%%%%%%%%%%%%%%%%%%%%%%%%%%%%%%%%%%%%%%%%%%%%%%%%%%%%%%%%%%%

\end{document}

%% file: tool/math_commands.tex
\usepackage{amsmath,amsfonts,bm}

\def\eqref#1{equation~\ref{#1}}
\def\1{\bm{1}}

\def\0{{\bm{0}}}
\def\1{{\bm{1}}}
\def\vepsilon{\bm{\epsilon}}

\def\vtheta{{\bm\theta}}

\def\vmu{\bm{\mu}}

\def\vu{{\bm{u}}}

\DeclareMathAlphabet{\mathsfit}{\encodingdefault}{\sfdefault}{m}{sl}
\SetMathAlphabet{\mathsfit}{bold}{\encodingdefault}{\sfdefault}{bx}{n}

\def\sB{{\mathbb{B}}}

\DeclareMathOperator*{\argmax}{arg\,max}

%% file: tool/my_math_commands.tex
\usepackage{color,soul}
\usepackage{enumitem}
\usepackage{nicefrac}
\usepackage{booktabs}

\renewcommand{\eqref}[1]{(\ref{#1})}

\newtheorem{definition}{Definition}

\newtheorem{theorem}{Theorem}
\def\0{\bm{0}}

\newcommand{\etal}{{et al.}}

%% file: text_revised.tex
\IEEEraisesectionheading{\section{Introduction}\label{sec:introduction}}
\IEEEPARstart{D}{eep} learning has achieved successful performance in many applications which deal with natural images. However, it has been shown that a small adversarially-designed perturbation to a natural image can fool deep neural networks\cite{szegedy2013intriguing}. %biggio2013evasion
Such perturbed images are called adversarial examples.
Subsequent research has been conducted on generating adversarial examples 
% using iterative optimization-based methods
and defending against these adversarial attacks\cite{goodfellow2014explaining,madry2017towards,zhang2019theoretically,carlini2017towards,gowal2020uncovering}. %moosavi2016deepfool
%guo2017countering,xie2017mitigating,dhillon2018stochastic,yang2019me,liu2018towards,liu2018adv,he2019parametric,
%kurakin2016adversarial_in_physical_world,moosavi2017universal,tramer2017ensemble,,yang2019me

One attempt to defend against adversarial attacks is to construct a randomized neural network which uses different networks for each inference\cite{guo2017countering,xie2017mitigating,dhillon2018stochastic}.
By doing so, even in the white-box setting, where the adversary has full access to the network and the defense strategy implemented, it is infeasible to utilize the gradient of the inference model to construct adversarial examples since the adversary cannot specify the model used in the inference phase.

However, the attack algorithm Expectation over Transformation (EOT) successfully circumvents the previous randomized defenses by using the expected gradient over the randomization as an alternative to the true gradient of the inference model\cite{athalye2017synthesizing,athalye2018obfuscated}.
EOT approximates the expected gradient by the sample mean using Monte Carlo estimation with sample models from the randomized network.
The EOT attack on a randomized network can be considered as a kind of substitute-based transfer attack in the sense that the adversary has no direct access to the inference model to be randomly sampled and utilizes the averaged classifier of the randomized network as a surrogate model.
Therefore, the success of EOT can be attributed to the high transferability among sample models of randomized networks.

\begin{figure}[t]
\begin{center}%[width=0.95\linewidth]
  \includegraphics[width=\linewidth]{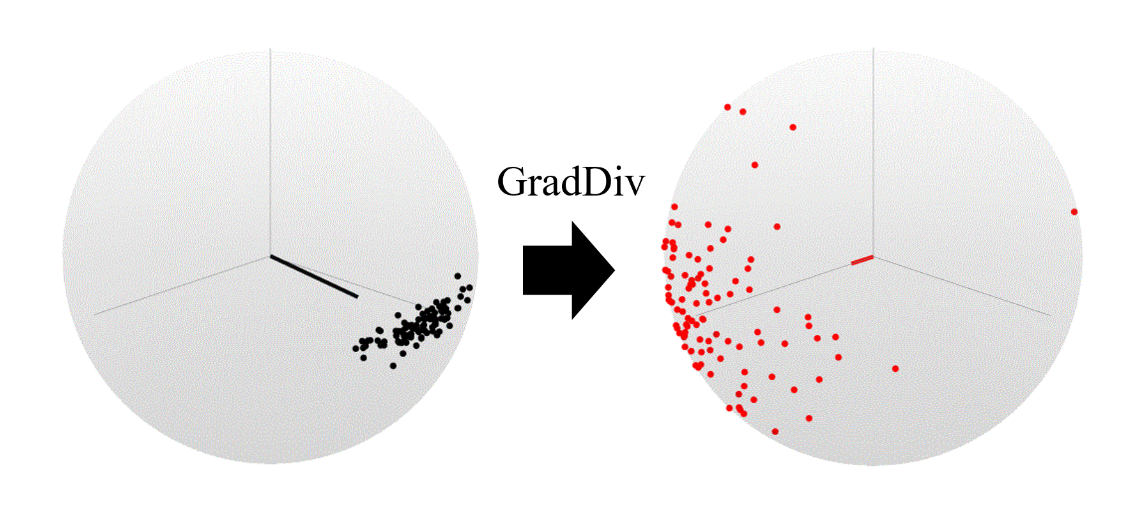}
\end{center}
   \caption{\textbf{Illustration of the proposed Gradient Diversity (GradDiv) regularization.} GradDiv encourages a randomized neural network to have dispersed loss gradients.
   The $xy$-plane is spanned by the sample mean vectors $\vu$ (black line) and $\vu_{reg}$ (red line) of the sample gradients for Baseline (black dots) and Baseline+GradDiv (red dots), respectively.
%   The $xy$-plane is spanned by the sample mean vectors $\vu$ (black line) and $\vu_{reg}$ (red line) of the sample gradients for Adv-BNN (black dots)\cite{liu2018adv} and Adv-BNN+GradDiv (red dots), respectively.
   The length of sample mean vector is called the Mean Resultant Length (MRL).
   The $z$-axis is chosen to be orthogonal to both $\vu$ and $\vu_{reg}$.}
\label{fig:intro}
\end{figure}

There has been a line of work on understanding transferability and measuring the effectiveness of transfer attacks\cite{papernot2017practical,papernot2016transferability,tramer2017space,liu2016delving}. %tramer2017ensemble,
One explanation of the transferability is that it comes from the similarity between the gradients of the target and surrogate model\cite{papernot2016transferability}.
In particular, sample models of a randomized network tend to have highly aligned gradients as shown in Figure \ref{fig:intro} (left, black).
% based on Adv-BNN\cite{liu2018adv} 

In this paper, we aim to answer the following question: \textit{"How can we build a robust randomized neural network against adversarial attacks?"} %, especially against the EOT attack % in the white-box setting
To answer this question, we first investigate the effect of adversarial attacks using proxy gradients on the randomized neural networks and find that the expected gradient used in the EOT attack is locally the optimal direction to maximize the expected loss increase.
Next, we demonstrate that the expected loss increase caused by proxy gradients is upper bounded by the Mean Resultant Length (MRL) of the loss gradients. 
Intuitively, in the extreme case in which the gradient directions are uniformly distributed (zero MRL), any proxy gradient (for example, the sample mean used in the EOT attack) is meaningless in constructing adversarial examples.
Therefore, we propose regularizations called \textit{GradDiv} to disperse the gradient distribution as in Figure \ref{fig:intro} (right, red).
In this way, we could build a robust randomized network against adversarial attacks.

% First, we consider the cosine similarities of the gradient samples and propose regularizers which penalize these similarities.
% Second, we use the estimated concentration parameter as a regularizer. 
% Lastly, we use the Determinantal Point Process (DPP) framework to enhance the \textit{diversity} of the sample models in the sense that they are less likely to transfer to each other.
% To the best of our knowledge, our work is the first to study the limitations of the EOT attack and utilize its properties to build a robust randomized network.

To summarize, the main contributions of this paper are as follows:
\begin{itemize}
    \item We analyze the relationship between a randomized neural network's gradient distribution and its robustness against adversarial attacks using proxy gradients and show that the adversarial attacks are less effective when the gradients are more scattered.

    \item We propose regularization methods collectively called GradDiv which encourage a randomized network's gradient dispersal by penalizing concentration.

    \item We test GradDiv on MNIST, STL10, and CIFAR10, and empirically demonstrate that GradDiv successfully leads to gradient dispersion and improves the robustness of the randomized network. % based on Adv-BNN\cite{liu2018adv}. 
    Moreover, GradDiv makes sample models of the randomized network significantly less transferable to each other.
\end{itemize}

\section{Literature Review}
\noindent\textbf{Stochastic defenses.}\quad
% \subsection{Randomized defenses}
Many methods\cite{guo2017countering,xie2017mitigating,dhillon2018stochastic,liu2018towards,liu2018adv,he2019parametric,pang2019mixup,eustratiadis2021weightcovariance} have been proposed to defend against adversarial attacks by introducing randomness into the classifier.
However, they have been broken by a stronger adaptive proxy-gradient-based attack \cite{athalye2018obfuscated,croce2020reliable,tramer2020adaptive}.

%using random transformations to the inputs before feeding them into the classifier.
One line of work introduced random transformations to the inputs before feeding them into the classifier.
Guo \etal\cite{guo2017countering} used input transformations such as bit-depth reduction, JPEG compression, TV minimization, and image quilting, and Xie \etal\cite{xie2017mitigating} used the transformation that resizes the input to a random size and randomly pads the resized input with zeros.
% Dhillon \etal\cite{dhillon2018stochastic} prunes a random subset of activations and compensates this by scaling up the remaining activations. 
However, Athalye \etal\cite{athalye2018obfuscated} have shown that these defenses\cite{guo2017countering,xie2017mitigating} rely on obfuscated gradients and can be completely broken by the EOT attack.

Another line of work proposed to introduce randomness on the layerwise input/activation or the weight.
Dhillon \etal\cite{dhillon2018stochastic} proposed to prune a random subset of activations and compensate this by scaling up the remaining activations, but it has also been broken by the EOT attack. 
The Random Self-Ensemble (RSE) approach introduced random noise layers to make the network stochastic and ensembled the prediction over the randomness to achieve stable performance\cite{liu2018towards}. 
While RSE introduced randomness by perturbing the inputs of each layer, another approach called Adv-BNN\cite{liu2018adv} used Bayesian neural network structure embedded with adversarial training\cite{madry2017towards} to make the weight of the model stochastic.
Moreover, He \etal \cite{he2019parametric} proposed a method called Parametric Noise Injection (PNI) injecting learnable noise on the layerwise weight or inputs.
However, these approaches \cite{liu2018towards,liu2018adv,he2019parametric} still evaluated their robustness against a weak attack rather than the EOT attack.
Especially, it has been shown (for the STL10 dataset) that Adv-BNN may not improve robustness against the EOT attack compared to the standard adversarial training\cite{rol2019comment}.
These randomized defense methods failed because they
% aim to improve performance by introducing randomness into the classifier, but they 
did not consider interactions among sample models which are important features in the robustness of randomized neural networks, and thus we utilize them to design regularizers.

% PNI \cite{he2019parametric} - n/ot evaluated under the EOT attack

Recently, Mixup Inference (MI)\cite{pang2019mixup} 
has been proposed, but it is broken by a stronger adaptive attack with the EOT principle\cite{tramer2020adaptive}. % Adaptive attack에서 EOT랑 비슷한 방법으로 (사실상 그냥 EOT) broken
Moreover, Weight-Covariance Alignment network (WCA-Net) \cite{eustratiadis2021weightcovariance} achieved a strong robustness against the PGD attack, but it seems to cause gradient obfuscation as the black-box attack (the Square attack \cite{andriushchenko2020square}) performs better than white-box attack (PGD) against WCA-Net (see their Table 4).\\
% L2P\cite{jeddi2020learn2perturb}\\ 

% transferable adversarial attack\cite{liu2016delving}

% substitute attack\cite{papernot2017practical}

% AA\cite{croce2020reliable}
% adaptive attacks\cite{tramer2020adaptive}

\noindent\textbf{Transferability.}\quad
% \subsection{Transferability}
It has been shown that an attacker can fool a target model without knowledge of the target model using a substitute model, which is called a transfer attack\cite{papernot2017practical}. There has been a subsequent line of work on transferability from a source model to a target model\cite{papernot2016transferability,tramer2017space,liu2016delving}.
Papernot \etal\cite{papernot2016transferability} attribute the transferability to the similarity between the loss gradients of the target and source models, Tramer \etal\cite{tramer2017space} to the high dimensionality of adversarial subspace, and Liu \etal\cite{liu2016delving} to the alignment of decision boundaries of the target and the source models.\\

\noindent\textbf{Ensemble-based defenses.}\quad 
% \subsection{Ensemble-based defenses} 
There is another line of work on the (deterministic) ensemble-based defenses.
Although not our focus here, they have a similar goal/methodology with ours.
Kariyappa and Qureshi\cite{kariyappa2019improving} proposed an ensemble of models with misaligned loss gradients to improve the robustness against transfer attack in the black-box setting, whereas we focus on the randomized neural networks in the white-box setting.
Also, Pang \etal\cite{pang2019improving} proposed a regularization called ensemble diversity which uses the Determinantal Point Process (DPP) framework to enhance the robustness of an ensemble model. 
The main difference is that Pang \etal\cite{pang2019improving} use the DPP framework on non-maximal predictions, while we use it on the loss gradients.
DVERGE\cite{yang2020dverge} diversifies and distills the non-robust features of sub-models, and thus the ensemble achieves higher robustness against transfer attack, but not against the white-box attack.

% TRS\cite{yang2021trs} % 우리 논문과 비슷한걸 하려나

\section{Background}
\subsection{Notations}
% \noindent\textbf{Notations}\quad
A randomized classifier can be represented as $F\sim q_{\vtheta}(F)$ with a parametric distribution $q_{\vtheta}(F)$, and thus the randomized classifier $F$ can be learned via optimizing the parameter $\vtheta$. A sample model from the randomized classifier $F$ is denoted as $f:\mathcal{X}\subset\mathbb{R}^p\rightarrow\mathcal{Y}$, where the input space is denoted as $\mathcal{X}$ and the output space as $\mathcal{Y}=\mathbb{R}^c$ with the number of classes $c$. The cross-entropy loss function is represented as $\mathcal{L}\left(f(X),Y\right)$, where $(X,Y)$ is a pair of the input $X$ and the corresponding label $Y$.
% To avoid confusion,
We alternatively use the short notation $l_f(X)=\mathcal{L}\left(f(X),Y\right)$. % The predicted class output of a classifier $f$ for the input $X$ is denoted as $\hat{y}(X)=\argmax_i f(X)_i\in \{0,1, \cdots, c-1\}$.

% \subsection{Adversarial Attack against Deterministic Classifier}
\subsection{Gradient-based Adversarial Attacks}
% 기본적인 문제화 TODO: 양봐서 결정
Within a bound $\mathbb{B}(X_0,\epsilon)$ with the radius of the perturbation bound $\epsilon$ and the center of an original image $X_0$, finding an adversarial example $X_0+\vepsilon$ is often formulated as a maximization problem of the loss function as follows:
\begin{equation}
\label{main_opt}
 \max_{X_0+\vepsilon\in\mathbb{B}(X_0, \epsilon)} \mathcal{L}(f(X_0+\vepsilon),Y)
\end{equation}

\noindent\textbf{Fast Gradient (Sign) Method (FG(S)M).}\quad
The Fast Gradient Sign Method (FGSM) is an efficient one-step gradient-based adversarial attack\cite{goodfellow2014explaining}. Starting from the original image $X_0$, it can be formulated as follows:
\begin{equation}
    X\leftarrow X_0+\epsilon\cdot\text{sign}({\nabla_X\mathcal{L}(f(X_0),Y)})
    \label{eqn:FGS_basic}
\end{equation}
If the adversary uses the $l_2$ normalization function instead of the sign function in (\ref{eqn:FGS_basic}), it is called the Fast Gradient Method (FGM).
Both methods can be integrated into a single formulation:
\begin{equation}
\label{eqn:FGS_proj}
    X\leftarrow X_0+\epsilon\cdot\Pi_{\partial\mathbb{B}(0,1)}{\nabla_X\mathcal{L}(f(X_0),Y)}
\end{equation}
where $\Pi_S$ is the projection onto the set $S$, and $\partial\mathbb{B}$ represents the boundary of $l_\infty$-ball or $l_2$-ball for the FGSM and the FGM, respectively. The perturbation in (\ref{eqn:FGS_proj}) provides the largest inner product with the gradient $\nabla_{X}\mathcal{L}(f(X_0),Y)$ within the ball $\mathbb{B}(X_0,\epsilon)$ around the image $X_0$. Therefore, (\ref{eqn:FGS_proj}) can be written as the following equivalent update:
\begin{equation}
    X\leftarrow X_0+\argmax_{g\in\mathbb{B}(0,\epsilon)} \left(g^T {\nabla_X\mathcal{L}(f(X_0),Y)}\right)
    \label{eqn:FGS_argmax}
\end{equation}

\noindent\textbf{Projected Gradient Descent (PGD).}\quad Projected Gradient Descent (PGD) is one of the strongest adversarial attacks\cite{madry2017towards}.
While FG(S)M approximates the loss function as a linear function in the whole bound $\mathbb{B}(X_0,\epsilon)$, PGD updates the perturbed input $X_k$ iteratively by approximating the loss function as a linear function in a smaller local region around the current $k$-th point $X_k$. 
Starting from the original image $X_0$ $(k=0)$, it performs one FG(S)M attack with step size $\alpha(\leq\epsilon)$ and the projection onto the $l_\infty$-ball or $l_2$-ball around the original image $X_0$ for each iteration as follows:
\begin{align}
\label{eqn:PGD_step}
X_k'&\leftarrow X_k+\Pi_{\partial\mathbb{B}(0,\alpha)}{\nabla_X\mathcal{L}(f(X_k),Y)},\\
X_{k+1}&\leftarrow \Pi_{\mathbb{B}(X_0,\epsilon)} X'_k
\end{align}
As in (\ref{eqn:FGS_argmax}), the projection in (\ref{eqn:PGD_step}) can be replaced with the following update:
\begin{equation}
\label{eqn:PGD_argmax}
X'_k\leftarrow X_k+\argmax_{g \in \mathbb{B}(0,\alpha)}\left({g^T \nabla_X\mathcal{L}(f(X_k),Y)}\right)
\end{equation}
\subsection{Proxy-gradient-based Adversarial Attacks}%Adversarial Attack against Randomized Classifier}
% \subsection{Expectation over Transformation (EOT)}
% \noindent\textbf{Expectation over Transformation (EOT)}\quad
For a randomized network, simple gradient-based methods like the PGD attack are infeasible to obtain the true gradient of the inference model because of its test-time randomness.\\
% Therefore, using an estimate of the true gradient.\\
% an attack algorithm called Expectation over Transformation (EOT)\cite{athalye2017synthesizing,athalye2018obfuscated} has been proposed which uses the expected gradient over the randomness as an estimate of the true gradient.

\noindent\textbf{Expectation over Transformation (EOT).}\quad
EOT
% is a modified version of the PGD attack
uses the expected gradient over the randomness as a proxy gradient.
In particular, for each step of the PGD attack, the update can be written as:
\begin{equation}
    X'_k\leftarrow X_k+\argmax_{g \in \mathbb{B}(0,\alpha)}\left(g^T\hat{\mathbb{E}}_{F\sim q_{\vtheta}}\left[\nabla_X\mathcal{L}(F(X_k),Y)\right]\right)
\end{equation} in place of (\ref{eqn:PGD_step}), and we call this attack EOT-PGD.
It approximates the expected gradient by a sample mean of sample gradient vectors from the randomized neural network.
Note that it is not limited to the input transformation case first presented in \cite{athalye2017synthesizing}.
The EOT principle can also be applied to other adversarial attacks such as APGD\cite{croce2020reliable}, B\&B\cite{brendel2019accurate}, and the Square attack\cite{andriushchenko2020square}.

% Note that standard gradient-based attacks are special instances of the EOT attack when $q_\vtheta(F)=\delta(f)$ and $\delta$ is the Dirac delta function.
% $F\sim\delta(f)$ and $\delta$ is the Dirac delta function.

\section{Gradient diversity regularization}
In this section, we first investigate the relationship between the gradient distribution and the robustness against adversarial attacks using proxy gradients, and then based on this analysis, we propose new regularization methods collectively called \textit{GradDiv} to mitigate the effect of the adversarial attacks on the randomized networks.

\subsection{vMF Distribution and the Concentration Parameter}
\label{sec:vMF}
We model directional data of loss gradients of a randomized neural network with a $p$-dimensional von Mises-Fisher (vMF) distribution\cite{mardia2009directional} which arises naturally in directional data where $p=\text{dim}(\mathcal{X})$.
In particular, the vMF distribution is obtained from the normal distribution constraining on the unit hypersphere $S^{p-1}$.
In this way, we could utilize estimated parameters of the vMF distribution from the gradient data to construct regularizers.

\begin{definition}
The pdf of the vMF distribution is defined as
\begin{equation}
    pdf(v;\mu,\kappa)=C_p(\kappa)\exp(\kappa \mu^T v)
\end{equation}
where $v\in S^{p-1}$ is a variable vector on the $(p-1)$-dimensional hypersphere $S^{p-1}$, $\kappa\geq 0$ is the concentration parameter, $\mu\in S^{p-1}$ is the mean direction, $C_p(\kappa)=\frac{\kappa^{p/2-1}}{(2\pi)^{p/2}I_{p/2-1}(\kappa)}$ is the normalization constant and $I_r$ is the $r$-th modified Bessel function.
\end{definition}

The (population) mean resultant length (MRL) $\rho$ is defined as $\rho=\Vert \mathbb{E}[v] \Vert_2\leq1$ and it satisfies the equation $\rho=A_p(\kappa)$ with the concentration parameter $\kappa$ where $A_p(\cdot)=\frac{I_{p/2}(\cdot)}{I_{p/2-1}(\cdot)}$  is a monotonically increasing function.
In addition, the sample MRL $\hat{\rho}$ is defined as $\hat{\rho}=\Vert \bar{v} \Vert_2$ where the sample mean $\bar{v}=\frac{\sum_{i=1}^n v_i}{n}$ of the samples $v_i$. 
The concentration parameter $\kappa$ is used as a measure of the concentration of the directional distribution.
To estimate the parameter $\kappa$ based on a maximum likelihood estimation, the approximation $\hat{\kappa}\approx {\hat{\rho}(p-\hat{\rho})}/({1-\hat{\rho}^2})$
% \begin{equation}
%     \hat{\kappa}\approx {\hat{\rho}(p-\hat{\rho})}/({1-\hat{\rho}^2})
%     \label{eqn:kappa_hat}
% \end{equation}
is often used since the exact estimation is intractable\cite{banerjee2005clustering}. 
Similarly, we also define the $\ell_q$-MRL $\rho_q\equiv\|\mathbb{E}[v]\|_q$ and the sample $\ell_q$-MRL $\hat{\rho}_q\equiv\|\overline{v}\|_q$.
Throughout the paper, we use $n$ to denote the number of directional data samples, especially the gradient samples.

\begin{figure}[t]
\begin{center}
   \includegraphics[width=.8\linewidth]{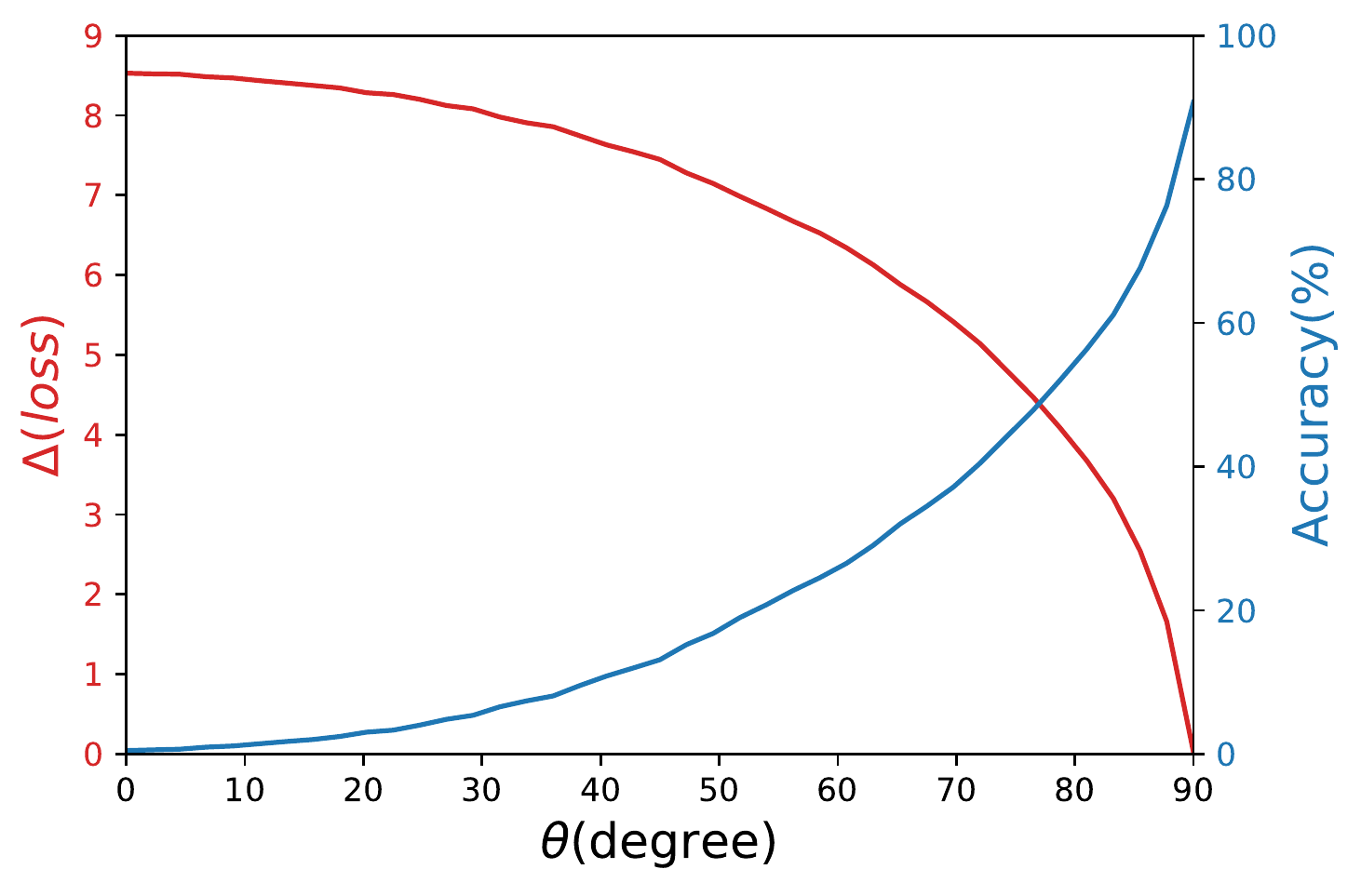}
\end{center}
   \caption{\textbf{The effectiveness of the PGD attack using proxy gradients instead of the actual gradients.} The proxy gradients are obtained by rotating the true gradients by $\theta$ with a random axis at every iterations of the attack.
   The loss difference caused by the proxy gradients (red) decreases, and the accuracy under the PGD attack (blue) increases as $\theta$ increases to $90^\circ$.
   Note that the graph of the loss difference (red) appears similar to the cosine curve, empirically demonstrating (\ref{eqn:delta}).
   }
\label{fig:cosgraph}
\end{figure}

\subsection{Relationship between the Gradient Distribution and the Proxy-gradient-based Attacks}
When the gradient of the target inference model $f$ is not accessible, it is crucial to make a proper estimate of the actual gradient to construct adversarial examples. % via optimization-based attacks.
% In other words,
For example, it requires solving the following problem (maximizing the inner product) in (\ref{eqn:PGD_argmax}) in each iteration:
\begin{equation}
  \argmax_{g \in \mathbb{B}(0,\alpha)}\left(g^T\nabla_X\mathcal{L}(f(X),Y)\right)
\end{equation}
without direct access to the true gradient $\nabla_X\mathcal{L}(f(X),Y)$.
Figure \ref{fig:cosgraph} shows that it is crucial for the proxy gradient to maximize the inner product (small $\theta$) to increase the loss and decrease the accuracy.
% 에서 볼 수 있듯이 inner product를 가장 크게하면 하는 proxy를 쓰면 쓸수록 attack 효과가 크다. 그러나
However, it needs not to be optimal to fool the network and
% As shown in Figure \ref{fig:cosgraph}, 
a proxy gradient with a sufficiently large inner product with the true gradient is sufficient to cause misclassification.

The above discussion can be extended to our case, the white-box attack against a randomized network.
In this case, it is infeasible to directly utilize the gradient of the inference model since the adversary cannot specify the model used in the inference phase.
Therefore, we further investigate the effect of a proxy gradient $g$ to the randomized network $F$ in terms of the expected loss increase along the direction $g$. 
Using the first-order approximation, we can prove that the increase is proportional to the inner product value with the expectation $\hat{g}=\mathbb{E}_{q_{\vtheta}}[\nabla_X l_{F}(X)]$ of the gradients as follows:
\begin{equation}
    \Delta(\alpha g)\equiv\mathbb{E}_{q_{\vtheta}}\left[ l_{F}(X+\alpha{g})-l_{F}(X)\right]=\alpha \hat{g}^T {g}+O(\alpha^2)
\label{eqn:delta}
\end{equation}
This indicates that the expected gradient $\hat{g}$ used in the EOT attack, i.e., $\alpha \hat{g}/\|\hat{g}\|_2$ and $\alpha \text{sign}(\hat{g})$ for $\ell_2$- and $\ell_\infty$-norm bounded perturbations, respectively, is locally the optimal direction to maximize the expected loss increase $\Delta(\alpha {g})$.
% For example, we can represent $\Delta(\alpha g)\leq\Delta(\alpha \hat{g}/\|\hat{g}\|_2)+O(\alpha^2)$ for $\alpha g\in\sB_2(0,\alpha)$.
% In particular, $\Delta(\alpha {g}/\|g\|_2)$ and $\Delta(\alpha \text{sign}{g})$ for $\ell_2$- and $\ell_\infty$-norm bounded perturbations.
Moreover, \eqref{eqn:delta} explains why the graph of the loss difference (red) in Figure \ref{fig:cosgraph} looks similar to the cosine curve.
% due to $\hat{g}^Tg=\|\hat{g}\|_2\|g\|_2\cos(\hat{g},g)$ where $\cos(u,v)$ is the cosine of the angle between two directions $u$ and $v$.
% If we plug the normalized expected gradient $\hat{g}/\|\hat{g}\|_2$ into $g$ in \eqref{eqn:delta}, then 
From \eqref{eqn:delta}, we can upper bound the expected loss increase as in the following theorem:

\begin{theorem}
\label{thm}
For a randomized neural network $F$ with $\ell_2$-bounded gradients at a given point $X$, the expected loss increase at $X$ along $g$
% the direction of the expected gradient $\hat{g}$ 
with the step size $\alpha$ is upper bounded as follows:
    \begin{align}
        % \Delta(\alpha\hat{g}/\normm{\hat{g}}_2)
        \Delta(\alpha g)&\leq\alpha M_X \rho_q +O(\alpha^2),~\forall\alpha g \in \mathbb{B}_p(0,\alpha)\label{eqn:thm}
        % \Delta(\alpha g)&\leq\alpha M_X \rho_1 +O(\alpha^2),~\forall\alpha g \in \mathbb{B}_\infty(0,\alpha)\label{eqn:cor}
    \end{align}
% for $\alpha  g \in \mathbb{B}_2(0,\alpha)$
for some $M_X>0$, where $\nicefrac{1}{p}+\nicefrac{1}{q}=1$ and $\rho_q$ is the $\ell_q$-MRL of the gradient direction $v={\nabla_X l_F(X)}/{\normm{\nabla_X l_F(X)}_2}$. %\frac{\nabla_X l_F(X)}{\normm{\nabla_X l_F(X)}_2}$.
\end{theorem}
\begin{proof}

From the boundedness of the gradient of the loss function, i.e., $\|\nabla_X l_{F}(X)\|_2\leq M_X$, we can derive
\begin{equation}
\normm{\hat{g}}_q=\normm{\mathbb{E}_{q_{\vtheta}}[\nabla_X l_{F}(X)]}_q\leq\normm{M_X\mathbb{E}_{q_{\vtheta}}[\frac{\nabla_X l_{F}(X)}{\normm{\nabla_X l_{F}(X)}_2}]}_q,
\end{equation}
Therefore, for $\alpha g \in \mathbb{B}_p(0,\alpha)$,
% if we choose to attack with the direction $\hat{g}$,
% /then 
the following inequality for the expected loss increase is obtained from (\ref{eqn:delta}):
\begin{align}
    \Delta(\alpha g)&\leq\max_{\alpha g\in\sB_p(0,\alpha)} \hat{g}^T(\alpha  g)+O(\alpha^2)\\
    % \label{eqn:thm_proof_last_eqn}\\
    &=\alpha \normm{\hat{g}}_q+O(\alpha^2)\leq\alpha M_X \rho_q +O(\alpha^2)
\end{align}
\end{proof}
% In the proof, by replacing $\hat{g}/\|\hat{g}\|_2$ with $\text{sign}(\hat{g})$ in \eqref{eqn:thm_proof_last_eqn}, we can easily derive the following corollary for the $\ell_\infty$-case:
% \begin{corollary}
% \label{cor}
% Under the same assumption in Theorem \ref{thm},
%     \begin{equation}
%         \Delta(\alpha g)\leq\alpha M_X \rho_1 +O(\alpha^2),~\forall\alpha g \in \mathbb{B}_\infty(0,\alpha)
%     \label{eqn:cor}
%     \end{equation}
% % for $\alpha g\in\sB_\infty(0,\alpha)$ 
% where $\rho_1$ is the $\ell_1$-MRL of the gradient direction. % $\rho_1=\|\mathbb{E}[v]\|_1$.
% \end{corollary}
% 여기에 최악의 경우의 논의를 진행, random network이므로 mean vector가 가장 '효율적'인 attack
Therefore, we penalize MRL $\rho$ of the gradient directions to lower the upper bound in (\ref{eqn:thm}), and thus build a robust randomized neural network against proxy-gradient-based attacks.
To this end, we propose the following regularizers:
\begin{align}
    R_{\kappa}(X;\vtheta)&=\frac{1}{p}\hat{\kappa}=\frac{1}{p}\frac{\hat{\rho}(p-\hat{\rho})}{1-\hat{\rho}^2},\label{eqn:reg_kappa}\\
%    L_{cos}(\theta)&=\hat{\mathbb{E}}\left[\cos(g_i, g_j)\right]\\
    R_\text{mean}(X;\vtheta)&=\frac{1}{n(n-1)}\sum_{i\neq j}\cos(g_i, g_j),\label{eqn:reg_cos}
\end{align} where $\{g_i\}_{i=1}^n$ is a set of sample gradients at the point $X$ and $\hat{\rho}$ is the sample MRL of the sample gradients.
% , and $\cos(u,v)$ is the cosine of the angle between two directions $u$ and $v$.
Note that the regularizer $R_\text{mean}$ in (\ref{eqn:reg_cos}) is the average value of the cosine between gradient samples which directly penalizes the left-hand side of (\ref{eqn:delta}).
We also tested variants of the regularizer (\ref{eqn:reg_cos}) (see the supplementary material for the details).

\subsection{Reducing Transferability using DPP}
A Determinantal Point Process (DPP) is a stochastic point process with a probability distribution over subsets of a given ground set $\mathcal{G}$ and the sampling of each element in the subsets being negatively correlated\cite{kulesza2012determinantal}. 
Therefore, it assigns higher probabilities to subsets which are diverse, and thus the probability relies on a measure used to evaluate the diversity among items in a subset.
DPPs are often applied to select a subset configuration with high diversity, but we focus on learning the parameter $\vtheta$ of the distribution $q_\vtheta$ of the randomized neural network $F$ that would give diverse sample models.

We consider the population $\Omega$ of the randomized neural network $F$ as a ground set $\mathcal{G}=\Omega$, and subsets $S_{n}=\{f_1,\cdots,f_n\}$ of sample models as point configurations of the ground set $\mathcal{G}$.
We suppose that the sampling follows $F\sim q_{\vtheta}(F)$, and aim to learn the parameter $\vtheta$ to make the point configurations diversified, i.e., to be less transferable to each other. 
In this setting, we focus on a specific class of DPPs called L-ensembles\cite{borodin2005eynard}.
An L-ensemble models the probabilities for subset $S=\{s_i\}$ as
\begin{equation}
    \mathcal{P}_L(S)=\frac{\det(L_S)}{\sum_{T\subset \mathcal{G}}\det(L_T)}=\frac{\det(L_S)}{\det(L+I)}\propto \det(L_S)
\end{equation} with a positive semi-definite matrix $L$, where $L_S\in\mathcal{R}^{|S|\times |S|}$ is the restriction of $L$ to the entries indexed by elements of $S$, where $(L_S)_{i,j}=k(s_i,s_j)$ and $k(\cdot,\cdot)$ is a kernel function that measures the similarity between two inputs.
As mentioned, the diversity/similarity measure is crucial to determine the characteristics of a DPP.
For our purposes, we define it by the inner product between the corresponding gradients, i.e., $k(f_i,f_j)=\langle g_i, g_j \rangle$, where $g_i =\nabla_{X}\mathcal{L}(f_i(X),Y)/\normm{\nabla_{X}\mathcal{L}(f_i(X),Y)}_2$.
Therefore, we model the probability for the sample models $S_{n}$ as $\mathcal{P}_L(S_{n})\propto \det(G_n^TG_n)$ where $G_n = [g_1, \cdots, g_n]$ is a $p\times n$ matrix.

Therefore, to maximize the likelihood, we use the negative log-likelihood as a regularizer:
\begin{align}
\label{eqn:reg_DPP}
    % R_\text{DPP}(X;\vtheta)&=-\log \mathcal{P}_L(S_{n}|X;\vtheta)\nonumber\\
    % &=-\log\det(G_n^TG_n)
    R_\text{DPP}(X;\vtheta)=-\log \mathcal{P}_L(S_{n}) %|X;\vtheta
    =-\log\det(G_n^TG_n)
\end{align}
Intuitively, $\mathcal{P}_L(S_{n})\propto \det(G_n^TG_n)=Vol^2(G_n)$, where $Vol(G_n)$ is the volume of the n-dimensional parallelotope spanned by the columns of $G_n$, i.e., $\{g_i\}_{i=1}^n$.
% Note that it requires some computation when $n$ is large compared to $R_{\kappa}$ and $R_\text{mean}$.

Finally, we consider the total objective of the randomized neural network with the parameter $\vtheta$ formulated as:
\begin{equation}
    L(X;\vtheta) + \lambda R_\text{GradDiv}(X;\vtheta),
    \label{eqn:final}
\end{equation}
where $L(X;\vtheta)$ is the original objective,
% for the randomized neural network,
$R_\text{GradDiv}$ is one of the following GradDiv regularizers $\{R_{\kappa},R_\text{mean},R_\text{DPP} \}$, and $\lambda$ is the regularization parameter.

\section{Experiments}
\noindent\textbf{Datasets and Setup.}\quad To evaluate our methods, we perform experiments on the MNIST\cite{lecun1998gradient}, CIFAR10\cite{krizhevsky2009learning}, and STL10\cite{coates2011analysis} datasets.
During the training, we generate adversarial examples on the fly using PGD with $\epsilon_\text{train}=0.3,\nicefrac{8}{255},0.03$ for MNIST, CIFAR10, and STL10, respectively, the step size $\alpha_\text{train}=\epsilon_\text{train}/4$ and the attack iterations $m=10$.

We refer to the GradDiv regularizations in (\ref{eqn:reg_kappa}), (\ref{eqn:reg_cos}), and (\ref{eqn:reg_DPP}) as GradDiv-$\kappa$, GradDiv-mean, and GradDiv-DPP, respectively. 
To compute the GradDiv regularization terms, we have to sample the loss gradients. 
We empirically found that three gradient samples ($n=3$) for each input are sufficient. % to improve performance.
To stabilize the training with the objective (\ref{eqn:final}), we start with the regularization parameter $\lambda =0$ during a warm-up period and linearly ramp it up to a target value during a ramp-up period. 
To cross-validate the target value, we extracted the validation set from the training set.
% hyperparameters of the regularization methods.
We used the last 5,000 training examples as the validation set for the MNIST datasets, and subsampled the last 10\% of the training examples for the validation set for the CIFAR10 and STL10 datasets.
We cross-validated over the range of the regularization weight $\lambda\in[0.05,10]$ with the validation set to select the best model for each dataset. 
The best regularizers and the corresponding regularization weights for each dataset are as follows: MNIST (GradDiv-$\kappa$, 1), CIFAR10 (GradDiv-DPP, 1), and STL10 (GradDiv-mean, 0.7).
We refer the readers to Table \ref{supp:params} in the supplementary material for more details about the training parameters.\\

\begin{figure*}[t]
\begin{center}
   \includegraphics[width=0.32\linewidth]{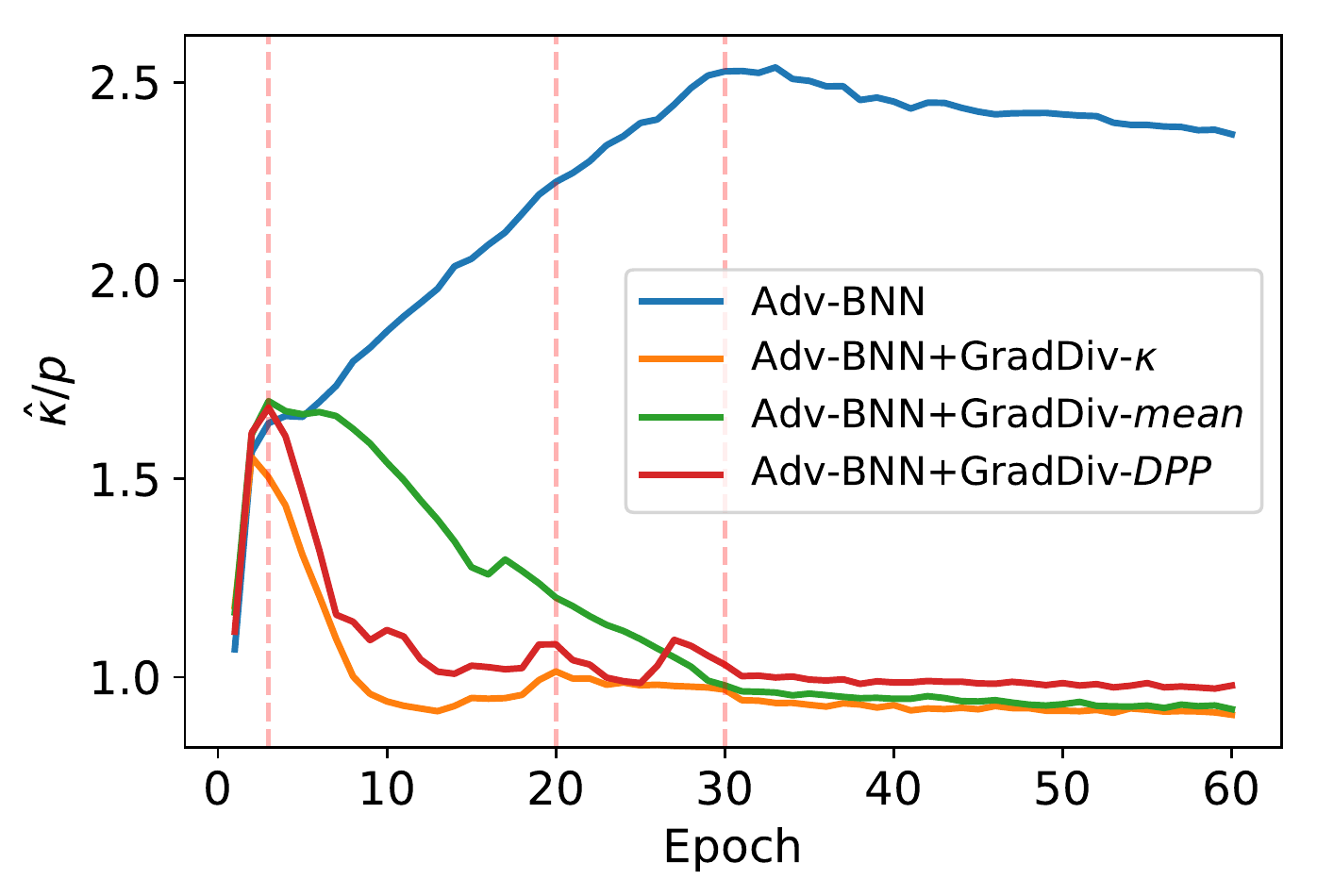}
   \includegraphics[width=0.32\linewidth]{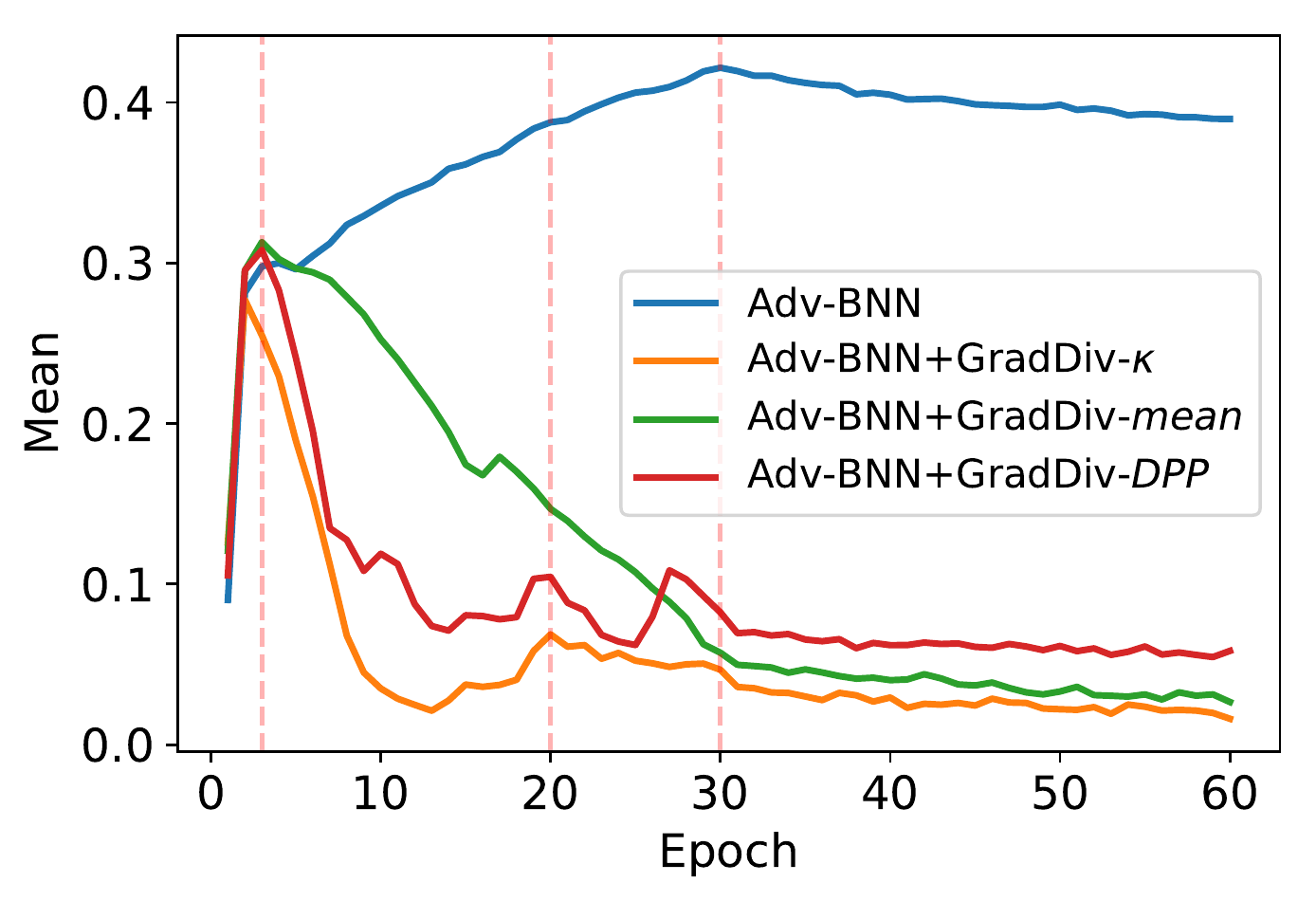}
   \includegraphics[width=0.32\linewidth]{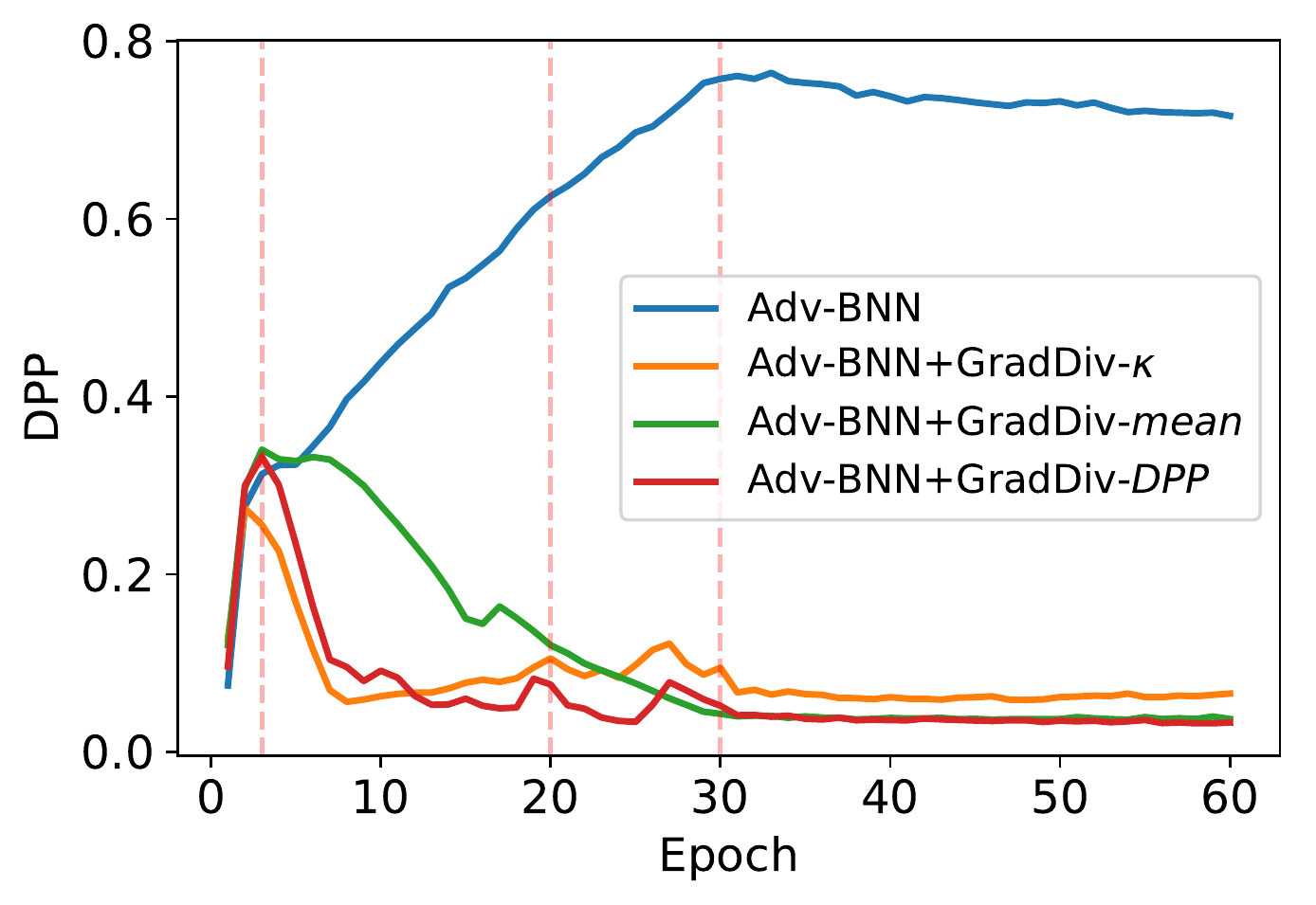}
\end{center}
    \caption{\textbf{The change in the concentration measures (\ref{eqn:reg_kappa}) (left), (\ref{eqn:reg_cos}) (middle), and (\ref{eqn:reg_DPP}) (right) during training on MNIST.} The two leftmost vertical lines in each graph indicate the end of the warm-up and ramp-up periods, and the third vertical line indicates when the learning rate has decayed.} %TODO: 혹시 다끝나고 순서확인
\label{fig:reg}
\end{figure*}

\noindent\textbf{Baseline.}\quad We use Adv-BNN\cite{liu2018adv} as a baseline randomized network, and we test the effect of GradDiv on the baseline.
In addition, we set the hyperparameters in the Adv-BNN objective as the KL regularization factor $\alpha_{KL}=0.02$ and the prior standard deviation $\sigma_0=0.05,0.1,0.15$ for MNIST, CIFAR10, and STL10, respectively. 
The hyperparameters were chosen to achieve similar performance on the Adv-BNN model as in\cite{liu2018adv}.
Note that GradDiv can be applied to any other randomized defenses, but we mainly focus on applying it to Adv-BNN because by doing so it can directly optimize the parameter on the randomness, while other methods such as\cite{xie2017mitigating,guo2017countering,liu2018towards} introduce randomness using manual configuration which can not be learned via stochastic gradient descent.
See the supplementary material for the results using RSE\cite{liu2018towards} as a baseline. \\

\begin{figure}[t]
\begin{center}
   \includegraphics[width=.6\linewidth]{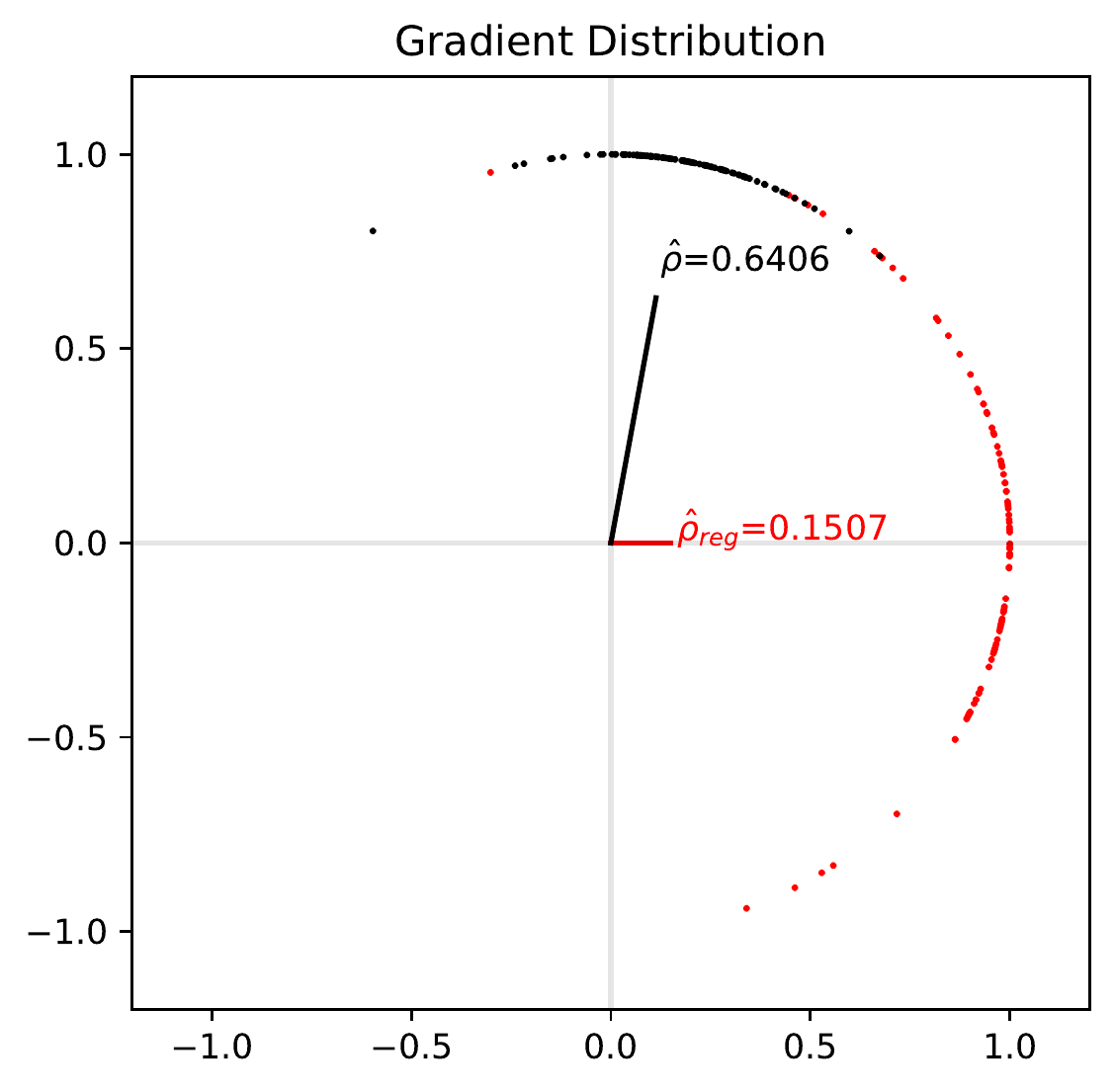}
   \includegraphics[width=.9\linewidth]{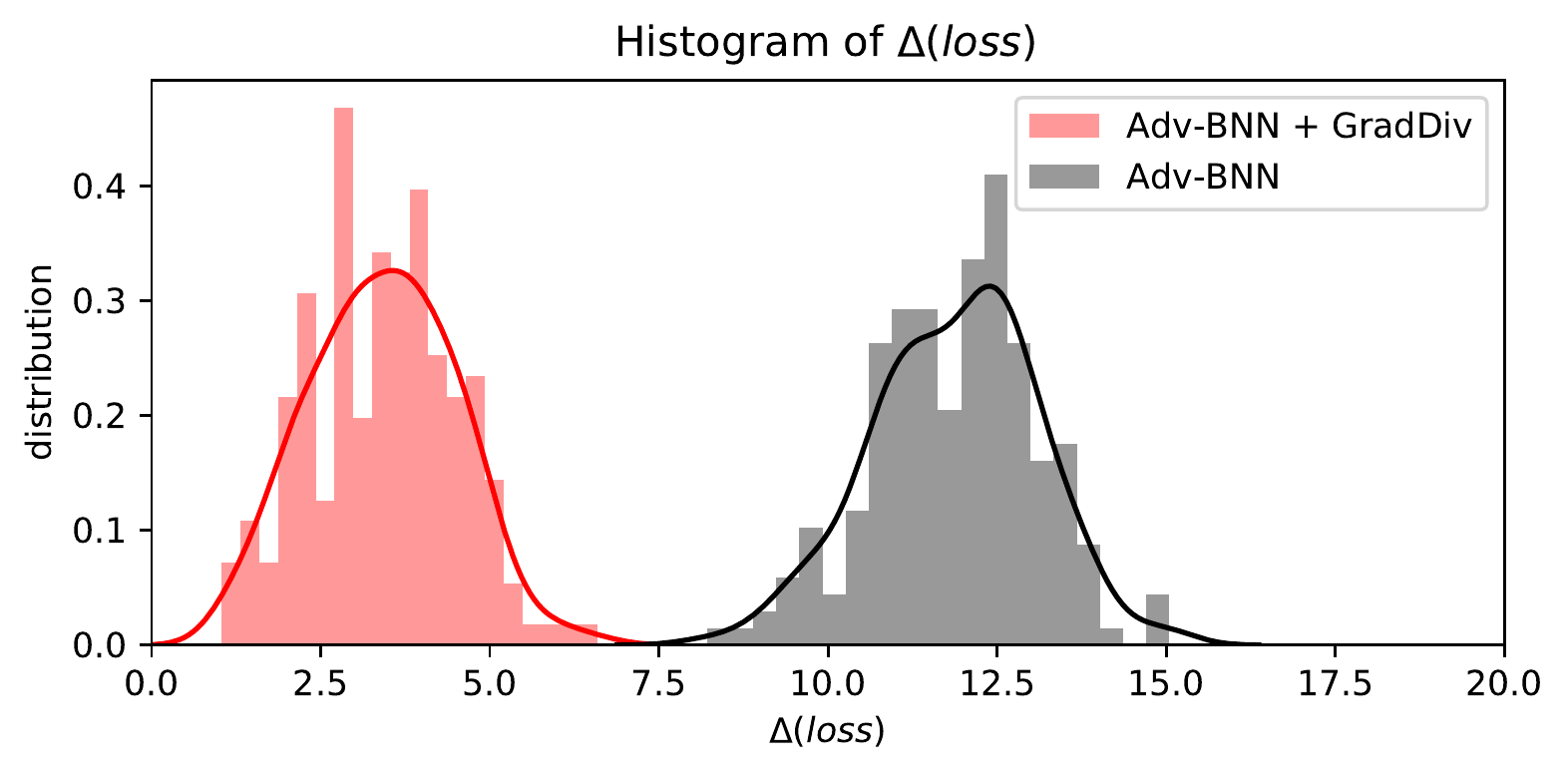}
\end{center}
   \caption{\textbf{Top: The scatter plot of the gradient samples on the unit circle.} The dots indicate the gradient samples from the randomized neural network and the lines indicate the sample mean vectors of the gradients with the length of the sample MRL $\hat{\rho}$ (black: Adv-BNN, red: Adv-BNN+GradDiv).
   \textbf{Bottom: The density plots of the loss increase under the EOT-PGD attack.}
   GradDiv effectively reduces the effects of the proxy-gradient-based attack as indicated in \eqref{eqn:thm}.
   }
\label{fig:2d}
\end{figure}

% \begin{figure}[t]
% \begin{center}
% %   \includegraphics[width=.9\linewidth]{0fig/MNIST_2d.pdf}
%   %\includegraphics[width=.95\linewidth]{0fig/Delta_MNIST_63.pdf}
%   \includegraphics[width=\linewidth]{0fig/Delta_MNIST_final.pdf}
% \end{center}
%   \caption{\textbf{The density plots of the loss increase under the EOT attack.}
%   GradDiv effectively reduces the effects of the attack as indicated in \eqref{eqn:thm}.
%   See Figure \ref{fig:2d} together.
%   }
% \label{fig:Delta}
% \end{figure}

\noindent\textbf{Effects of GradDiv during Training.}\quad Figure \ref{fig:reg} shows the effects of GradDiv on the concentration measures we used, i.e., the normalized estimate $\hat{\kappa}/p$ of the concentration parameter (\ref{eqn:reg_kappa}), the mean of the cosine similarity (\ref{eqn:reg_cos}), and the DPP loss (\ref{eqn:reg_DPP}) during training on MNIST.
The values are averaged over each epoch. The concentration measures are successfully reduced by introducing GradDiv after the warm-up period, while the measures tend to increase without GradDiv, especially for the first 30 epochs with large learning rates.
Even though each regularizer has different objectives, they have similar regularization effects.
We also observed similar results for CIFAR10 and STL10 (see the supplementary material for the details).
Moreover, the concentration measures for the networks trained with GradDiv converge close to the optimal values. 
For example, the DPP loss in Figure \ref{fig:reg} (Right) shows that the DPP loss reaches near to 0, which is optimal, when the n-dimensional parallelotope has the largest volume of 1, i.e., when every gradient sample is orthogonal to each other.
\\

\begin{table*}[]
\caption{\textbf{The test accuracy (mean$\pm$std\% for stochastic defenses) against a set of adversarial attacks with $\epsilon=\nicefrac{8}{255}$ on the CIFAR10 dataset.}  % under the white-box/black-box (WB/BB) settings
Bold numbers are the highest accuracy.
The last row (Total) shows the robust accuracy against the combined attacks.
% Each column indicates the results according to the different perturbation bounds $\epsilon$.
% See the supplementary material for the detailed settings.
% We use the EOT attack with the gradient sample sizes $n=10$.
% Note that the performances of deterministic neural networks (No defense and Adv.train) do not depend on the gradient sample size $n$.
\label{table:acc_cifar10}
}
\def\arraystretch{1.} %1.5
\resizebox{\textwidth}{!}{
\begin{tabular}{l|cc|cccc|c}
\toprule
& \textbf{No defense} & \textbf{Adv.train} & \textbf{RSE} & \textbf{PNI} & \textbf{MI} & \begin{tabular}[c]{@{}c@{}}\textbf{Baseline}\\\textbf{(=Adv-BNN)}\end{tabular} &  \begin{tabular}[c]{@{}c@{}}\textbf{Baseline}\\\textbf{+GradDiv}\end{tabular} \\ \hline
% \multicolumn{2}{c|}{\textbf{None}}
\textbf{None}
&\textbf{92.39}&78.83&84.27$\pm$0.16&81.19$\pm$0.07&84.08$\pm$0.10&75.78$\pm$0.45&77.51$\pm$1.30 \\ \hline
% \multirow{15}{*}{\textbf{WB}} 
\textbf{FGSM}                   
& 8.11&48.61&45.07$\pm$0.12&51.18$\pm$0.14&47.35$\pm$0.11&63.14$\pm$0.26&\textbf{71.30$\pm$1.59}\\
\textbf{PGD}                    
& 0.00&37.38&30.75$\pm$0.14&44.09$\pm$0.44&35.41$\pm$0.32&66.32$\pm$0.68&\textbf{75.72$\pm$1.62} \\
\textbf{APGD$_\text{CE}$}       
& 0.00&36.79&42.21$\pm$0.17&46.54$\pm$0.46&36.80$\pm$0.41&69.09$\pm$0.19&\textbf{74.06$\pm$1.49} \\
\textbf{APGD$_\text{DLR}$}       
& 0.04&38.69&58.95$\pm$0.20&56.07$\pm$0.65&43.68$\pm$0.46&73.09$\pm$0.42&\textbf{74.93$\pm$1.17} \\      
\textbf{B\&B}                   
& 0.43&41.88&\textbf{81.54$\pm$0.10}&76.04$\pm$0.23&76.17$\pm$0.32&75.09$\pm$0.51&75.38$\pm$0.62\\    
\textbf{FAB}                   
& 0.01&36.94&\textbf{83.37$\pm$0.13}&77.24$\pm$0.12&77.04$\pm$0.21&74.94$\pm$0.37&77.04$\pm$1.24\\
\textbf{EOT$^1$-FGSM}                   
& .&.&41.84$\pm$0.16&51.61$\pm$0.22&46.89$\pm$0.24&63.02$\pm$0.55&\textbf{70.34$\pm$1.84} \\
\textbf{EOT$^1$-PGD}                    
& .&.&22.74$\pm$0.13&37.58$\pm$0.18&31.51$\pm$0.16&50.31$\pm$0.40&\textbf{59.24$\pm$1.69}\\
\textbf{EOT$^1$-APGD$_\text{CE}$}       
& .&.&35.00$\pm$0.20&42.97$\pm$0.11&32.46$\pm$0.18&67.27$\pm$0.40&\textbf{75.26$\pm$1.43}\\
\textbf{EOT$^1$-APGD$_\text{DLR}$}      
& .&.&40.24$\pm$0.30&46.24$\pm$0.15&35.20$\pm$0.42&70.20$\pm$0.30&\textbf{76.27$\pm$1.40}\\
\textbf{EOT$^1$-B\&B}                   
& .&.&51.35$\pm$0.19&66.32$\pm$0.87&65.49$\pm$0.70&\textbf{75.49$\pm$0.61}&\textbf{75.92$\pm$0.61}\\
\textbf{EOT$^1$-FAB}                   
& .&.&\textbf{83.12$\pm$0.20}&78.78$\pm$0.23&81.95$\pm$0.33&75.66$\pm$0.45&77.49$\pm$1.30\\
\textbf{EOT-FGSM}                
& .&.&51.45$\pm$0.21&46.32$\pm$0.14&45.39$\pm$0.09&51.17$\pm$0.45&\textbf{63.26$\pm$1.95}\\
\textbf{EOT-PGD}                
& .&.&39.66$\pm$0.28&35.56$\pm$0.21&31.09$\pm$0.11&43.39$\pm$0.48&\textbf{45.88$\pm$1.68} \\
\textbf{EOT-APGD$_\text{CE}$}   
& .&.&28.40$\pm$0.07&38.89$\pm$0.26&32.06$\pm$0.21&59.74$\pm$0.32&\textbf{68.91$\pm$1.56}\\
\textbf{EOT-APGD$_\text{DLR}$} 
& .&.&31.56$\pm$0.10&40.98$\pm$0.23&34.06$\pm$0.26&66.09$\pm$0.21&\textbf{73.91$\pm$1.41} \\ \hline
% & \textbf{EOT-B\&B}                
% & -     & -     & 00.00$\pm$0.00 & 00.00$\pm$0.00 & 00.00$\pm$00.00 & 00.00$\pm$0.00 & 00.00$\pm$0.00 \\ \hline
% \multirow{3}{*}{\textbf{BB}} 
% & \textbf{Square}                 
% & 5.73 & 49.54 & \textbf{83.75$\pm$0.00} & 73.87$\pm$0.00 & 72.17$\pm$0.10 & 75.32$\pm$0.00 & 75.88$\pm$0.00 \\
% & \textbf{EOT$^1$-Square}                 
% & -     & -     & \textbf{82.73$\pm$1.84} & 80.51$\pm$0.00 & \textbf{81.46$\pm$0.20} & 76.16$\pm$0.07 & 76.79$\pm$0.25 \\
% & \textbf{EOT-Square}             
% & -     & -     & \textbf{81.21$\pm$0.01} & 77.31$\pm$0.00 & 73.60$\pm$0.03 & 72.85$\pm$0.00 & 70.04$\pm$0.00 \\
\textbf{Total}    & 0.00&35.44&19.34$\pm$0.25&34.82$\pm$0.23&30.38$\pm$0.08&42.95$\pm$0.45&\textbf{45.77$\pm$1.63} \\
% \textbf{Total}    & 0.00     & 35.61     & 18.49$\pm$0.13 & 34.63$\pm$0.10 & 29.18$\pm$0.03 & 38.66$\pm$0.16 & \textbf{39.17$\pm$0.41} \\
                                    \bottomrule
\end{tabular}
}
\end{table*}

\noindent\textbf{Gradient Distribution and Loss Increase.}\quad We first demonstrate the effects on the directional distribution of the gradients when the regularizations are applied on MNIST as in Figure \ref{fig:2d} (Top). 
To visualize the high dimensional gradient samples, we project them onto a 2D plane.
The projection plane is spanned by two vectors $\bar{g}$ (black line) and $\bar{g}_{reg}$ (red line), where $\bar{g}$ is the sample mean vector of the gradient samples for the model trained without GradDiv and $\bar{g}_{reg}$ is the counterpart for the model trained with GradDiv.
We sample 100 gradients at the first test image. We also noted the sample MRLs $\hat{\rho}$ and $\hat{\rho}_{reg}$ near the mean vectors $\bar{g}$ and $\bar{g}_{reg}$, respectively. 
The mean vectors are on the projection plane, so their lengths can be compared directly. 
The gradients appear to be more dispersed when the regularizations are applied (red dots in Figure \ref{fig:2d} (Top)).
The dispersion is evaluated by the sample MRL, and it is shortened about 75\% from $\hat{\rho}=0.64$ to $\hat{\rho}_{reg}=0.15$ after applying GradDiv.

In Theorem \ref{thm}, we argued that the loss increase under the proxy-gradient-based attack is upper bounded by the MRL $\rho$ of the gradients, and this is empirically proven in Figure \ref{fig:2d} (Bottom).
In detail, GradDiv successfully reduces the loss increase by EOT-PGD and the loss increase is about 75\% smaller after applying GradDiv, similar to the ratio of the sample MRLs, $\hat{\rho}_{reg}/\hat{\rho}$.
In the experiment, we use EOT-PGD with the number of sample gradients $n=20$, $\epsilon=0.4$, $\alpha=0.1$, and the attack iterations $m=20$ on MNIST.\\

\begin{figure}[t]
    \begin{center}
       \includegraphics[width=.8\linewidth]{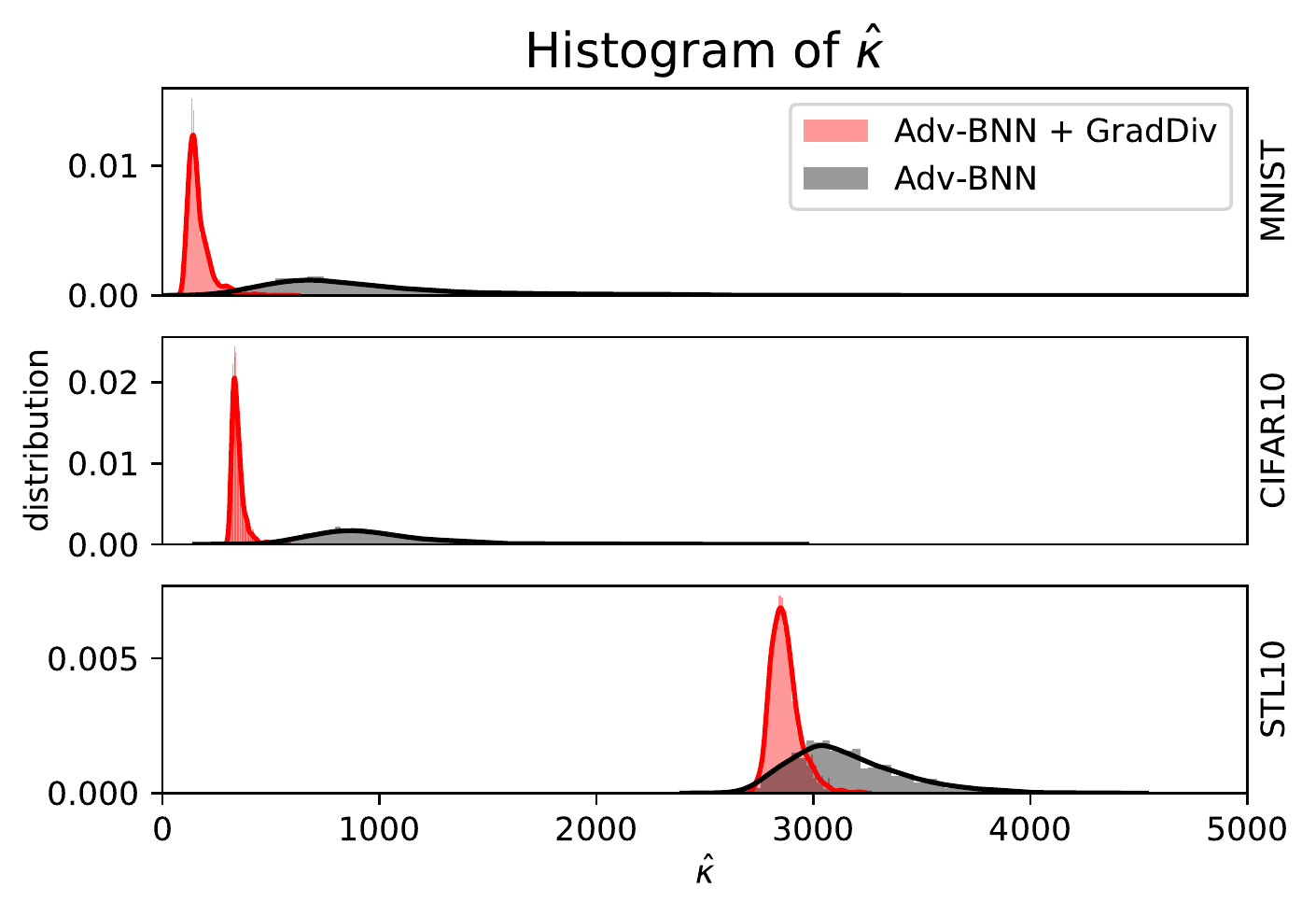}
    \end{center}
   \caption{\textbf{The density plots of the estimated concentration parameters $\hat{\kappa}$ for the test examples of MNIST (top), CIFAR10 (middle), and STL10 (bottom).}
   GradDiv effectively reduces the concentration parameter $\hat{\kappa}$.
   }
\label{fig:kappa}
\end{figure}

% \begin{figure}[t]
%     \begin{center}
%         \includegraphics[width=.4\linewidth]{0fig/2d_last.pdf}
%       \includegraphics[width=.55\linewidth]{0fig/kappa_hist_final3.pdf}
%     \end{center}
%   \caption{\textbf{The density plots of the estimated concentration parameters $\hat{\kappa}$ for the test examples of MNIST (Top), CIFAR10 (Middle), and STL10 (Bottom).}
%   GradDiv effectively reduces the concentration parameter $\hat{\kappa}$.
%   }
% \label{fig:kappa}
% \end{figure}

\noindent\textbf{The Estimated Concentration Parameter $\hat{\kappa}$.}\quad In Figure \ref{fig:kappa}, we draw density plots of the estimated concentration parameters $\hat{\kappa}$ of the sample gradients for every test image.
While Figure \ref{fig:2d} (Top) shows the distribution of gradients at a single test image, the density plots can represent whole test images.
We sample 100 gradients for the estimation of the concentration parameter $\kappa$. 
As desired, the density plots show that the regularized models have lower estimated concentration parameters $\hat{\kappa}$ compared to the baseline, Adv-BNN. 
Note that, compared to the other datasets, the estimated concentration parameters $\hat{\kappa}$ for the STL10 dataset have higher values because the STL10 images are embedded in a higher-dimensional ($3\times96\times96$) space.
\\% than the other datasets. \\

\begin{figure}[t]
\begin{center}%[width=0.95\linewidth]
   \includegraphics[width=0.95\linewidth]{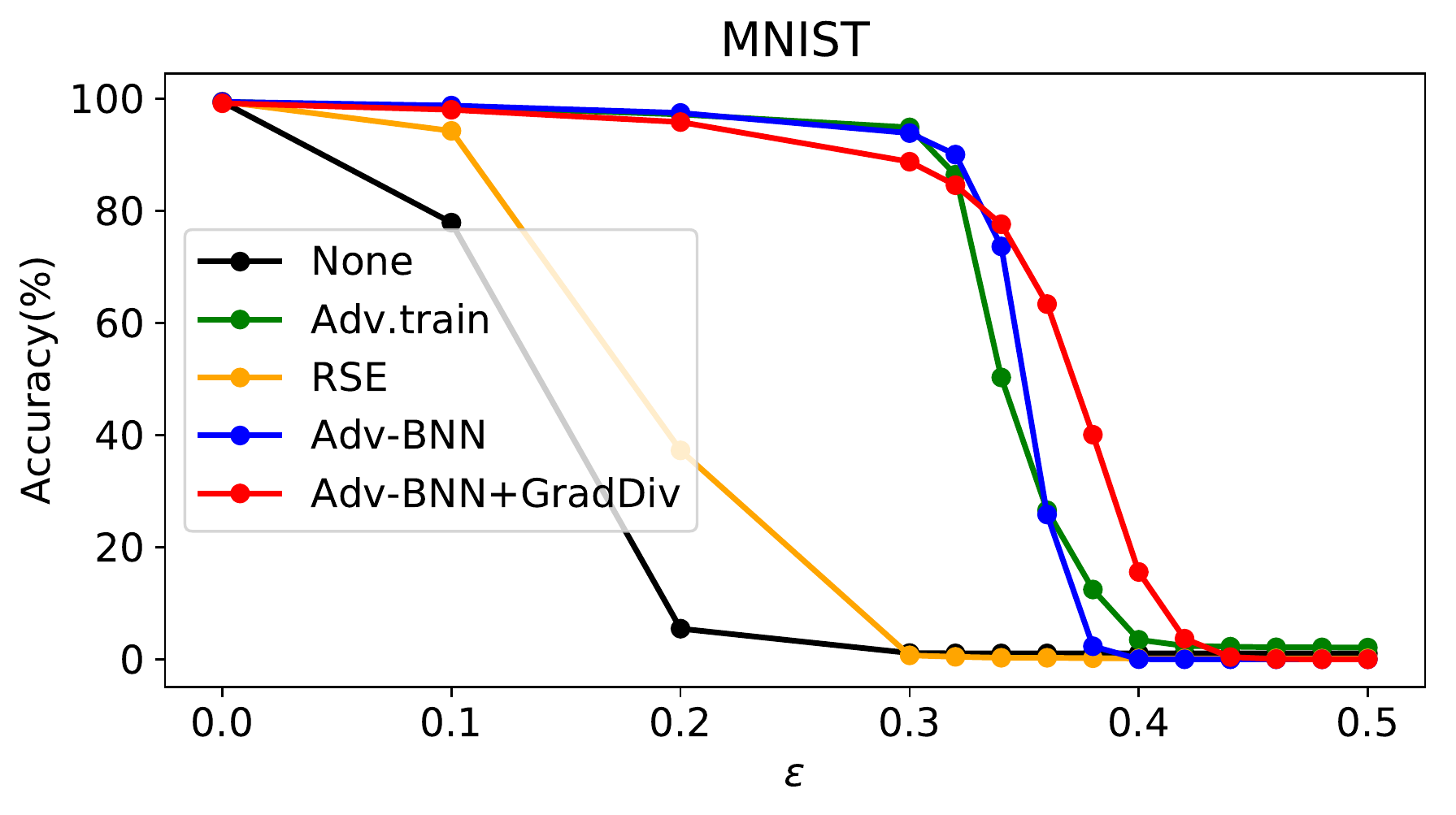}
   \includegraphics[width=0.95\linewidth]{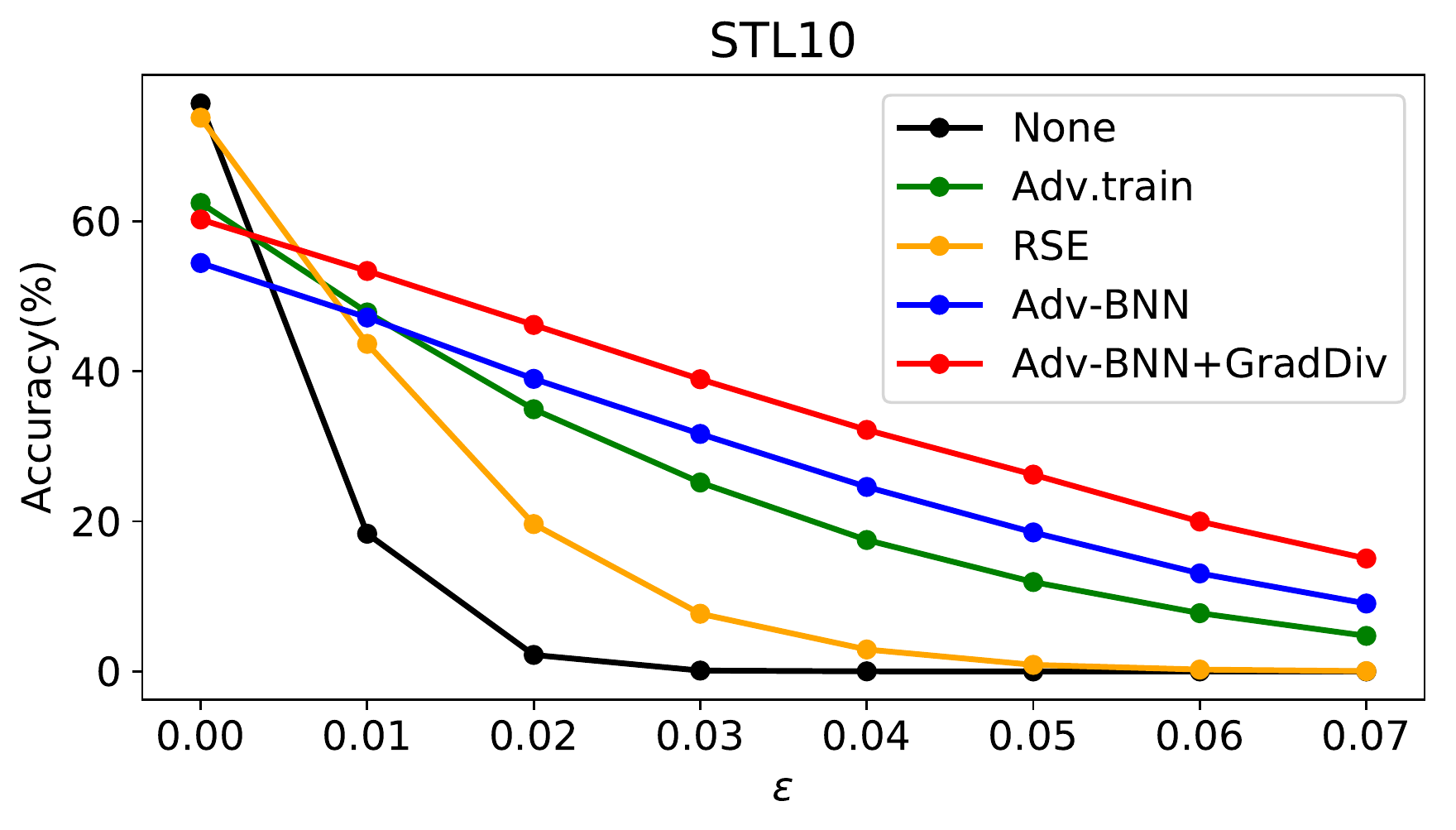}
   \caption{\textbf{The test accuracy against the EOT-PGD attack on MNIST (top) and STL10 (bottom).}}
\label{fig:acc_mnist}
\end{center}
\end{figure}
\noindent\textbf{Robustness against Adversarial Attacks.}\quad
For comparison, we use several deterministic/stochastic defense models: (1) a standard deterministic neural network trained with clean training images referred as \textit{No defense}, (2) a neural network trained with the PGD adversarial images referred as Adversarial Training (\textit{Adv.train})\cite{madry2017towards}, (3) Random Self-Ensemble (\textit{RSE})\cite{liu2018towards}, (4) Parametric Noise Injection (\textit{PNI})\cite{he2019parametric}, (5) Mixup Inference (\textit{MI})\cite{pang2019mixup}, (6) \textit{Adv-BNN}\cite{liu2018adv}, and (7) \textit{Adv-BNN + GradDiv}, an Adv-BNN model trained with GradDiv.
For all randomized networks, we use 20 sample models for the ensemble.
We note that the number of the ensemble is not sensitive above 20.
% For RSE, we set the init-noise level as $\sigma_\text{init}=0.2$ for the first noise layer and the inner-noise level as $\sigma_\text{inner}=0.1$ for the other noise layers, the same as in\cite{liu2018towards}.
% Among many GradDiv regularization models, 

% We note that there is no significant difference between the regularization methods as implied in Figure \ref{fig:reg}.

% To these defense models, we apply a diverse set of adversarial attacks. 
% We emphasize that we follow\cite{athalye2018obfuscated} and evaluate the robustness against the EOT attacks, i.e., every attack is based on the expected gradients. 
We test the defense models against a diverse set of state-of-the-art adversarial attacks: FGSM\cite{goodfellow2014explaining}, PGD\cite{madry2017towards}, APGD$_\text{CE}$\cite{croce2020reliable}, APGD$_\text{DLR}$\cite{croce2020reliable}, B\&B\cite{brendel2019accurate}, FAB\cite{croce2020minimally}, and Square\cite{andriushchenko2020square}.
For each attack, we use three types of the attack: 
(1) one uses a fixed sample model throughout attack iterations,
(2) another one uses a sample model for each iteration,
and (3) the other uses $n$ sample models for each iteration.
% (1) one uses a gradient from a fixed sample model,
% (2) another one uses a sample gradient for each gradient update,
% and (3) the other uses a sample mean for each update.
We name them [Attack], EOT$^{1}$-[Attack] and EOT-[Attack], respectively.

Table \ref{table:acc_cifar10} shows the test accuracy against several attacks on CIFAR10. %, including EOT-FGSM and EOT-PGD with various sample sizes $n$
We use the number of sample models $n=10$ for the EOT attacks and attack iteration $m=50$ for the PGD attacks.
% For EOT-FGSM, we use the number of sample gradients $n=10$, the step-size $\alpha=\epsilon$, and the attack iteration $m=1$.
% For EOT-PGD, we use $n=10$, $\alpha=\epsilon/4$, and $m=50$.
% The comparison results of the models are shown in Table \ref{table:acc_cifar10}. % and Figure \ref{fig:acc_mnist}./
GradDiv successfully improves the robustness of the baseline, outperforming the other methods in most of the cases. 
% 더자세히
% with a cost of standard accuracy drop similar to that of Adv.train and Adv-BNN.
% While the improvement caused by GradDiv in the case $n=20$ is marginal, there are significant performance gaps between Adv-BNN+GradDiv and the others in other cases, especially for the smaller $n$ and smaller attack iterations (EOT-FGSM).
% For the results on MNIST and STL10, we refer the readers to the supplementary material.
Surprisingly, GradDiv often improves the clean accuracy of the model compared to the baseline. 
This improvement can be attributed to the diversity of the sample models which improves the performance of the ensemble inference model.
% \hl{blackbox, score-based attack}
% Black-box attack (the Square attack) is not as effective as white-box attacks against the stochastic defenses.

% EOT with n=20 얘기
Figure \ref{fig:acc_mnist} shows the test accuracy against EOT-PGD ($n=20,m=20$) with the perturbation bound $\epsilon$ on MNIST and STL10. 
It shows that our models outperform the other models in robustness under the EOT-PGD attack.
Especially, it shows high accuracy for large $\epsilon$ on MNIST.
As shown in Figure \ref{fig:acc_mnist} (top), using GradDiv, Adv-BNN can improve the performance by 37.52\%p, 37.72\%p and 15.55\%p on $\epsilon=0.36$, 0.38 and 0.4, respectively. 
Moreover, Adv-BNN+GradDiv shows higher accuracy than Adv-BNN for every $\epsilon$ on STL10 as shown in Figure \ref{fig:acc_mnist} (bottom).
We observed no significant difference when using a larger $n$ in this setting.
\\

\begin{figure}[b]
\begin{center}
       \includegraphics[width=\linewidth]{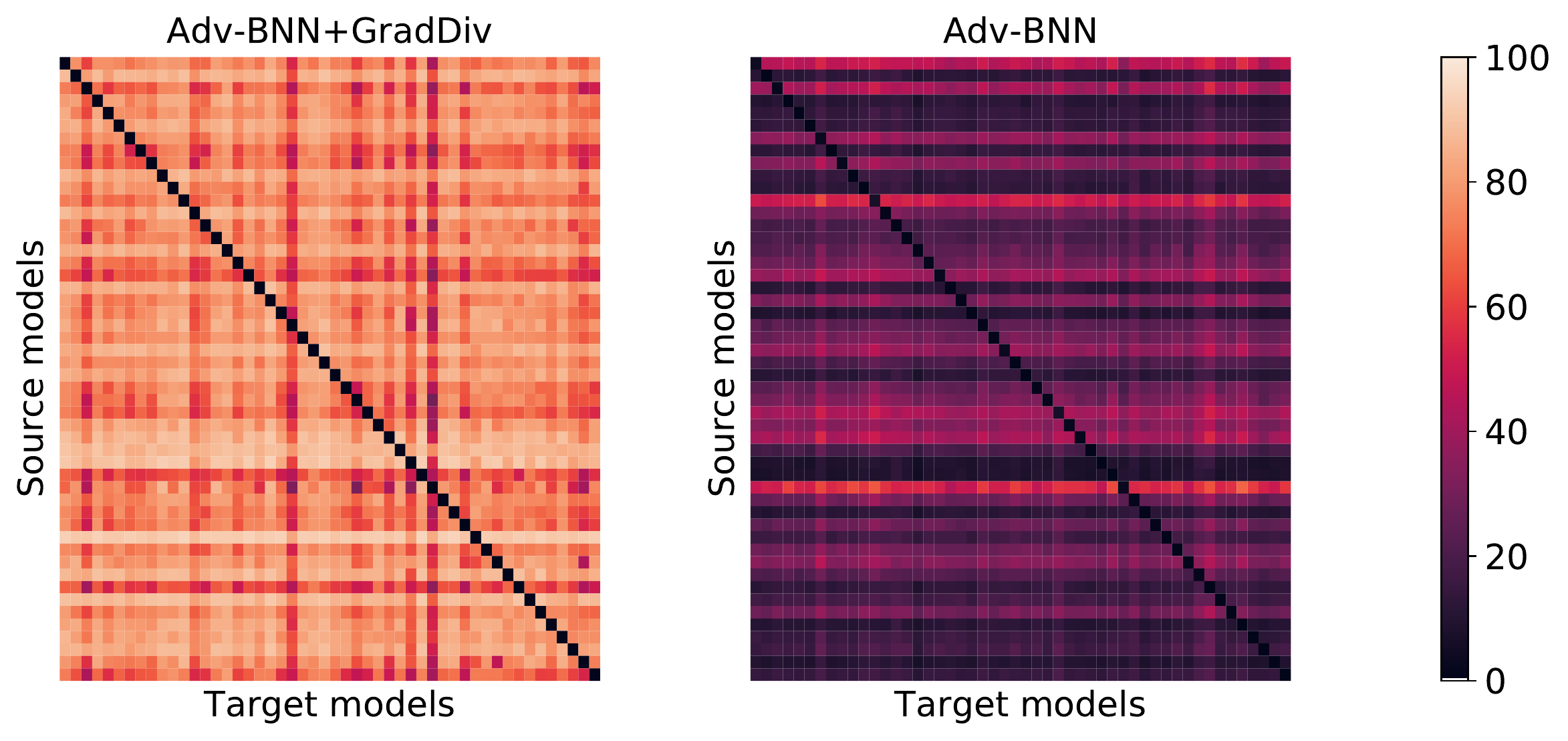}
\end{center}
   \caption{\textbf{Transferability among sample models of Adv-BNN+GradDiv (left) and Adv-BNN (right).} Lighter colors indicate higher/better test accuracy, i.e., lower transferability.}
\label{fig:blackbox}
\end{figure}

\noindent\textbf{Discussion on the Empirical Robustness.}\quad
Empirical study on defenses against a finite set of adversarial attacks has inherent limitations that they cannot guarantee the non-existence of a stronger adaptive attack which can break the defenses.
Thus, we evaluate GradDiv with a diverse set of state-of-the-art adversarial attack methods.
% Even though we focus on the white-box setting, 
We emphasize that randomized neural networks, at least, have advantages over deterministic models under query-based black-box attacks
% such as\cite{chen2017zoo,bhagoji2017exploring,ilyas2018black,ilyas2018prior}
since it is much harder for the adversary to estimate the actual gradient when the oracle is stochastic.
Furthermore, with limited queries, the model trained with GradDiv is more effective than the other methods as shown in the experiments on the limited sample size $n$.\\ % in Table \ref{table:acc_cifar10}.\\

\noindent\textbf{Checklist for Gradient Obfuscation.}\quad
Following \cite{athalye2018obfuscated}, we check whether GradDiv shows the following characteristic behaviors of defenses which cause obfuscated gradients:
\begin{itemize}%[leftmargin=20pt]
    \item[1.] One-step attacks perform better than iterative attacks.
    \item[2.] Black-box attacks are better than whit-box attacks.
    \item[3.] Unbounded attacks do not reach 100\% success.
    \item[4.] Random sampling finds adversarial examples.
    \item[5.] Increasing distortion bound does not increase success.
\end{itemize}
The evidence that GradDiv does not show the behaviors is demonstrated in Table \ref{table:acc_cifar10}
for the first behaviors, and in Figure \ref{fig:acc_mnist} for the third and the last behaviors.
For the second behavior, we found that our model has achieved the robust accuracy $>$74\% against the Square attacks (together with the EOT variants) i.e., black-box attack is weaker than white-box attacks.
%See the supplementary material for the details under black-box settings.
For the fourth behavior, we found that the EOT-PGD attack is strong enough that if it fails, then brute-force random search with $10^5$ samples also can not find adversarial examples.
\\

\noindent\textbf{Transferability between Sample Models.}\quad
To further study the effect of the GradDiv regularizations, we evaluate the transferability among sample models of Adv-BNN and Adv-BNN+GradDiv-DPP. 
We sample 50 sample models for each randomized network and test PGD attacks with $\epsilon=0.4$ and the attack iterations $m=20$ on MNIST. 
% Note that we do not have to use the EOT attack because each sample model is deterministic.

We report the results of the transfer attacks from the source models (rows) to the target models (columns)
% in $50\times 50$ matrices 
as in Figure \ref{fig:blackbox}.
The sample models trained with GradDiv are less transferable to each other, achieving an average test accuracy (off-diagonal) of 75\% (min/max: 29/94\%), while the samples from Adv-BNN show higher transferability with an average test accuracy of 24\% (min/max: 4.5/68\%). 
This result implies that GradDiv significantly lowers the transferability among the sample models as desired.

% \noindent\textbf{Cosine Similarities}\quad To compare the cosine similarities among the sample gradients, we sample 100 sample gradients for Adv-BNN and Adv-BNN+GradDiv trained on MNIST, and plot the cosine similarity matrix as in Figure \ref{suppfig:cos_plot}. The sample gradients from Adv-BNN+GradDiv are significantly less aligned than those from Adv-BNN.

% \begin{figure}[t]
% \begin{center}
%   \includegraphics[width=.9\linewidth]{0fig/cos_plot_100.pdf}
% \end{center}
%   \caption{\textbf{Cosine similarities among gradient samples of Adv-BNN\cite{liu2018adv}+GradDiv (left) and Adv-BNN\cite{liu2018adv} (right).} Darker colors indicate lower/better cosine similarity.
%   }
% \label{suppfig:cos_plot}
% \end{figure}

\section{Conclusion}
Randomized defenses are potentially promising defense methods against adversarial attacks, especially in the black-box settings with limited query.
Unfortunately, previous attempts have been broken by the proxy-gradient-based attack.
In this paper, we investigate the effect of the proxy-gradient-based attack on the randomized neural networks and demonstrate that the proxy gradient is less effective when the gradients are more scattered. 
Based on the analysis, we propose GradDiv regularizations that penalize the gradients concentration to confuse the adversary about the true gradient.
We show that GradDiv improves the robustness of randomized neural networks.
We hope the future work on randomized defenses can adopt GradDiv to further improve their performance.

% \section*{Acknowledgment}
% This work was supported by the National Research Foundation of Korea (NRF) grant funded by the Korean government (MSIT) (NRF-2019R1A2C2002358).

\bibliographystyle{IEEEtran}
\bibliography{IEEEabrv,ms} %{IEEEabrv,../bib/paper}

\newpage

\onecolumn 

\section*{Supplements to GradDiv: Adversarial Robustness of Randomized Neural Networks via Gradient Diversity Regularization}

\mbox{}

\subsection*{Details of the Experimental Setup}
\subsubsection*{Network Architectures}
In this section, we provide the details of the network architectures used in the experiments. We denote the convolutional layer with the output channel c, the kernel k, the stride s as C(c, k, s) (or C(c, k, s, p) if it uses the padding p), the linear layer with the output channel c as F(c), the maxpool layer and the average pool layer with kernel size 2 and stride 2 as MP and AP, respectively, and the batch normalization layer as BN. We also denote ReLU layer and Leaky ReLU layer as ReLU and LReLU, respectively. Note that we omit the flatten layer before the first linear layer.%\cite{ioffe2015batch}
\begin{itemize}
    \item MNIST: C(32,3,1,1)-LReLU-C(64,3,1,1)-LReLU-MP-C(128,3,1,1)-LReLU-MP-F(1024)-LReLU-F(10)
    \item CIFAR10: VGG-16
    \item STL10: C(32,3,1,1)-BN-ReLU-MP-C(64,3,1,1)-BN-ReLU-MP-C(128,3,1,1)-BN-ReLU-MP-C(256,3,1,1)-BN-ReLU-MP\\-C(256,3,1)-BN-ReLU-C(512,3,1)-BN-ReLU-AP-F(10)
\end{itemize}

\subsubsection*{Batch-size, Training Epoch, Learning rate decay, Warm-up, and Ramp-up periods}

Table \ref{supp:params} summarizes the details of the training parameters. We use the Adam optimizer for all datasets.
\begin{table*}[b]
\caption{\textbf{The details of the training parameters.} The learning rate is decayed at the epochs listed in the lr decay column with decay rate 0.1. }
\centering
\begin{tabular}{|c|c|c|c|c|c|c|}
\hline
Datasets & batch-size           & epoch & learning rate (initial)     & lr decay             & warm-up & ramp-up \\ \hline
MNIST    & \multirow{3}{*}{128} & 60    & \multirow{3}{*}{0.001} & {[}30{]}          & 3       & 20      \\ \cline{1-1} \cline{3-3} \cline{5-7}
STL10    &                      & 120   &                        & {[}60,100{]}     & 6       & 60      \\ \cline{1-1} \cline{3-3} \cline{5-7}
CIFAR10  &                      & 200   &                        & {[}80,140,180{]} & 6       & 90      \\ \hline
\end{tabular}
\label{supp:params}
\end{table*}

\subsection*{Cosine Similarities}

\begin{figure}[h]
\begin{center}
  \includegraphics[width=.5\linewidth]{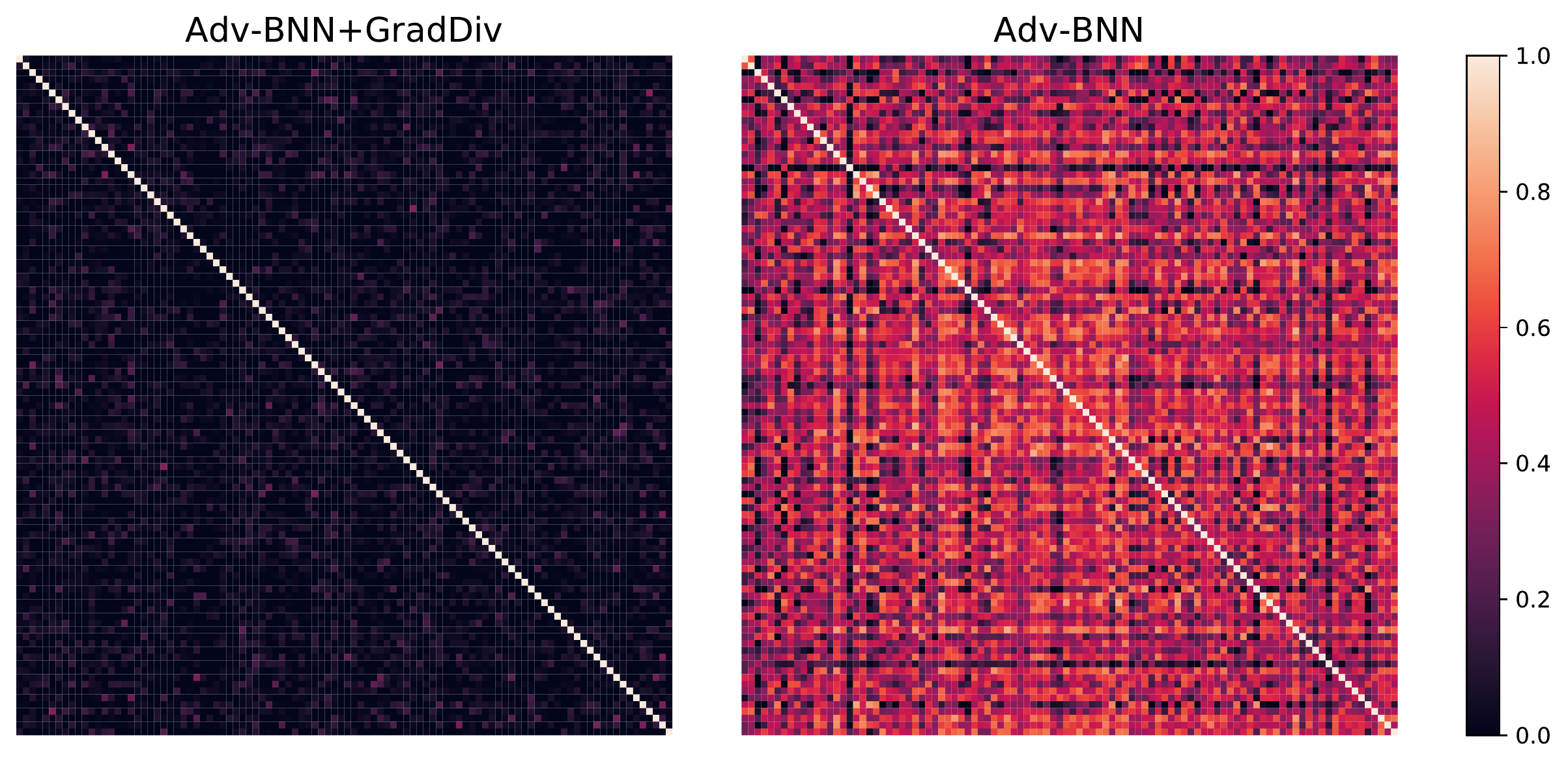}
\end{center}
  \caption{\textbf{Cosine similarities among gradient samples of Adv-BNN+GradDiv (left) and Adv-BNN (right).} Darker colors indicate lower/better cosine similarity.
  }
\label{suppfig:cos_plot}
\end{figure}

To compare the cosine similarities among the sample gradients, we sample 100 sample gradients for Adv-BNN and Adv-BNN+GradDiv trained on MNIST, and plot the cosine similarity matrix as in Figure \ref{suppfig:cos_plot}. The sample gradients from Adv-BNN+GradDiv are significantly less aligned than those from Adv-BNN.

\subsection*{Variants of GradDiv-mean (\ref{eqn:reg_cos})}
As mentioned in the main paper, we also tested variants of GradDiv-mean (\ref{eqn:reg_cos}). We propose two variants, $R_\text{max}(X;\theta)$ and $R_\text{smoothmax}(X;\theta)$. $R_\text{max}(X;\theta)$ uses the maximum value of the cosine similarity instead of the mean value: $R_\text{max}(X;\theta)=\max\{\cos(g_i, g_j)\}_{i\neq j}$, and $R_\text{smoothmax}(X;\theta)$ uses the smooth maximum (LogSumExp): $R_\text{smoothmax}(X;\theta)=\log\{\sum_{i\neq j}\exp\cos(g_i, g_j)\}$.

\begin{figure*}[b]
\begin{center}
   \includegraphics[width=0.32\linewidth]{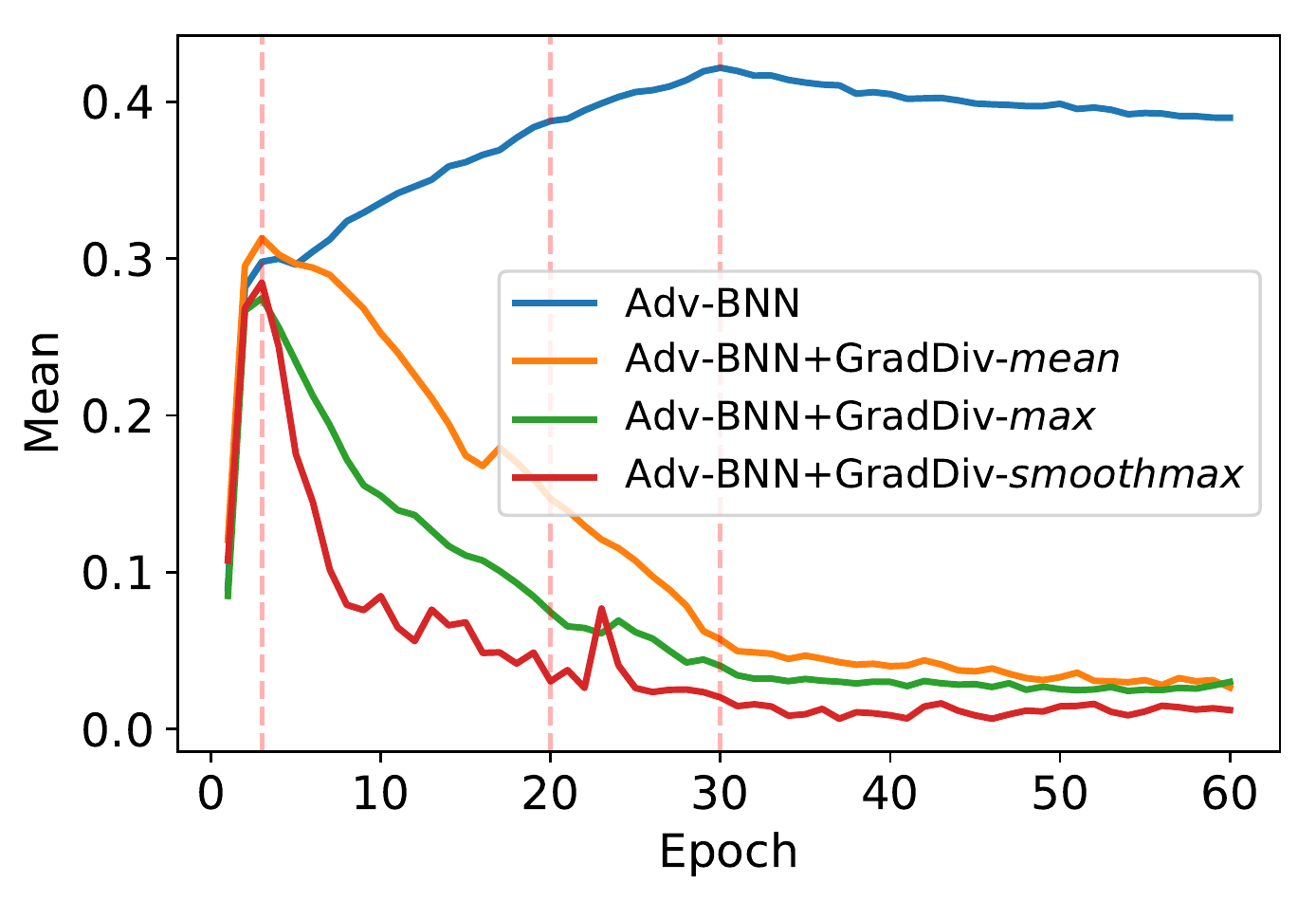}
   \includegraphics[width=0.32\linewidth]{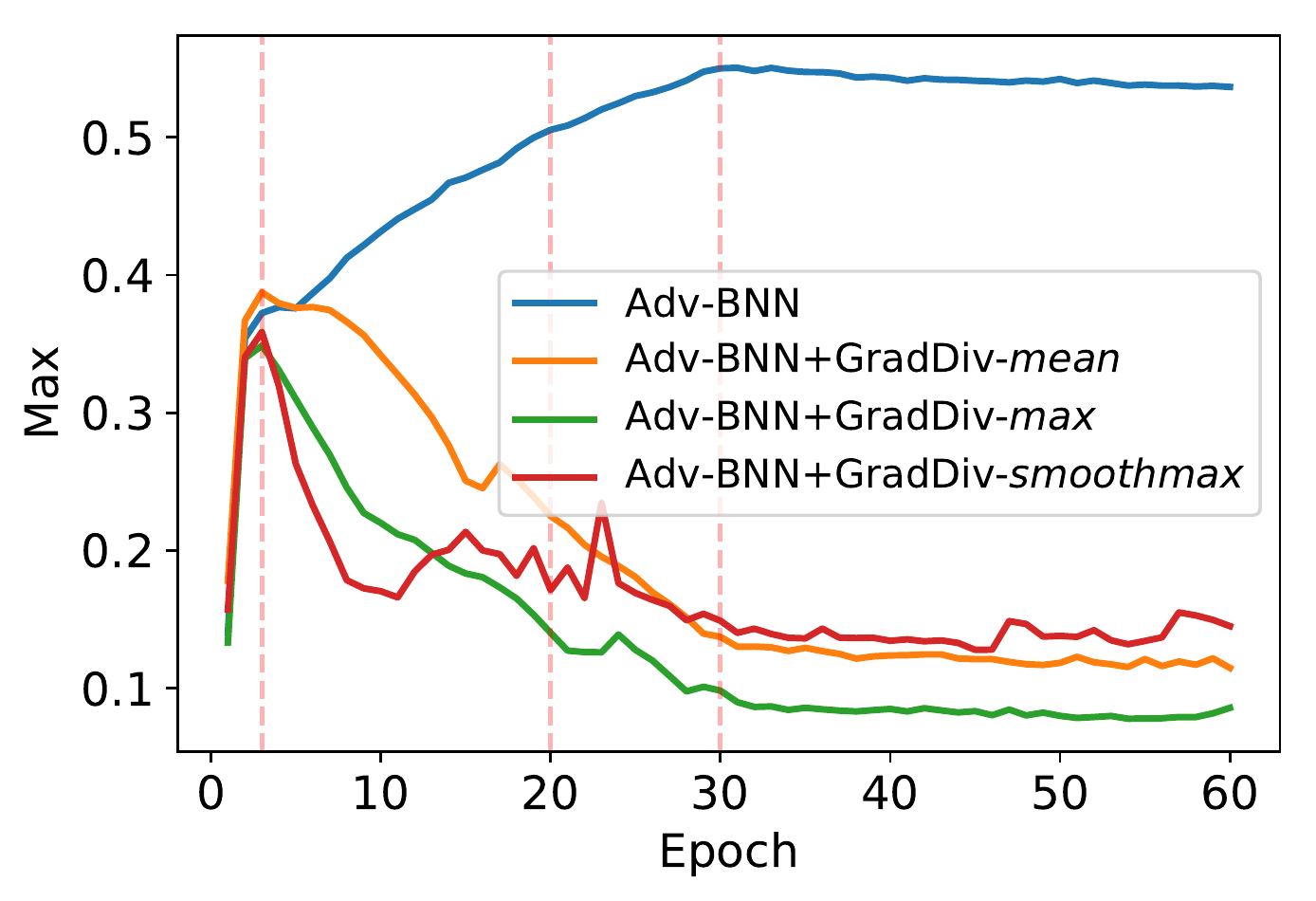}
   \includegraphics[width=0.32\linewidth]{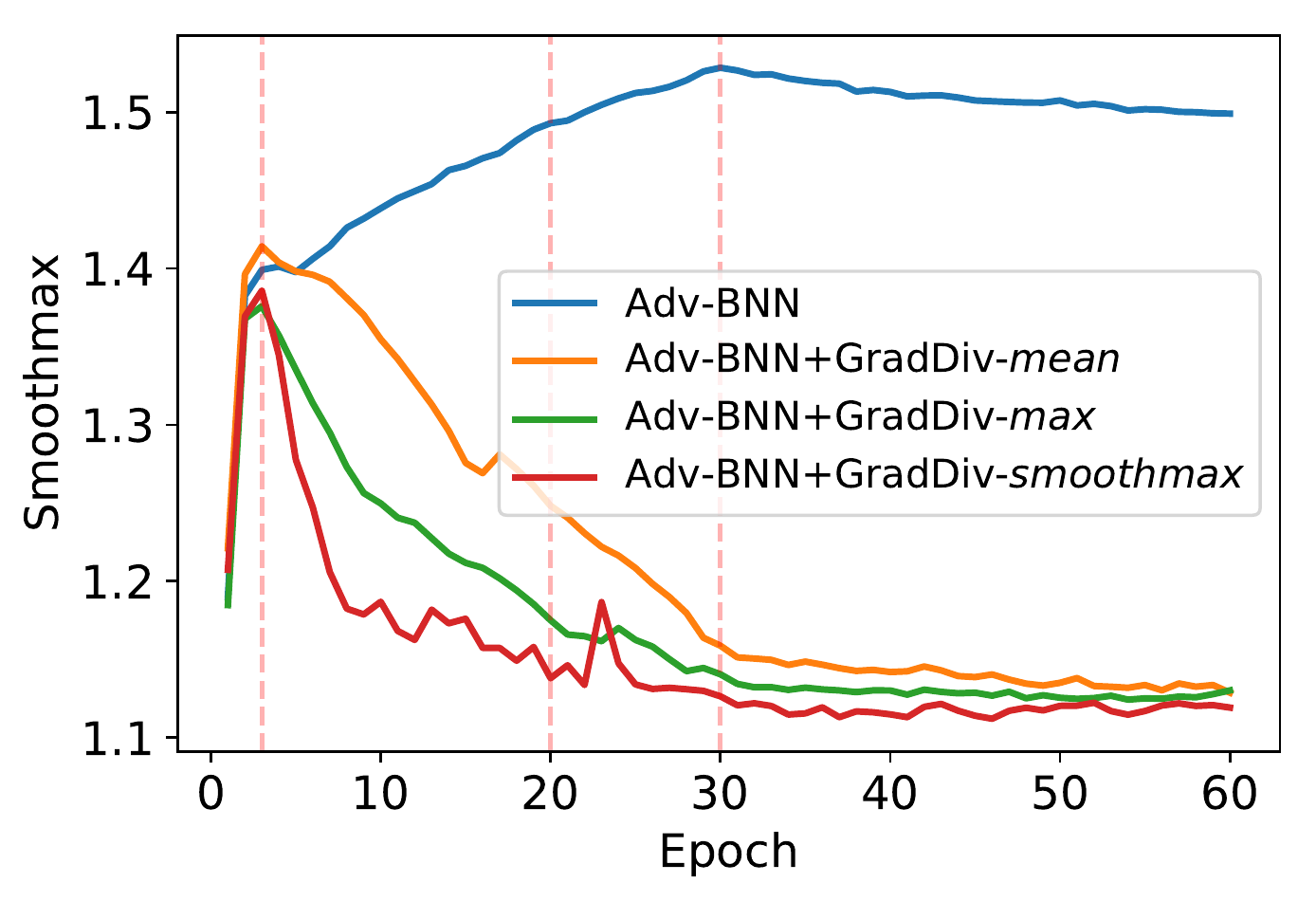}
\end{center}
    \caption{\label{suppfig:reg}\textbf{The change in the additional concentration measures during training.} The two leftmost vertical lines in each graph indicate the end of the warm-up and ramp-up periods, and the third vertical line indicates when the learning rate has decayed.}
\end{figure*}

\begin{figure*}[t]
\begin{center}%[width=0.95\linewidth]
   \includegraphics[width=0.95\linewidth]{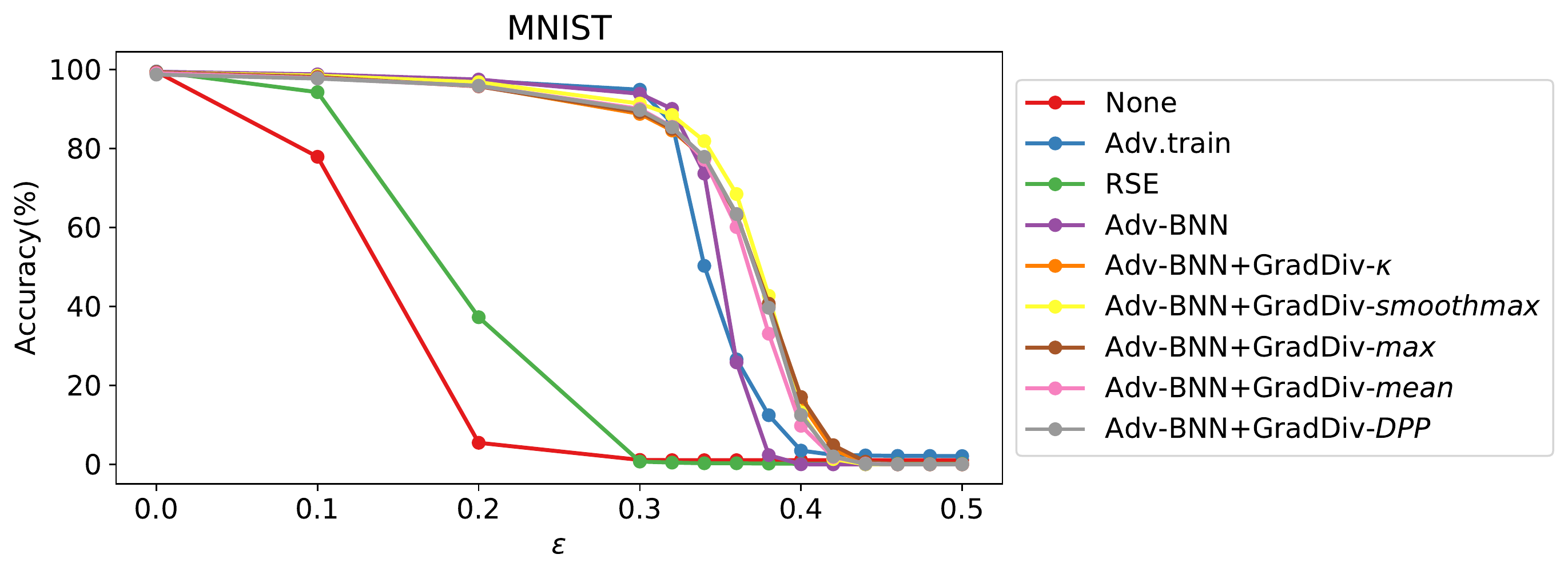}
   \caption{\textbf{The test accuracy against the EOT-PGD attack on MNIST with the proposed regularizers.}}
\label{suppfig:acc_mnist}
\end{center}
\end{figure*}

Figure \ref{suppfig:reg} shows the effects of the variants on each objective and Figure \ref{suppfig:acc_mnist} shows the test accuracy against EOT-PGD on MNIST ($n=20$, $\alpha=\epsilon/4$, and $m=20$). There is no significant performance difference among the regularizers.

%We can also consider variants of the (\ref{eqn:reg_cos}) such as $L_{max}(X;\theta)$ $=$ $\max\{\cos(g_i, g_j)\}$ and $L_{smooth\_max}(X;\theta)$ $=$ $\log\{\sum_{i\neq j}\exp\cos(g_i, g_j)\}$

\comment{
\subsection*{Note on Adv-BNN\cite{liu2018adv}}
%(randomized classifier가 중점이므로 깊게 갈 필요는 없음)
In the main paper, we use Bayesian Neural Network (BNN) structure for the randomized classifier $F=f_{W}$ which means that\shortcomment{ for each unit of the network} the weight $W$ is sampled from the posterior distribution $p(W|\mathcal{D})$ of weight given the training data $\mathcal{D}$, and the predictive probability $P(Y|X)$ for the output $Y$ given the test input $X$ is $P(Y|X)=\mathbb{E}_{p(W|\mathcal{D})}[P(Y|X,W)]$. However, it is intractable to compute the expectation over the true posterior $p(W|\mathcal{D})$. Therefore, variational learning utilizes a parameterized proxy distribution $q_\theta(W)$ and minimizes the Kullback-Leibler (KL) divergence with the true posterior $p(W|\mathcal{D})$ to learn the parameter $\theta$:
\begin{align} % TODO : equation numbering
    &\text{KL}(q_\theta(W)\Vert p(W|\mathcal{D}))=\int_{\Omega} q_\theta(W) \log \frac{q_\theta(W)}{p(W|\mathcal{D})} dW\nonumber\\
    &=\int_{\Omega}q_\theta(W) \log \frac{q_\theta(W)}{p(W)} -q_\theta(W) \log p(\mathcal{D}|W) dW\nonumber\\
    &=\text{KL}(q_\theta(W)\Vert p(W)) - \mathbb{E}_{q_\theta(W)}[\log p(\mathcal{D}|W)].
\end{align}

In the paper, we suppose the posterior is a diagonal normal distribution $W\sim\mathcal{N}(\vmu,\Sigma=\text{diag}\{\sigma_i\})$. Given training data $\mathcal{D}$, the parameter $\theta$ ($\vmu$ and $\Sigma$) of the posterior distribution is leaned by minimizing the following cost function:
\begin{equation}
\label{eqn:loss_bnn}
    \mathcal{J}(\mu,\Sigma)=\alpha_{KL}\text{KL}(q_{\mu,\Sigma}(W)\Vert p(W))-\mathbb{E}_{W\sim q_{\mu,\Sigma}}\text{log }p(\mathcal{D}|W)
\end{equation}
where the first KL divergence term is called the complexity cost, and the second term is called the likelihood cost. In the case of a diagonal Gaussian, the complexity term can be explicitly formulated as follows:
\begin{equation}
    \text{KL}(q_{\mu,\Sigma}(W)\Vert p(W))=\sum_i\log \frac{\sigma_0}{\sigma_i}+\frac{\sigma_i^2+(\mu_i-\mu_0)^2}{2\sigma_0^2}-\frac{1}{2}\label{eqn:KL}
\end{equation} with the prior parameters $\mu_0$ and $\sigma_0$. Also, the likelihood term is replaced by the standard cross entropy loss
\begin{equation}
    \sum_{(X_i,Y_i)\in \mathcal{D}} \mathbb{E}_{W\sim q_{\mu,\Sigma}} \mathcal{L}(f_{W}(X_i),Y_i).
    \label{eqn:cross_ent}
\end{equation}\shortcomment{ where $I_\text{train}$ is the index set of the training data.}

Therefore, the objective for each data $(X,Y)$ is represented as follows:
\begin{equation}
\label{eqn:bnn_last}
    \mathcal{J}(\mu,\Sigma|X)=\frac{\alpha_{KL}}{|\mathcal{D}|}\text{KL}(q_{\mu,\Sigma}(W)\Vert p(W))-\mathcal{L}(f_{w}(X),Y)
\end{equation} with a sample weight $w$.
In Adv-BNN\cite{liu2018adv}, they train a randomized defense model using an adversarial training data $\mathcal{D}^{adv}=\{(X_i^{adv},Y_i)\}$ instead of the standard training data $\mathcal{D}$ in (\ref{eqn:bnn_last}).
}

\subsection*{Additional Results on "Effects of GradDiv during Training"}
\begin{figure*}[t]
\begin{center}
  \includegraphics[width=0.32\linewidth]{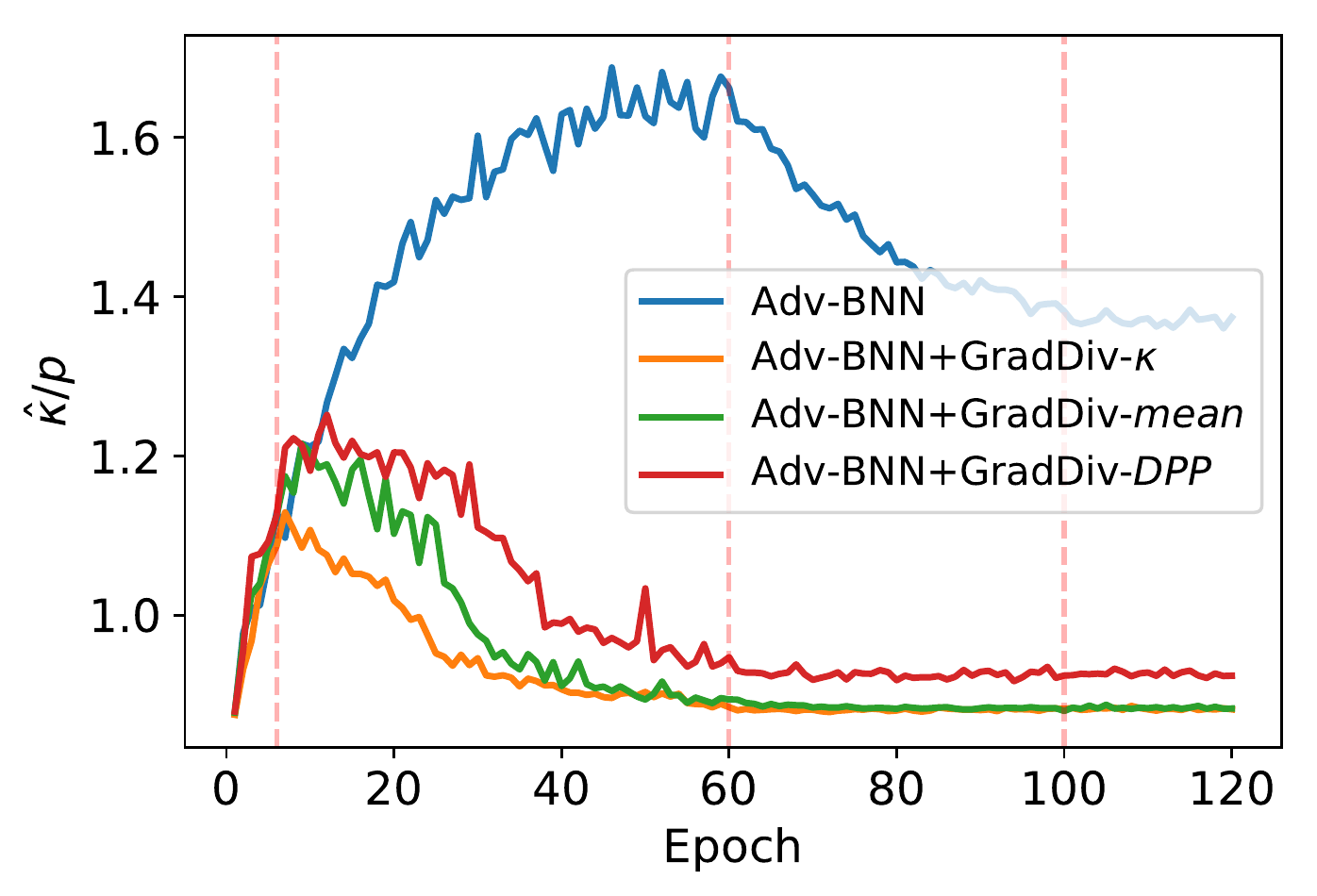}
  \includegraphics[width=0.32\linewidth]{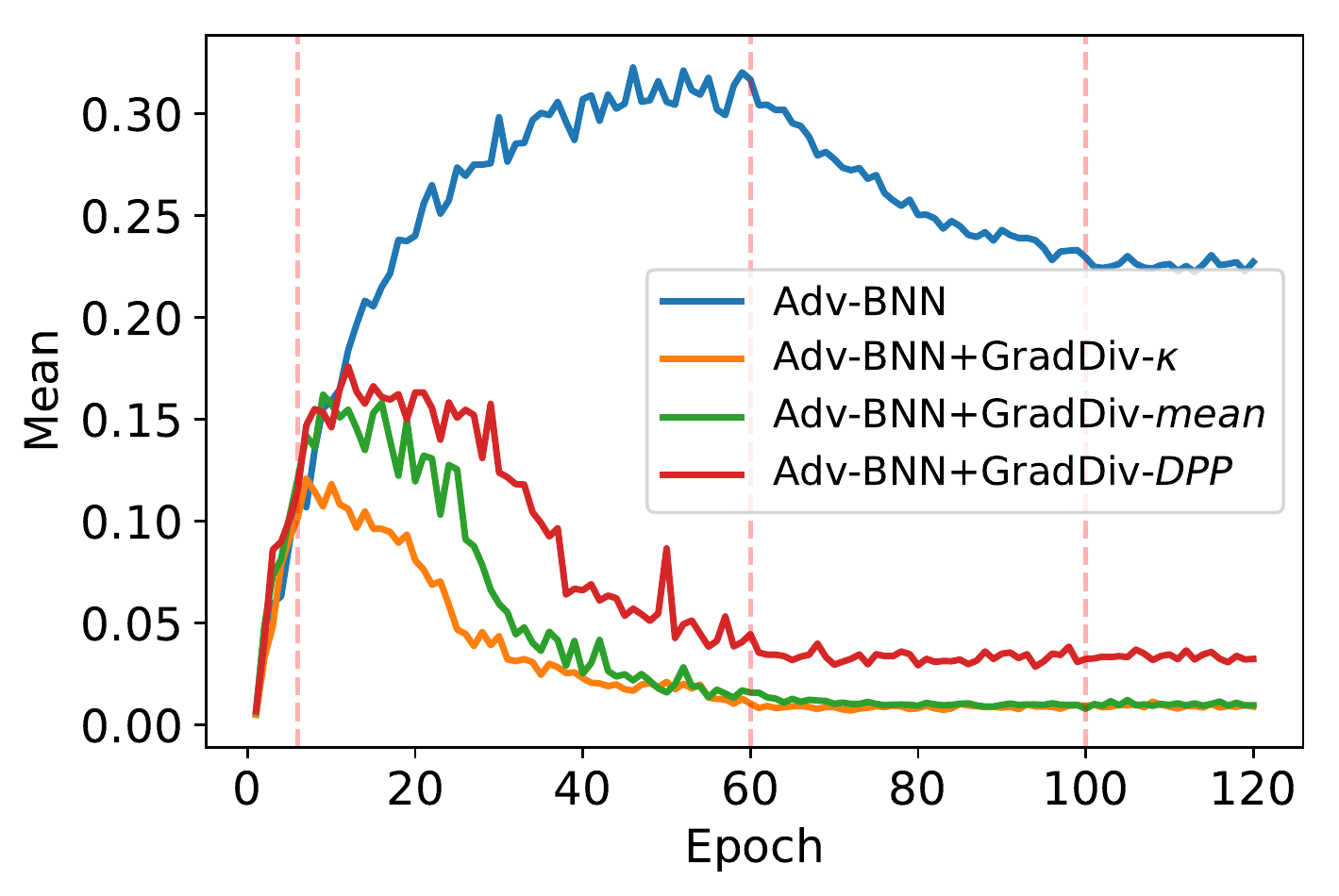}
  \includegraphics[width=0.32\linewidth]{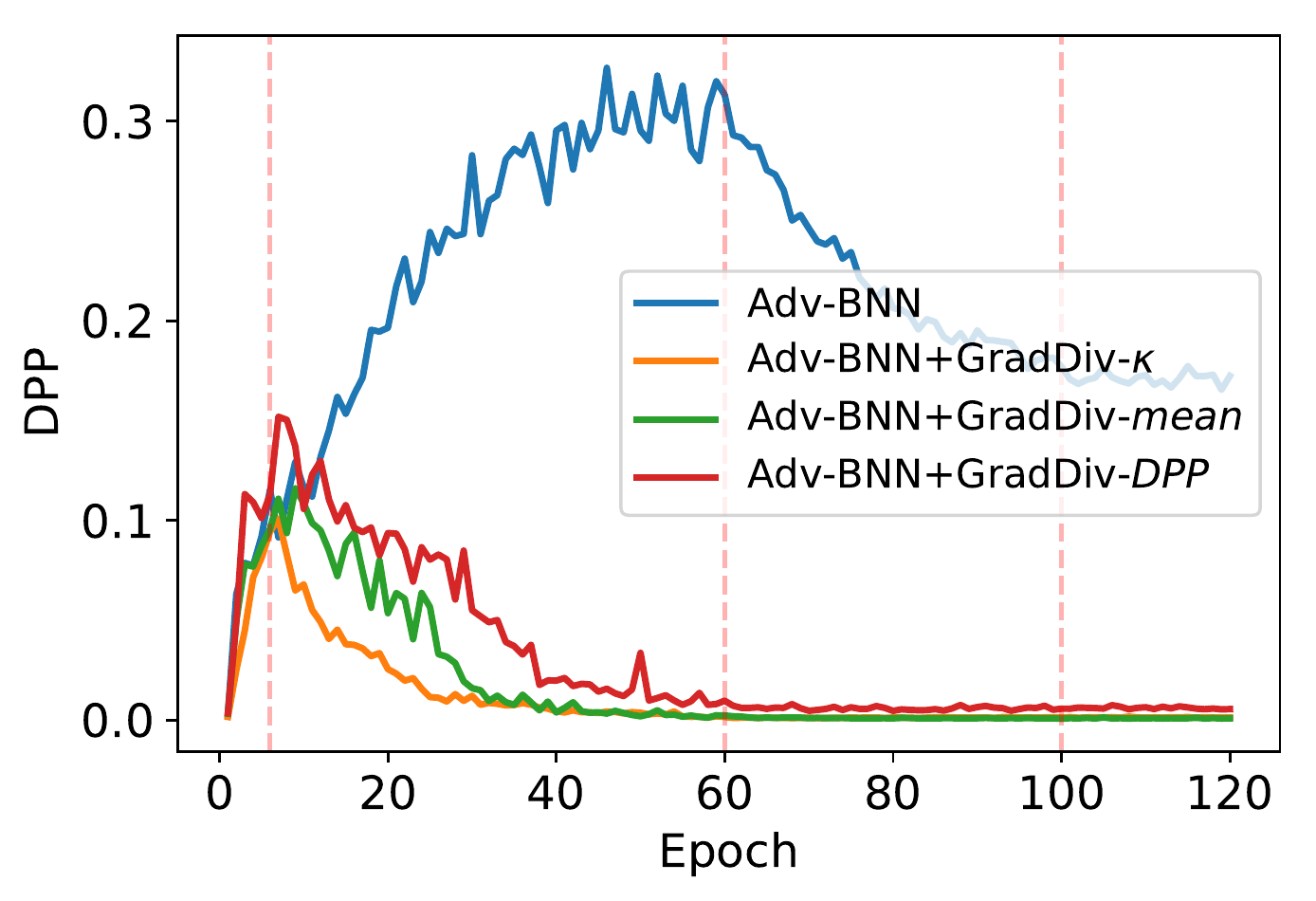}
\end{center}
    \caption{\textbf{The change in the concentration measures (\ref{eqn:reg_kappa}) (left), (\ref{eqn:reg_cos}) (middle), and (\ref{eqn:reg_DPP}) (right) during training on STL10.} The two leftmost vertical lines in each graph indicate the end of the warm-up and ramp-up periods, and the two rightmost vertical lines indicate when the learning rate has decayed.} %TODO: 혹시 다끝나고 순서확인
\label{suppfig:reg_stl}
\end{figure*}

\begin{figure*}[t]
\begin{center}
   \includegraphics[width=0.32\linewidth]{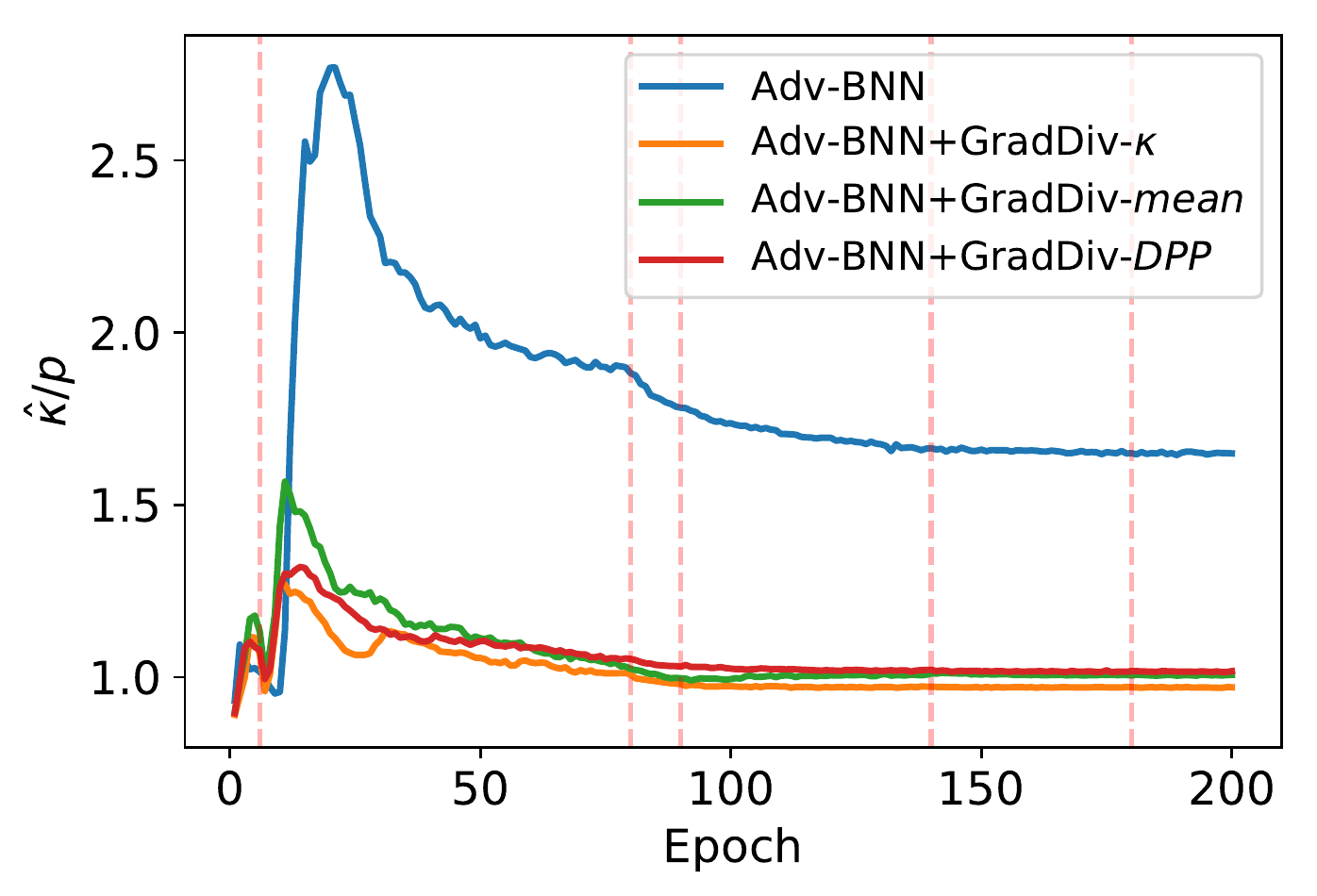}
   \includegraphics[width=0.32\linewidth]{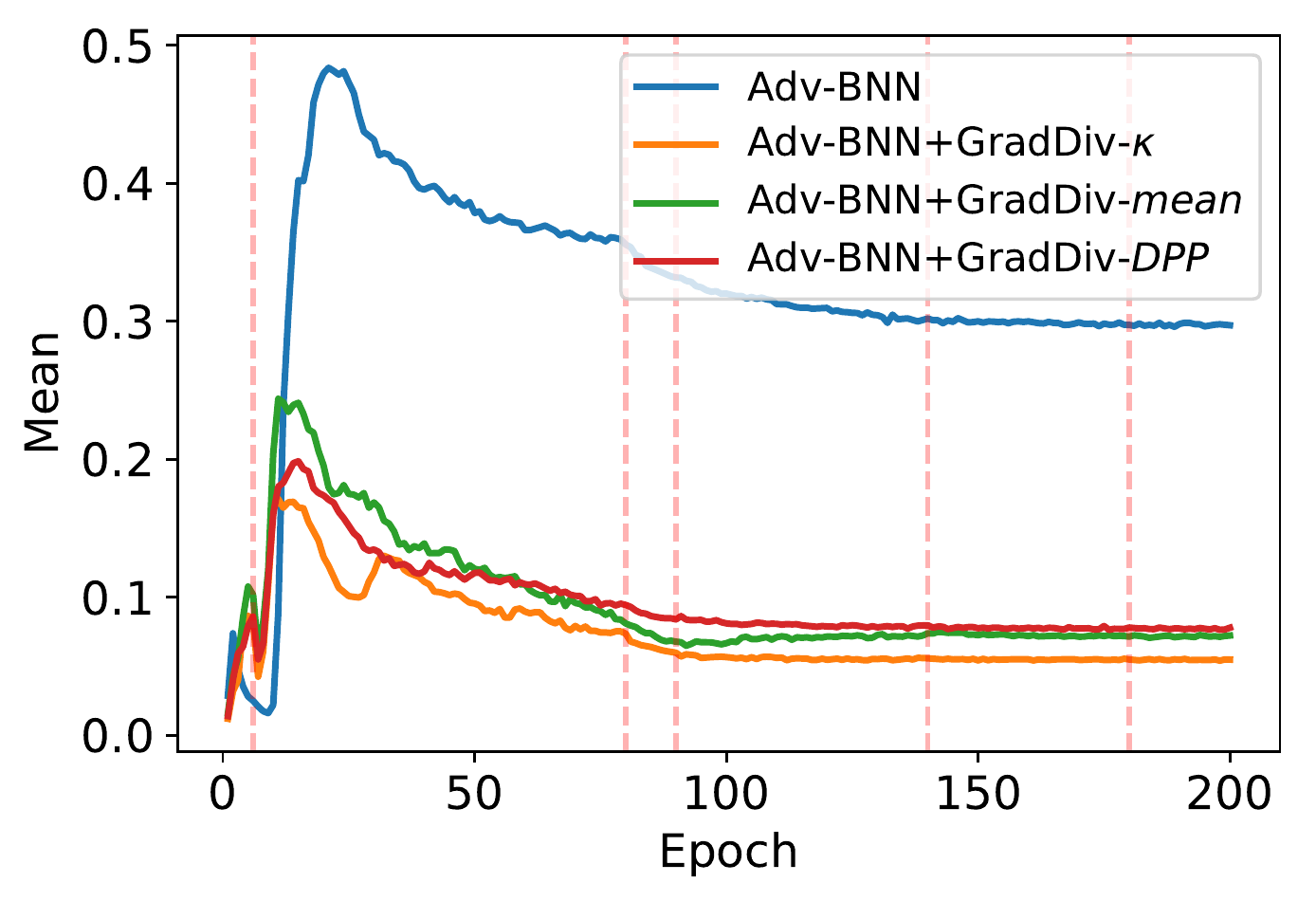}
   \includegraphics[width=0.32\linewidth]{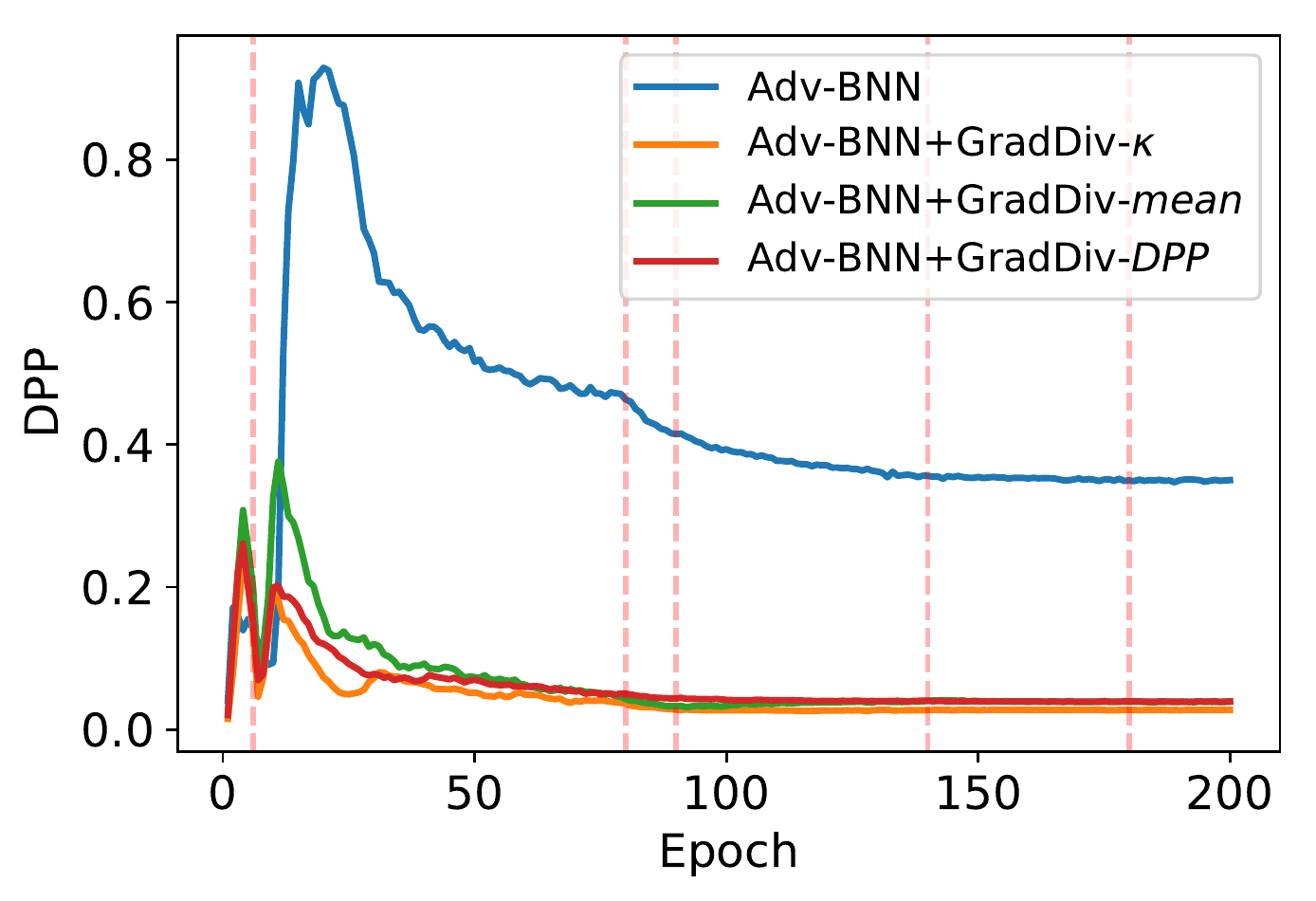}
\end{center}
    \caption{\textbf{The change in the concentration measures (\ref{eqn:reg_kappa}) (left), (\ref{eqn:reg_cos}) (middle), and (\ref{eqn:reg_DPP}) (right) during training on CIFAR10.} The two leftmost vertical lines in each graph indicate the end of the warm-up and ramp-up period, and the three rightmost vertical lines indicate when the learning rate has decayed.} %TODO: 혹시 다끝나고 순서확인
\label{suppfig:reg_cifar}
\end{figure*}

Figure \ref{suppfig:reg_stl} and \ref{suppfig:reg_cifar} show the effects of GradDiv on the concentration measures (\ref{eqn:reg_kappa}), (\ref{eqn:reg_cos}), and (\ref{eqn:reg_DPP}) during training on STL10 and CIFAR10, respectively. We observed similar results with Figure \ref{fig:reg}.

\subsection*{Additional Results on Table \ref{table:acc_cifar10}}
In Table \ref{supptable:acc_mnist} and \ref{supptable:acc_stl}, we also present additional results as Table \ref{table:acc_cifar10} in the main paper for the other datasets, MNIST and STL10.

\subsection*{In the Case of $n>20$ in Figure \ref{fig:acc_mnist}}
\begin{figure}[t]
\begin{center}
   \includegraphics[width=0.5\linewidth]{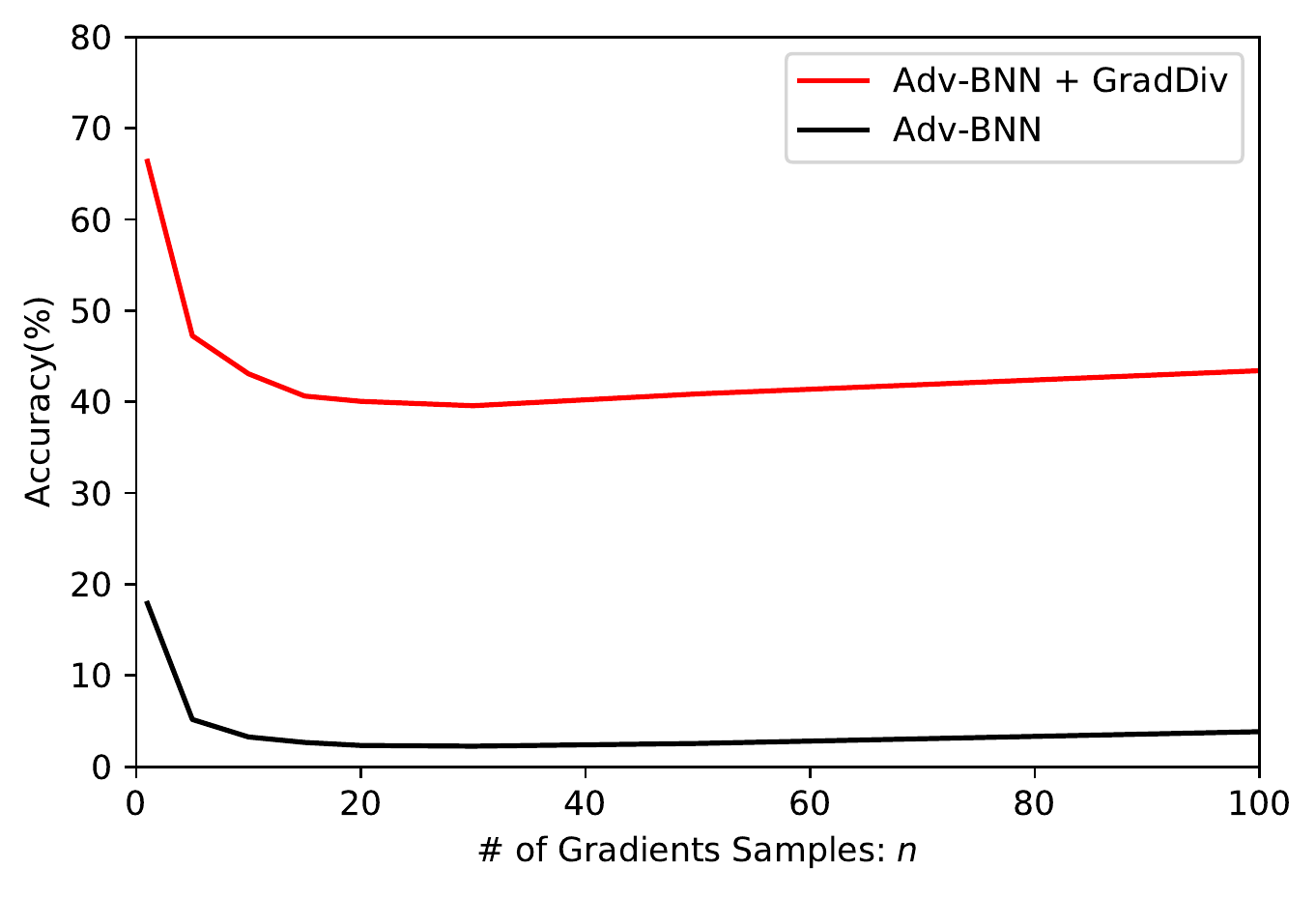}
\end{center}
   \caption{\textbf{The effectiveness of the EOT-PGD attack with different gradient sample size $n$.}
   }
\label{suppfig:sample_sens}
\end{figure}

As mentioned in the main paper, we use the gradient sample size $n=20$ in Figure \ref{fig:acc_mnist}. Figure \ref{suppfig:sample_sens} shows accuracy when different gradient sample sizes $n$ are used. Note that when $n>20$ the EOT-PGD attack on Adv-BNN+GradDiv becomes even weaker as $n$ increases. Therefore, we reported the worst-case results when $n=20$.
In the experiment, we use the EOT-PGD attack with $\epsilon=0.38$, $\alpha=\epsilon/4$ and the attack iteration $m=20$ on MNIST.

% \subsection*{Additional Results on "Transferability between Sample Models"}
% \red{결과넣기 호기}
% \subsection*{Cosine Similarities}
% To compare the cosine similarities among the sample gradients, we sample 100 sample gradients for Adv-BNN and Adv-BNN+GradDiv trained on MNIST, and plot the cosine similarity matrix as in Figure \ref{suppfig:cos_plot}. The sample gradients from Adv-BNN+GradDiv are less aligned than those from Adv-BNN.

\subsection*{RSE as a Baseline}
In the main paper, we use Adv-BNN\cite{liu2018adv} as a baseline and show increase of robustness with our regularizers. In addition to this, we test the proposed method on RSE\cite{liu2018towards} as a new baseline and demonstrate the results in Table \ref{supptable:acc_rse}. In the experiment, RSE using (GradDiv-$\kappa$, 1) outperforms the baseline on CIFAR10.

% \subsection*{$l_2$-Attack}
% \red{잘나오면 넣기 성윤}

% \begin{figure}[t]
% \begin{center}
%   \includegraphics[width=0.95\linewidth]{0fig/theta_40_.pdf} %cosgraph
% \end{center}
%   \caption{The effectiveness of PGD attack using proxy gradients instead of the actual gradients. The proxy gradients are obtained by rotating the true gradients by $\theta$ with a random axis at every iterations of the attack. The loss difference caused by the proxy gradients (red) decreases, and the accuracy under the PGD attack (blue) increases as $\theta$ increases to $90^\circ$.\shortcomment{ Note that for $\theta=90^\circ$, the accuracy is almost same as the accuracy for the clean data.} Note that the graph of the loss difference (red) appears similar to the cosine curve, empirically demonstrating (\ref{eqn:delta}).
%   }
% \label{fig:cosgraph}
% \end{figure}

% present the experimental results to

% \begin{figure*}[t]
% \begin{center}
%   \includegraphics[width=0.32\linewidth]{0fig/reg_kappa.pdf}
%   \includegraphics[width=0.32\linewidth]{0fig/reg_mean.pdf}
%   \includegraphics[width=0.32\linewidth]{0fig/reg_dpp.pdf}
% \end{center}
%     \caption{The change of the concentration measures (\ref{eqn:reg_kappa}) (left), (\ref{eqn:reg_cos}) (middle), and (\ref{eqn:reg_DPP}) (right) during training. The two leftmost vertical lines in each graph indicate the end of the warm-up and ramp-up period, and the third vertical line indicates when the learning rate is decayed.} %TODO: 혹시 다끝나고 순서확인
% \label{fig:reg}
% \end{figure*}

% \begin{figure}[t]
% \begin{center}
% %   \includegraphics[width=.9\linewidth]{0fig/MNIST_2d.pdf}
%   \includegraphics[width=.8\linewidth]{0fig/2d_plot_final0.pdf}
% \end{center}
%   \caption{The scatter plot of the gradient samples on the unit circle. The dots indicate the gradient samples from the randomized neural network and the lines indicate the sample mean vectors of the gradients with the length of the sample MRL $\hat{\rho}$ (black: Adv-BNN, red: Adv-BNN+GradDiv).}
% \label{fig:2d}
% \end{figure}

% \begin{figure}[t]
% \begin{center}
% %   \includegraphics[width=.9\linewidth]{0fig/MNIST_2d.pdf}
%   %\includegraphics[width=.95\linewidth]{0fig/Delta_MNIST_63.pdf}
%   \includegraphics[width=.95\linewidth]{0fig/Delta_MNIST_73.pdf}
% \end{center}
%   \caption{The density plots of the loss increase under the EOT attack. }
% \label{fig:Delta}
% \end{figure}
\begin{table*}[]
\caption{
\label{supptable:acc_mnist}T\textbf{he test accuracy against adversarial attacks on MNIST.} The best results are highlighted in bold.}
\resizebox{\textwidth}{!}{
\begin{tabular}{c|c|c|c|crrrllllr}
\hline
Dataset                 & \multicolumn{2}{c|}{Attack}                                                                    & Method          & 0                                  & \multicolumn{1}{c}{0.1} & \multicolumn{1}{c}{0.2} & \multicolumn{1}{c}{0.3} & 0.32           & 0.34           & 0.36           & 0.38           & \multicolumn{1}{c}{0.4} \\ \hline
\multirow{16}{*}{MNIST} & \multicolumn{2}{c|}{\multirow{5}{*}{\begin{tabular}[c]{@{}c@{}}EOT-\\ FGSM\end{tabular}}}      & None            & \multicolumn{1}{r}{99.41}          & 92.48                   & 61.24                   & 27.3                    & 23.66          & 20.67          & 18.50           & 17.08          & 15.95                   \\
                        & \multicolumn{2}{c|}{}                                                                          & Adv.train       & \multicolumn{1}{r}{99.40}           & 98.70                    & 97.95                   & 97.42                   & 96.69          & 93.22          & 86.47          & 77.05          & 62.74                   \\
                        & \multicolumn{2}{c|}{}                                                                          & RSE            & \multicolumn{1}{r}{\textbf{99.44}} & 96.08                   & 82.52                   & 47.76                   & 40.11          & 33.73          & 28.37          & 23.71          & 20.03                   \\
                        & \multicolumn{2}{c|}{}                                                                          & Adv-BNN         & \multicolumn{1}{r}{99.38}          & \textbf{98.85}          & \textbf{98.18}          & \textbf{97.23}          & \textbf{97.28} & \textbf{97.11} & \textbf{96.28} & 92.82          & 85.02                   \\
                        & \multicolumn{2}{c|}{}                                                                          & Adv-BNN+GradDiv & \multicolumn{1}{r}{99.14}          & 98.43                   & 97.39                   & 96.03                   & 95.46          & 95.02          & 94.18          & \textbf{93.57} & \textbf{92.32}          \\ \cline{2-13}
                        & \multirow{11}{*}{\begin{tabular}[c]{@{}c@{}}EOT\\ -PGD\end{tabular}} & \multirow{2}{*}{-}      & None            & -                                  & 77.87                   & 5.46                    & 1.12                    & 1.06           & 1.05           & 1.05           & 1.05           & 1.05                    \\
                        &                                                                      &                         & Adv.train       & -                                  & 98.53                   & 97.16                   & 94.85                   & 86.41          & 50.26          & 26.57          & 12.44          & 3.48                    \\ \cline{3-13}
                        &                                                                      & \multirow{3}{*}{$n=5$}  & RSE            & -                                  & 94.71                   & 47.55                   & 1.63                    & 0.82           & 0.47           & 0.38           & 0.34           & 0.33                    \\
                        &                                                                      &                         & Adv-BNN         & -                                  & \textbf{98.80}           & \textbf{97.65}          & \textbf{94.32}          & \textbf{90.62} & 74.83          & 28.55          & 5.18           & 0.15                    \\
                        &                                                                      &                         & Adv-BNN+GradDiv & -                                  & 98.27                   & 96.33                   & 90.12                   & 86.67          & \textbf{80.23} & \textbf{68.34} & \textbf{47.24} & \textbf{21.37}          \\ \cline{3-13}
                        &                                                                      & \multirow{3}{*}{$n=10$} & RSE            & -                                  & 94.37                   & 41.39                   & 1.03                    & 0.51           & 0.38           & 0.30            & 0.22           & 0.25                    \\
                        &                                                                      &                         & Adv-BNN         & -                                  & \textbf{98.76}          & \textbf{97.59}          & \textbf{93.88}          & \textbf{90.07} & 73.87          & 25.38          & 3.26           & 0.05                    \\
                        &                                                                      &                         & Adv-BNN+GradDiv & -                                  & 98.19                   & 96.08                   & 89.21                   & 85.36          & \textbf{78.34} & \textbf{64.88} & \textbf{43.07} & \textbf{17.31}          \\ \cline{3-13}
                        &                                                                      & \multirow{3}{*}{$n=20$} & RSE            & -                                  & 94.23                   & 37.25                   & 0.70                     & 0.44           & 0.27           & 0.27           & 0.19           & 0.15                    \\
                        &                                                                      &                         & Adv-BNN         & -                                  & \textbf{98.74}          & \textbf{97.43}          & \textbf{93.83}          & \textbf{90.00}    & 73.62          & 25.82          & 2.33           & 0.02                    \\
                        &                                                                      &                         & Adv-BNN+GradDiv & -                                  & 97.99                   & 95.80                    & 88.73                   & 84.54          & \textbf{77.58} & \textbf{63.34} & \textbf{40.05} & \textbf{15.57}          \\ \hline
\end{tabular}
}
\end{table*}

% sigma = 0.1
\begin{table*}[]
\caption{
\label{supptable:acc_stl}\textbf{The test accuracy against adversarial attacks on STL.} The best results are highlighted in bold.}
\resizebox{\textwidth}{!}{
\begin{tabular}{c|c|c|c|crrrrrrr}
\hline
Dataset                 & \multicolumn{2}{c|}{Attack}                                                                    & Method          & 0                                  & \multicolumn{1}{c}{0.01} & \multicolumn{1}{c}{0.02} & \multicolumn{1}{c}{0.03} & \multicolumn{1}{c}{0.04} & \multicolumn{1}{c}{0.05} & \multicolumn{1}{c}{0.06} & \multicolumn{1}{c}{0.07} \\ \hline
\multirow{16}{*}{STL10} & \multicolumn{2}{c|}{\multirow{5}{*}{\begin{tabular}[c]{@{}c@{}}EOT-\\ FGSM\end{tabular}}}      & None            & \multicolumn{1}{r}{\textbf{75.69}} & 31.08                    & 16.35                    & 10.5                     & 8.05                     & 7.14                     & 6.90                      & 6.66                     \\
                        & \multicolumn{2}{c|}{}                                                                          & Adv.train       & \multicolumn{1}{r}{62.43}          & 48.53                    & 37.70                     & 29.22                    & 22.96                    & 18.16                    & 15.10                     & 12.76                    \\
                        & \multicolumn{2}{c|}{}                                                                          & RSE            & \multicolumn{1}{r}{73.65}          & 47.86                    & 26.29                    & 13.01                    & 7.18                     & 4.43                     & 2.90                      & 1.99                     \\
                        & \multicolumn{2}{c|}{}                                                                          & Adv-BNN         & \multicolumn{1}{r}{54.85}          & 48.38                    & 41.64                    & 35.24                    & 30.21                    & 25.4                     & 20.46                    & 17.14                    \\
                        & \multicolumn{2}{c|}{}                                                                          & Adv-BNN+GradDiv & \multicolumn{1}{r}{60.31}          & \textbf{56.39}           & \textbf{51.76}           & \textbf{47.94}           & \textbf{43.84}           & \textbf{40.54}           & \textbf{35.93}           & \textbf{32.62}           \\ \cline{2-12}
                        & \multirow{11}{*}{\begin{tabular}[c]{@{}c@{}}EOT\\ -PGD\end{tabular}} & \multirow{2}{*}{-}      & None            & -                                  & 18.35                    & 2.20                      & 0.11                     & 0.01                     & 0.00                        & 0.00                        & 0.00                        \\
                        &                                                                      &                         & Adv.train       & -                                  & 47.84                    & 34.94                    & 25.18                    & 17.51                    & 11.91                    & 7.76                     & 4.74                     \\ \cline{3-12}
                        &                                                                      & \multirow{3}{*}{$n=5$}  & RSE            & -                                  & 44.99                    & 21.31                    & 8.55                     & 3.39                     & 1.18                     & 0.34                     & 0.08                     \\
                        &                                                                      &                         & Adv-BNN         & -                                  & 47.40                     & 40.49                    & 33.63                    & 27.18                    & 21.08                    & 16.48                    & 11.21                    \\
                        &                                                                      &                         & Adv-BNN+GradDiv & -                                  & \textbf{54.80}            & \textbf{48.60}            & \textbf{42.43}           & \textbf{36.49}           & \textbf{30.49}           & \textbf{25.28}           & \textbf{21.04}           \\ \cline{3-12}
                        &                                                                      & \multirow{3}{*}{$n=10$} & RSE            & -                                  & 44.33                    & 20.38                    & 8.10                      & 3.06                     & 0.99                     & 0.26                     & 0.05                     \\
                        &                                                                      &                         & Adv-BNN         & -                                  & 47.64                    & 39.98                    & 32.41                    & 25.69                    & 19.35                    & 13.84                    & 9.74                     \\
                        &                                                                      &                         & Adv-BNN+GradDiv & -                                  & \textbf{54.19}           & \textbf{47.14}           & \textbf{40.75}           & \textbf{34.56}           & \textbf{28.15}           & \textbf{22.79}           & \textbf{17.75}           \\ \cline{3-12}
                        &                                                                      & \multirow{3}{*}{$n=20$} & RSE            & -                         & 43.65                    & 19.64                    & 7.69                     & 2.93                     & 0.88                     & 0.25                     & 0.04                     \\
                        &                                                                      &                         & Adv-BNN         & -                                  & 47.16                    & 38.99                    & 31.65                    & 24.59                    & 18.53                    & 13.06                    & 9.05                     \\
                        &                                                                      &                         & Adv-BNN+GradDiv & -                        & \textbf{53.36}           & \textbf{46.18}           & \textbf{38.94}           & \textbf{32.19}           & \textbf{26.23}           & \textbf{19.98}           & \textbf{15.05}           \\ \hline
\end{tabular}
}
\end{table*}

\begin{table*}[]
\caption{
\label{supptable:acc_cifar}\textbf{The test accuracy against adversarial attacks on CIFAR10.} The best results are highlighted in bold.}
\resizebox{\textwidth}{!}{
\begin{tabular}{c|c|c|c|crrrrrrr}
\hline
Dataset                   & \multicolumn{2}{c|}{Attack}                                                                    & Method          & 0                                  & \multicolumn{1}{c}{0.01} & \multicolumn{1}{c}{0.02} & \multicolumn{1}{c}{0.03} & \multicolumn{1}{c}{0.04} & \multicolumn{1}{c}{0.05} & \multicolumn{1}{c}{0.06} & \multicolumn{1}{c}{0.07} \\ \hline
\multirow{16}{*}{CIFAR10} & \multicolumn{2}{c|}{\multirow{5}{*}{\begin{tabular}[c]{@{}c@{}}EOT-\\ FGSM\end{tabular}}}      & None            & \multicolumn{1}{r}{\textbf{92.39}} & 26.57                    & 12.06                    & 8.37                     & 7.18                     & 6.83                     & 6.81                     & 6.96                     \\
                          & \multicolumn{2}{c|}{}                                                                          & Adv.train       & \multicolumn{1}{r}{78.83}          & 68.12                    & 57.65                    & 49.63                    & 42.72                    & 37.38                    & 33.14                    & 29.26                    \\
                          & \multicolumn{2}{c|}{}                                                                          & RSE            & \multicolumn{1}{r}{84.06}          & 68.49                    & 51.83                    & 35.67                    & 21.60                     & 12.90                     & 7.08                     & 4.12                     \\
                          & \multicolumn{2}{c|}{}                                                                          & Adv-BNN         & \multicolumn{1}{r}{75.97}          & 67.94                    & 59.28                    & 49.9                     & 41.59                    & 33.91                    & 27.47                    & 22.10                     \\
                          & \multicolumn{2}{c|}{}                                                                          & Adv-BNN+GradDiv & \multicolumn{1}{r}{76.45}          & \textbf{71.37}           & \textbf{65.65}           & \textbf{59.02}           & \textbf{53.69}           & \textbf{47.71}           & \textbf{42.72}           & \textbf{37.69}           \\ \cline{2-12}
                          & \multirow{11}{*}{\begin{tabular}[c]{@{}c@{}}EOT\\ -PGD\end{tabular}} & \multirow{2}{*}{-}      & None            & -                                  & 3.16                     & 0.00                        & 0.00                        & 0.00                        & 0.00                        & 0.00                        & 0.00                        \\
                          &                                                                      &                         & Adv.train       & -                                  & 66.75                    & 52.84                    & 39.49                    & 28.32                    & 18.26                    & 11.47                    & 7.23                     \\ \cline{3-12}
                          &                                                                      & \multirow{3}{*}{$n=5$}  & RSE            & -                                  & 66.39                    & 44.50                     & 23.86                    & 10.48                    & 3.98                     & 1.40                      & 0.44                     \\
                          &                                                                      &                         & Adv-BNN         & -                                  & 67.57                    & 58.06                    & 46.81                    & 36.13                    & 25.15                    & 16.24                    & 9.18                     \\
                          &                                                                      &                         & Adv-BNN+GradDiv & -                                  & \textbf{69.81}           & \textbf{61.88}           & \textbf{53.63}           & \textbf{44.52}           & \textbf{36.40}            & \textbf{28.52}           & \textbf{20.91}           \\ \cline{3-12}
                          &                                                                      & \multirow{3}{*}{$n=10$} & RSE            & -                                  & 65.95                    & 43.10                     & 22.40                     & 9.43                     & 3.24                     & 1.19                     & 0.37                     \\
                          &                                                                      &                         & Adv-BNN         & -                                  & 67.41                    & 56.85                    & 45.46                    & 33.73                    & 23.04                    & 13.81                    & 7.31                     \\
                          &                                                                      &                         & Adv-BNN+GradDiv & -                                  & \textbf{68.59}           & \textbf{59.31}           & \textbf{49.17}           & \textbf{38.96}           & \textbf{29.53}           & \textbf{20.74}           & \textbf{13.56}           \\ \cline{3-12}
                          &                                                                      & \multirow{3}{*}{$n=20$} & RSE            & -                                  & 65.48                    & 42.12                    & 21.65                    & 9.02                     & 3.18                     & 1.01                     & 0.30                      \\
                          &                                                                      &                         & Adv-BNN         & -                                  & \textbf{67.17}           & 56.36                    & 44.49                    & 32.75                    & 21.50                     & 12.27                    & 6.29                     \\
                          &                                                                      &                         & Adv-BNN+GradDiv & -                                  & 66.39                    & \textbf{56.73}           & \textbf{44.88}           & \textbf{33.81}           & \textbf{22.95}           & \textbf{14.52}           & \textbf{8.60}             \\ \hline
\end{tabular}
}
\end{table*}

% \begin{figure}[t]
% \begin{center}
%   \includegraphics[width=0.5\linewidth]{0fig/cos_plot_100.pdf}
% \end{center}
%   \caption{Cosine similarities among gradient samples of Adv-BNN\cite{liu2018adv}+GradDiv (left) and Adv-BNN\cite{liu2018adv} (right). Darker colors indicate lower cosine similarity.
%   }
% \label{suppfig:cos_plot}
% \end{figure}

\begin{table*}[]\caption{
\label{supptable:acc_rse} \textbf{The test accuracy against adversarial attacks on the baseline, RSE\cite{liu2018towards} on CIFAR10.} The best results are highlighted in bold.}
\resizebox{\textwidth}{!}{
\begin{tabular}{c|c|c|c|crrrrrrr}
\hline
Dataset                   & \multicolumn{2}{c|}{Attack}                                                                   & Method      & 0                                  & \multicolumn{1}{c}{0.01} & \multicolumn{1}{c}{0.02} & \multicolumn{1}{c}{0.03} & \multicolumn{1}{c}{0.04} & \multicolumn{1}{c}{0.05} & \multicolumn{1}{c}{0.06} & \multicolumn{1}{c}{0.07} \\ \hline
\multirow{10}{*}{CIFAR10} & \multicolumn{2}{c|}{\multirow{2}{*}{\begin{tabular}[c]{@{}c@{}}EOT-\\ FGSM\end{tabular}}}     & RSE        & \multicolumn{1}{r}{84.06}          & 68.49                    & 51.83                    & 35.67                    & 21.60                     & 12.90                     & 7.08                     & 4.12                     \\
                          & \multicolumn{2}{c|}{}                                                                         & RSE+GradDiv & \multicolumn{1}{r}{\textbf{84.88}} & \textbf{77.77}           & \textbf{69.42}           & \textbf{59.88}           & \textbf{50.10}            & \textbf{41.53}           & \textbf{33.71}           & \textbf{28.52}           \\ \cline{2-12}
                          & \multirow{8}{*}{\begin{tabular}[c]{@{}c@{}}EOT\\ -PGD\end{tabular}} & \multirow{2}{*}{$n=1$}  & RSE        & -                                  & 68.76                    & 48.15                    & 28.07                    & 14.06                    & 5.84                     & 2.29                     & 0.77                     \\
                          &                                                                     &                         & RSE+GradDiv & -                                  & \textbf{72.05}           & \textbf{54.75}           & \textbf{36.95}           & \textbf{21.83}           & \textbf{11.40}            & \textbf{6.09}            & \textbf{3.12}            \\ \cline{3-12}
                          &                                                                     & \multirow{2}{*}{$n=5$}  & RSE        & -                                  & 66.39                    & 44.50                     & 23.86                    & 10.48                    & 3.98                     & 1.40                      & 0.44                     \\
                          &                                                                     &                         & RSE+GradDiv & -                                  & \textbf{68.49}           & \textbf{47.90}            & \textbf{27.69}           & \textbf{12.83}           & \textbf{5.57}            & \textbf{2.38}            & \textbf{1.11}            \\ \cline{3-12}
                          &                                                                     & \multirow{2}{*}{$n=10$} & RSE        & -                                  & 65.95                    & 43.10                     & 22.40                     & 9.43                     & 3.24                     & 1.19                     & 0.37                     \\
                          &                                                                     &                         & RSE+GradDiv & -                                  & \textbf{67.61}           & \textbf{44.91}           & \textbf{24.53}           & \textbf{10.46}           & \textbf{4.12}            & \textbf{1.69}            & \textbf{0.73}            \\ \cline{3-12}
                          &                                                                     & \multirow{2}{*}{$n=20$} & RSE        & -                                  & 65.48                    & 42.12                    & 21.65                    & 9.02                     & 3.18                     & 1.01                     & 0.30                      \\
                          &                                                                     &                         & RSE+GradDiv & -                                  & \textbf{67.07}           & \textbf{44.88}           & \textbf{25.43}           & \textbf{11.60}           & \textbf{5.20}            & \textbf{2.59}            & \textbf{1.46}            \\ \hline
\end{tabular}
}

\end{table*}

\subsection*{Diversity of Decision Boundary}
\begin{figure}[]
    \centering
    \includegraphics[width=0.19\linewidth]{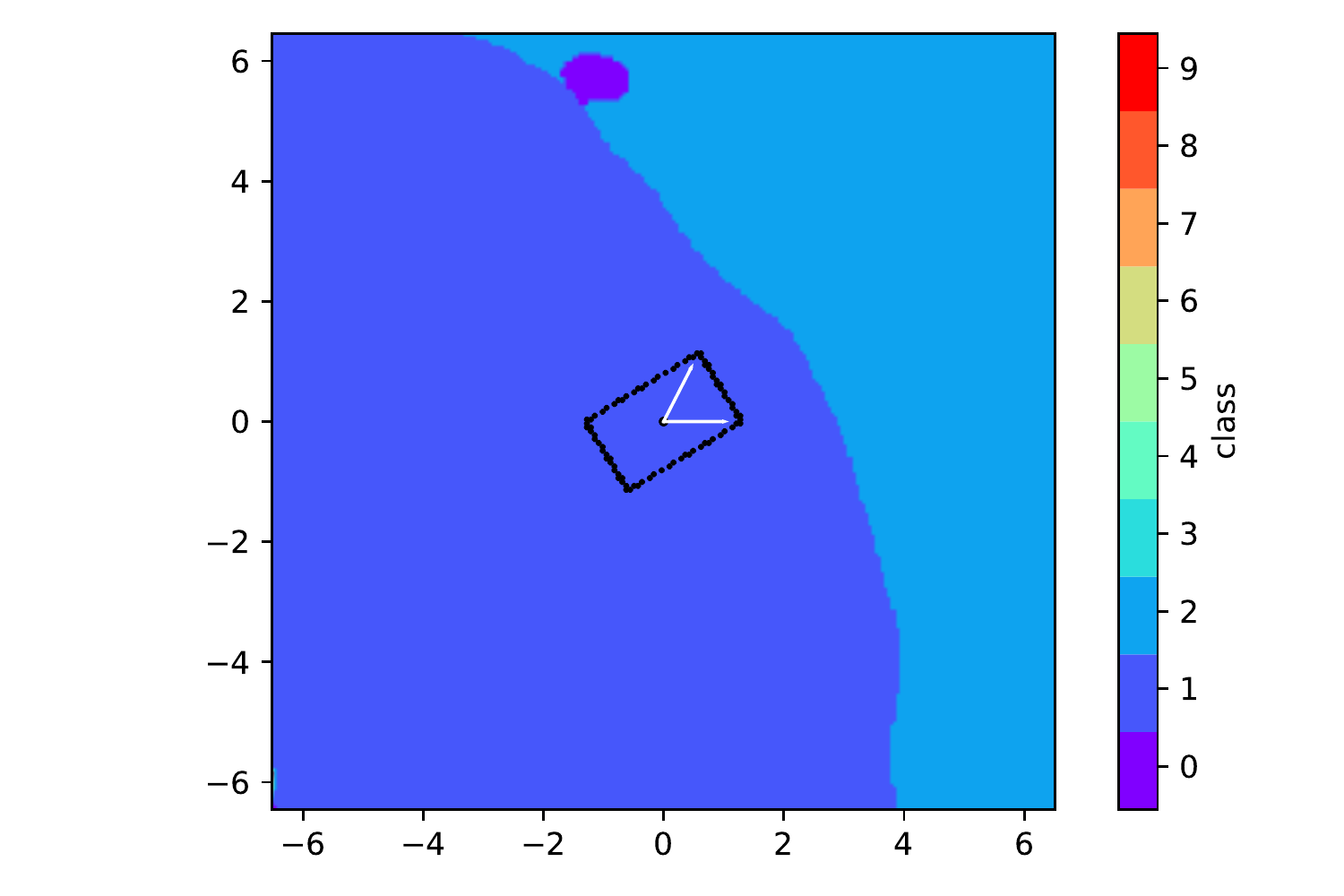}
    \includegraphics[width=0.19\linewidth]{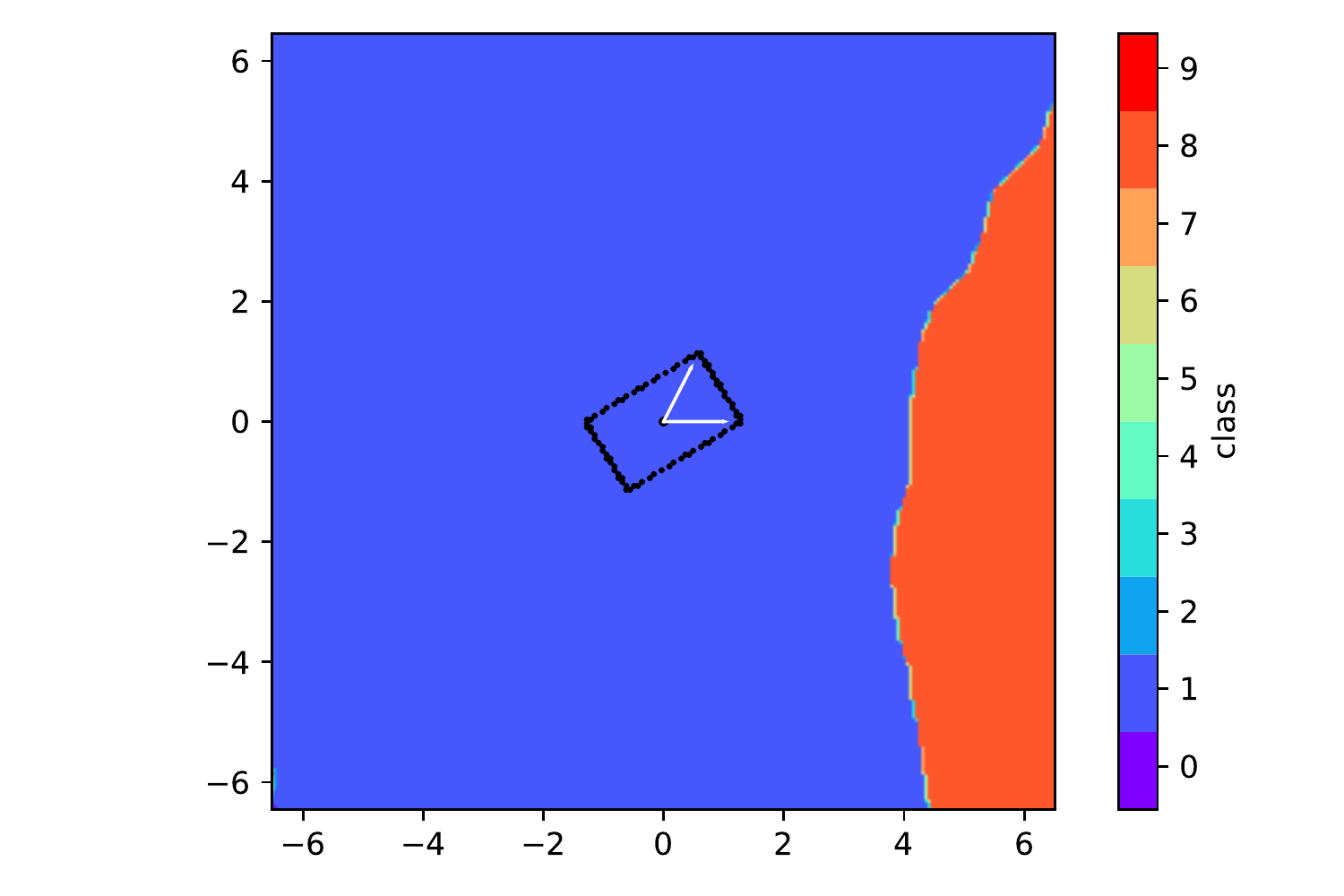}
    \includegraphics[width=0.19\linewidth]{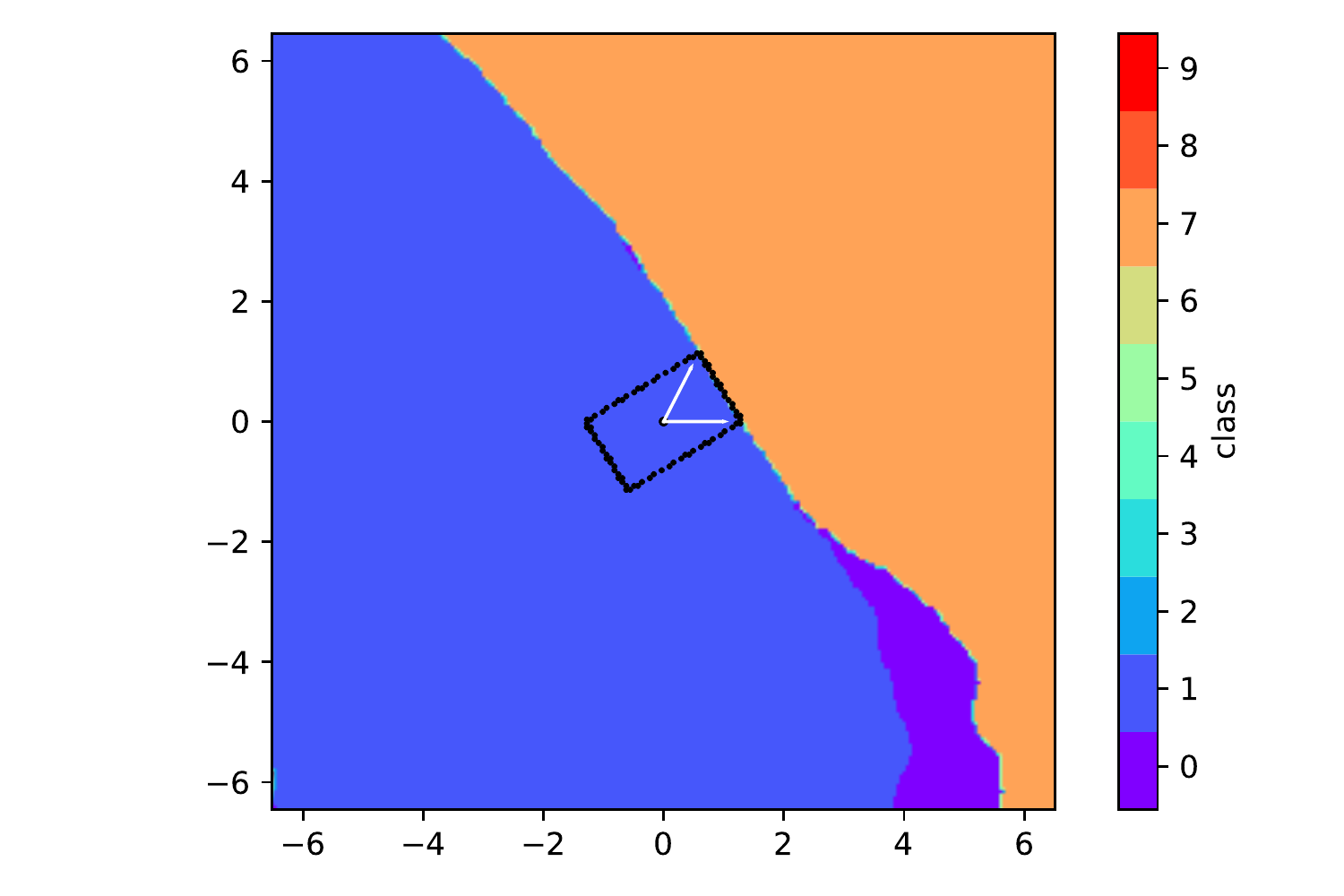}
    \includegraphics[width=0.19\linewidth]{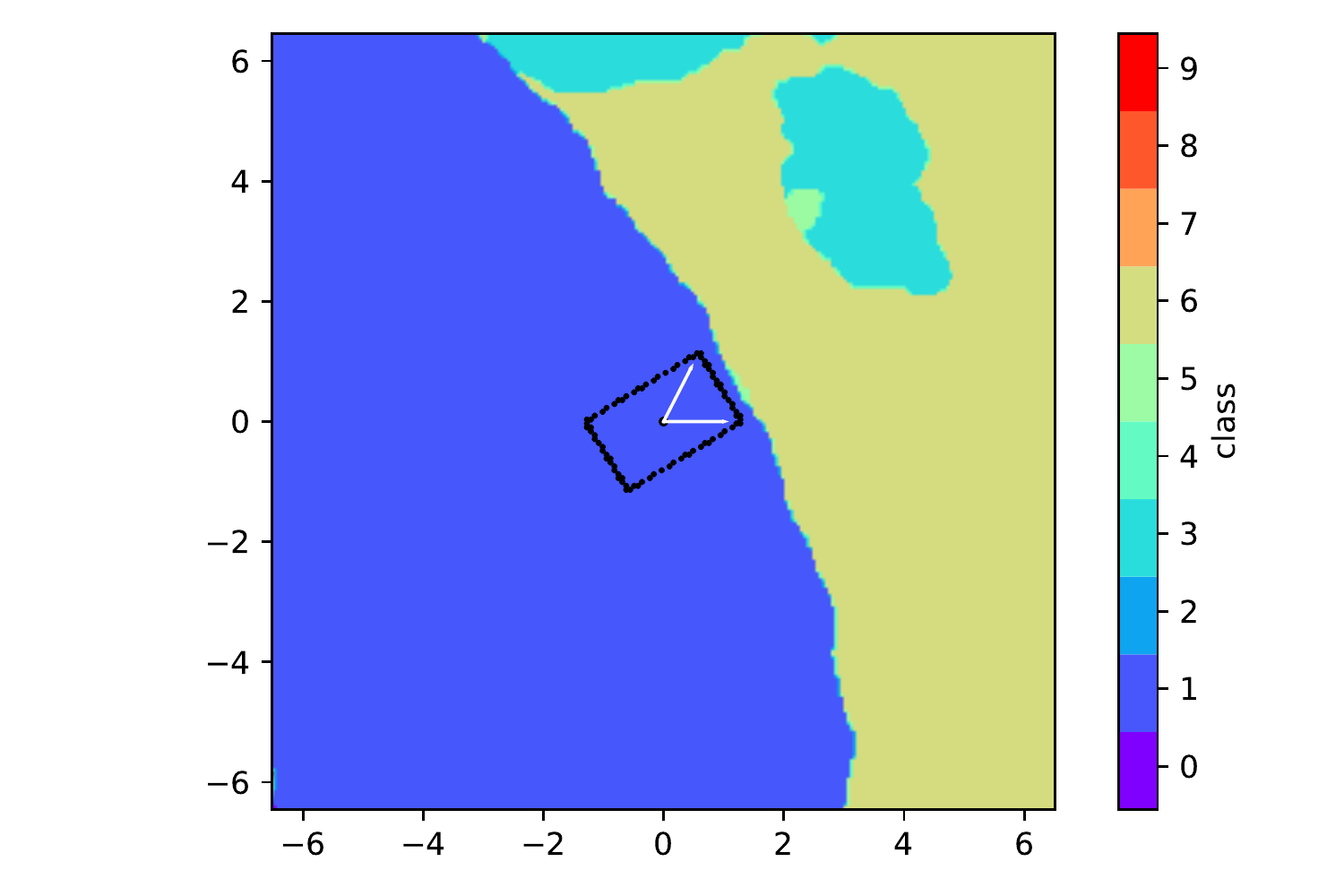}
    \includegraphics[width=0.19\linewidth]{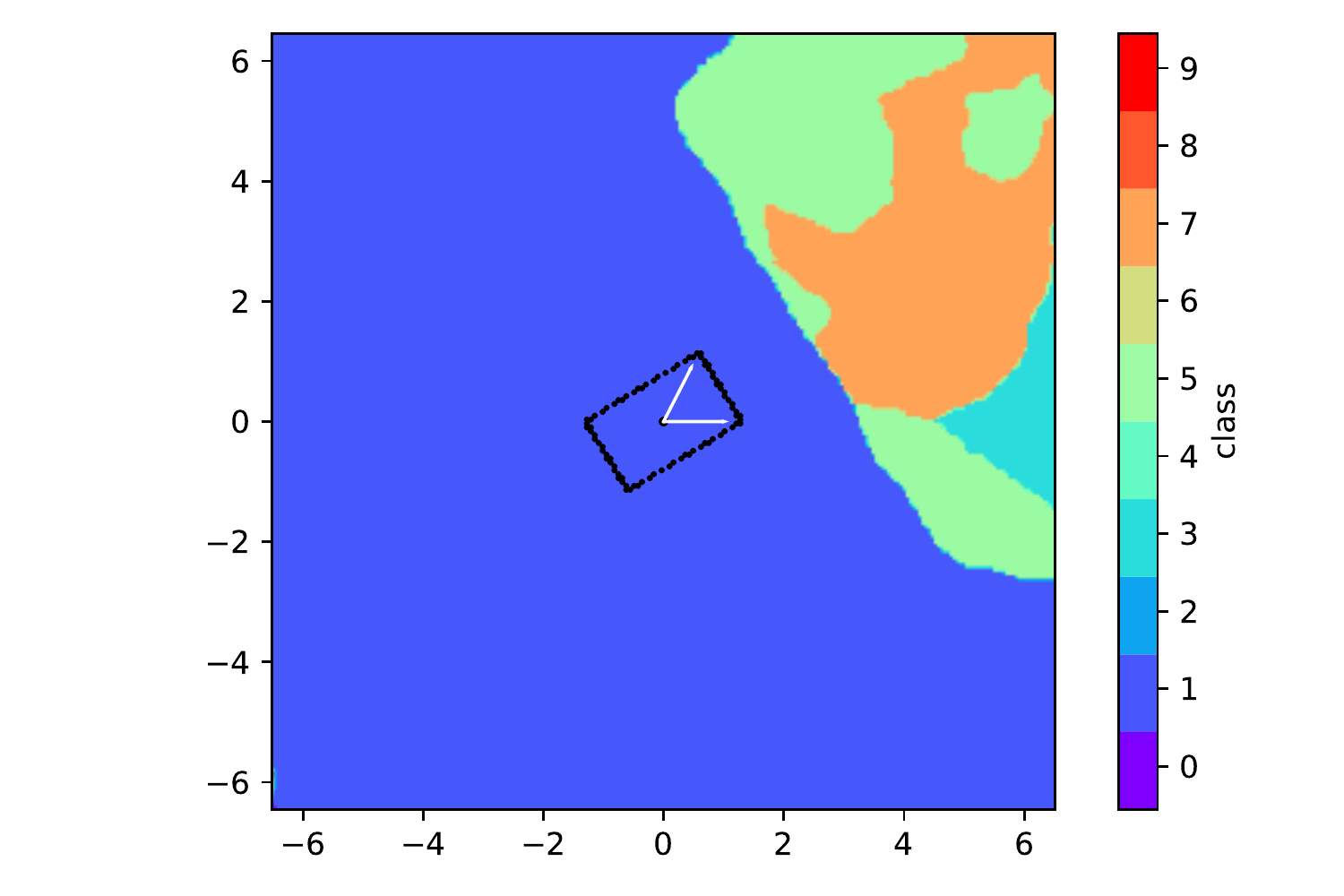}
    \includegraphics[width=0.19\linewidth]{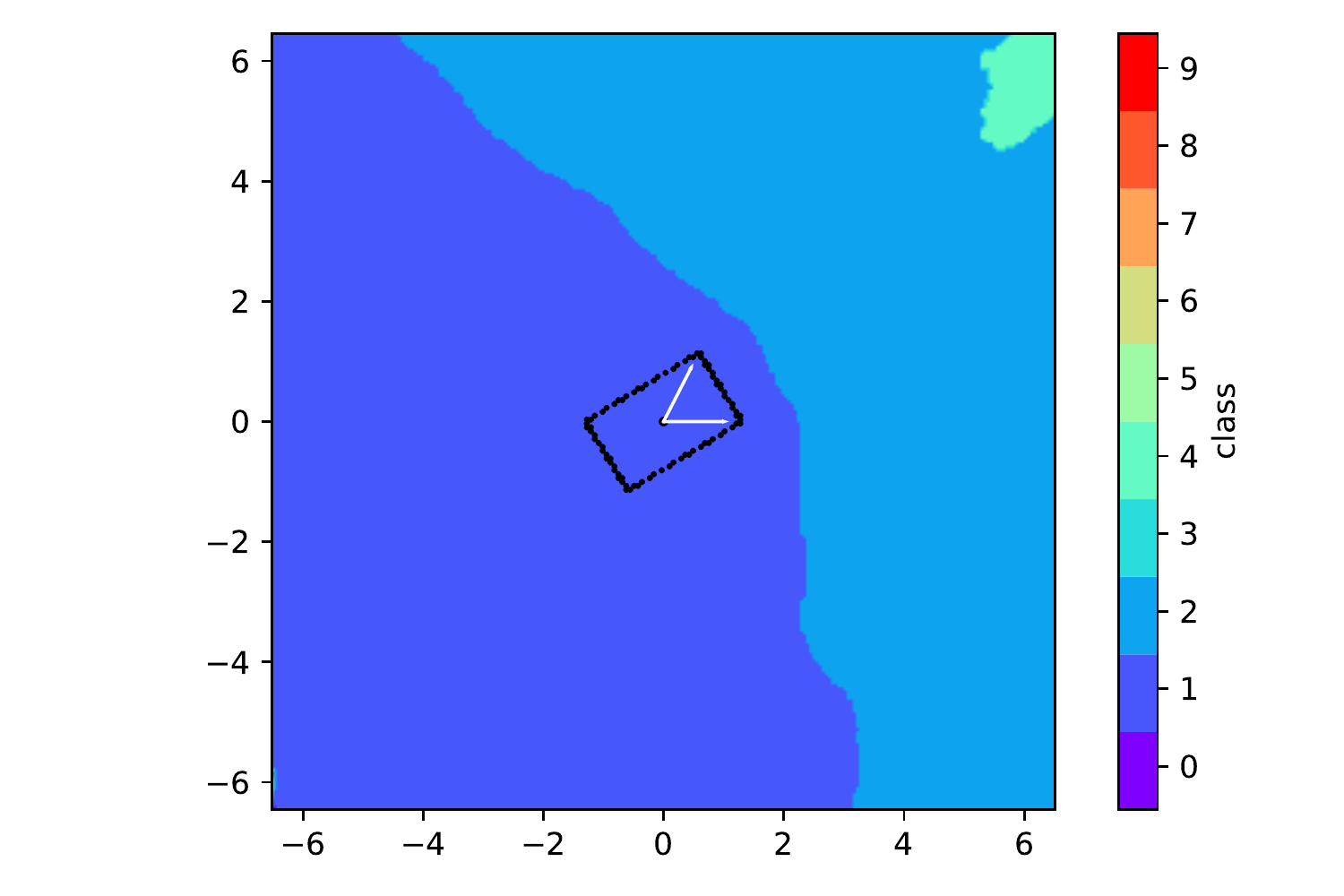}
    \includegraphics[width=0.19\linewidth]{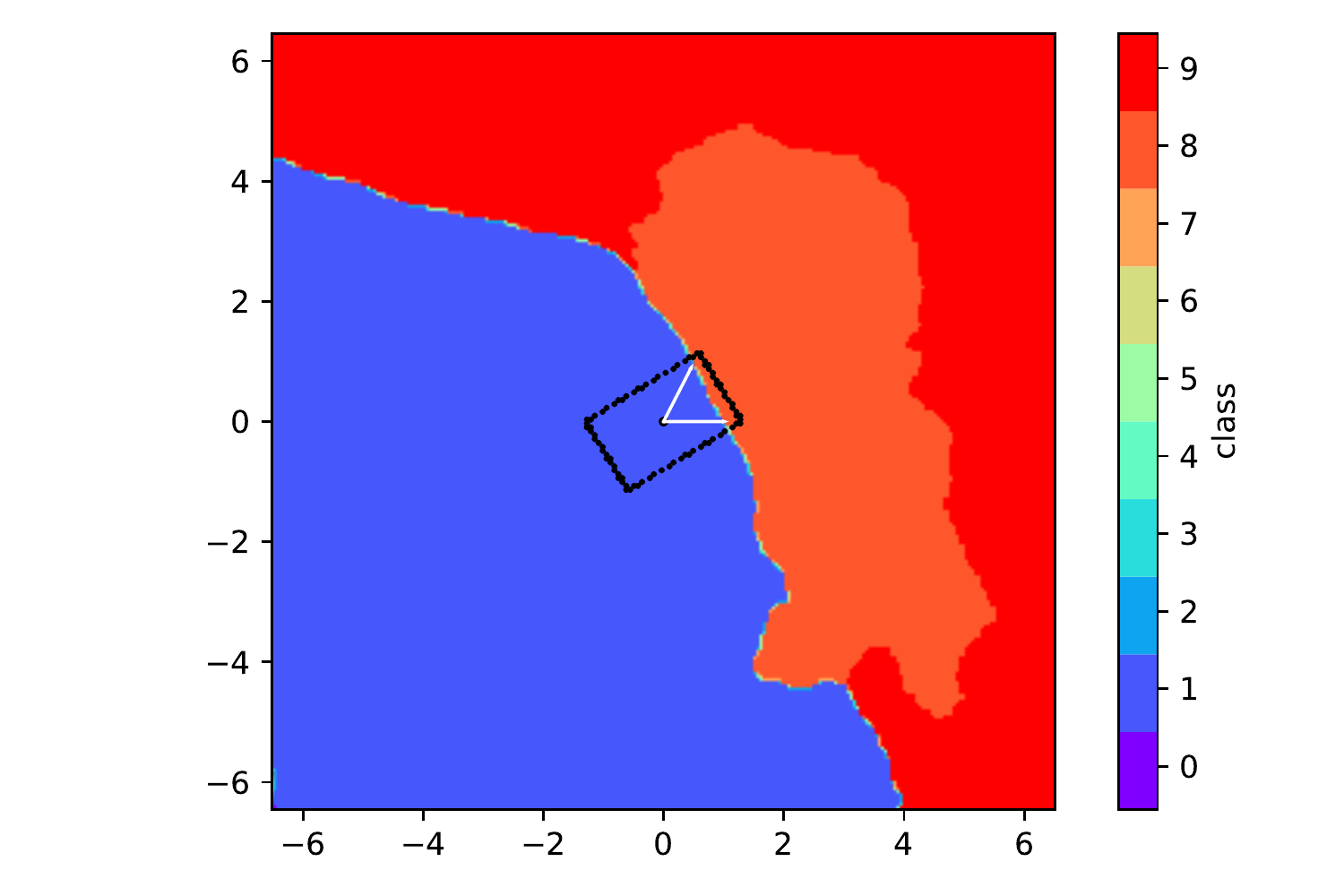}
    \includegraphics[width=0.19\linewidth]{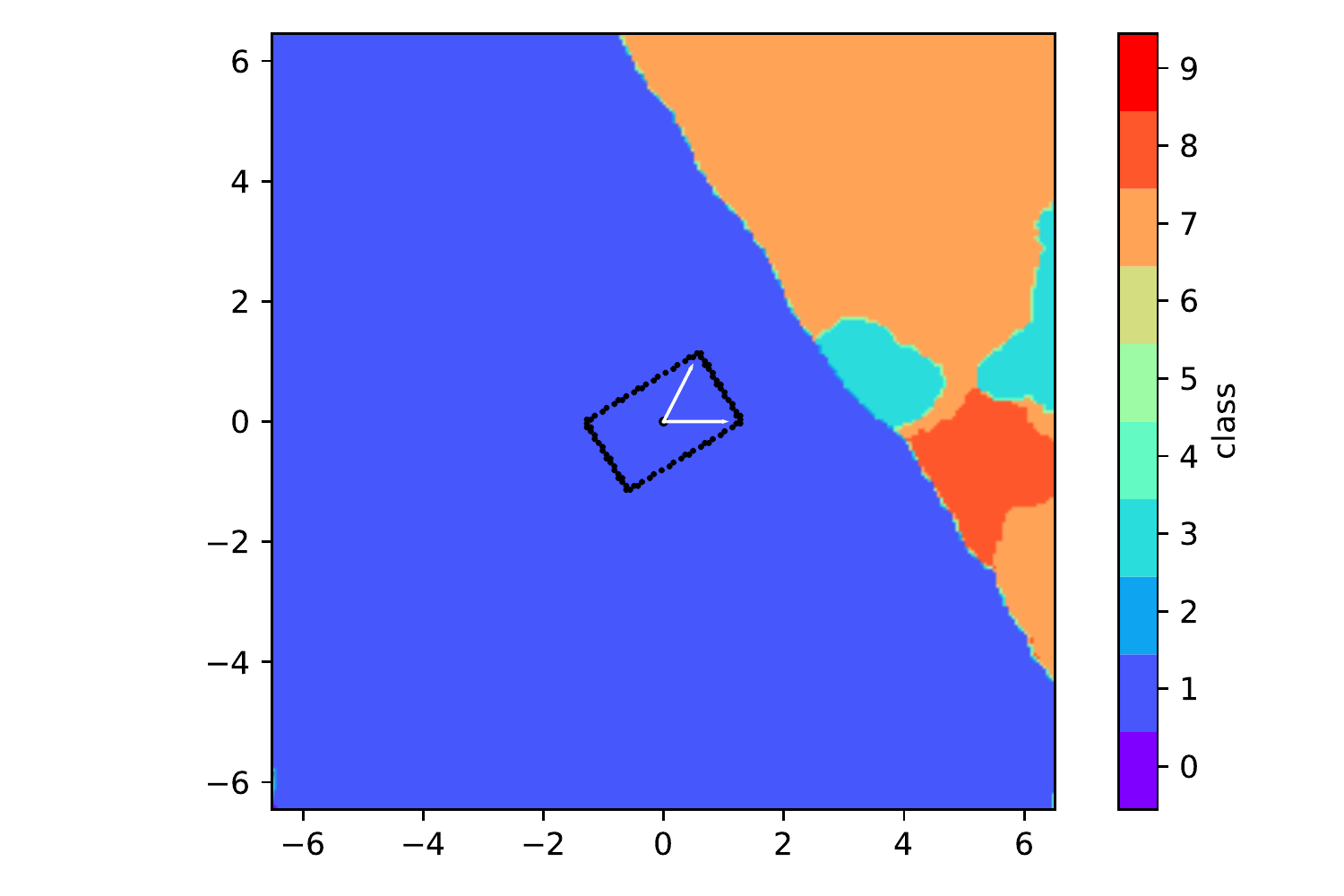}
    \includegraphics[width=0.19\linewidth]{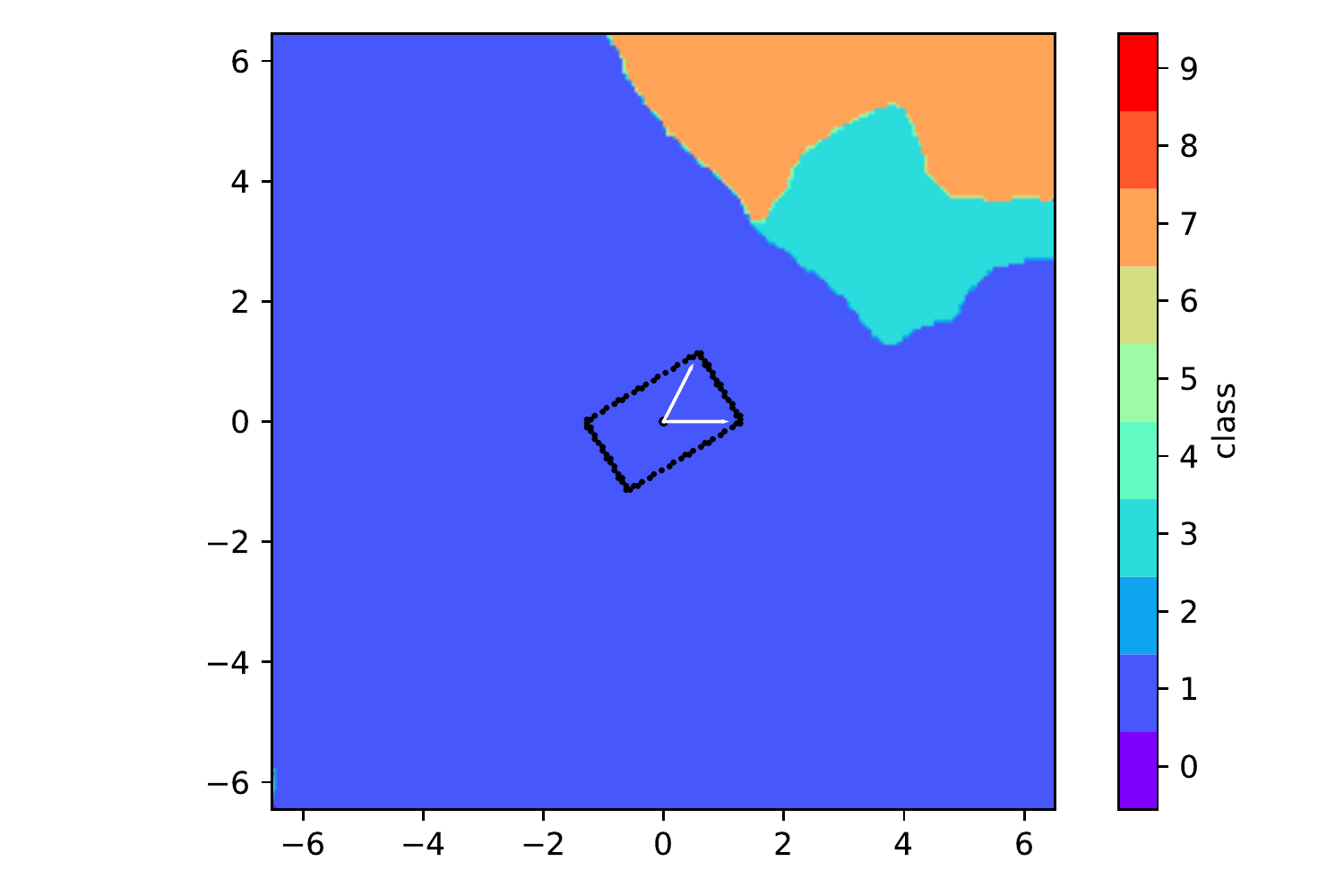}
    \includegraphics[width=0.19\linewidth]{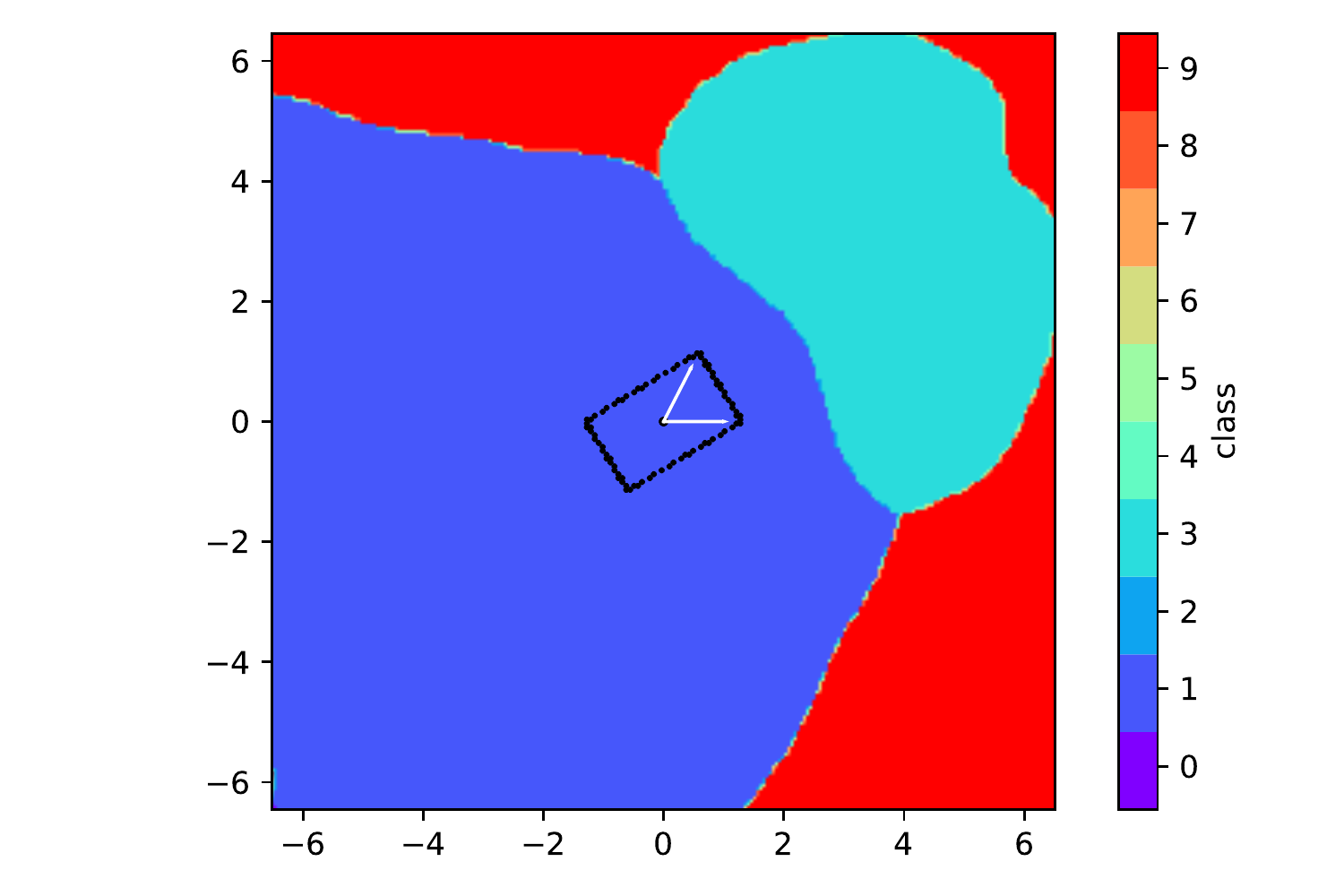}
    \vspace{.5cm}\\
    \includegraphics[width=0.19\linewidth]{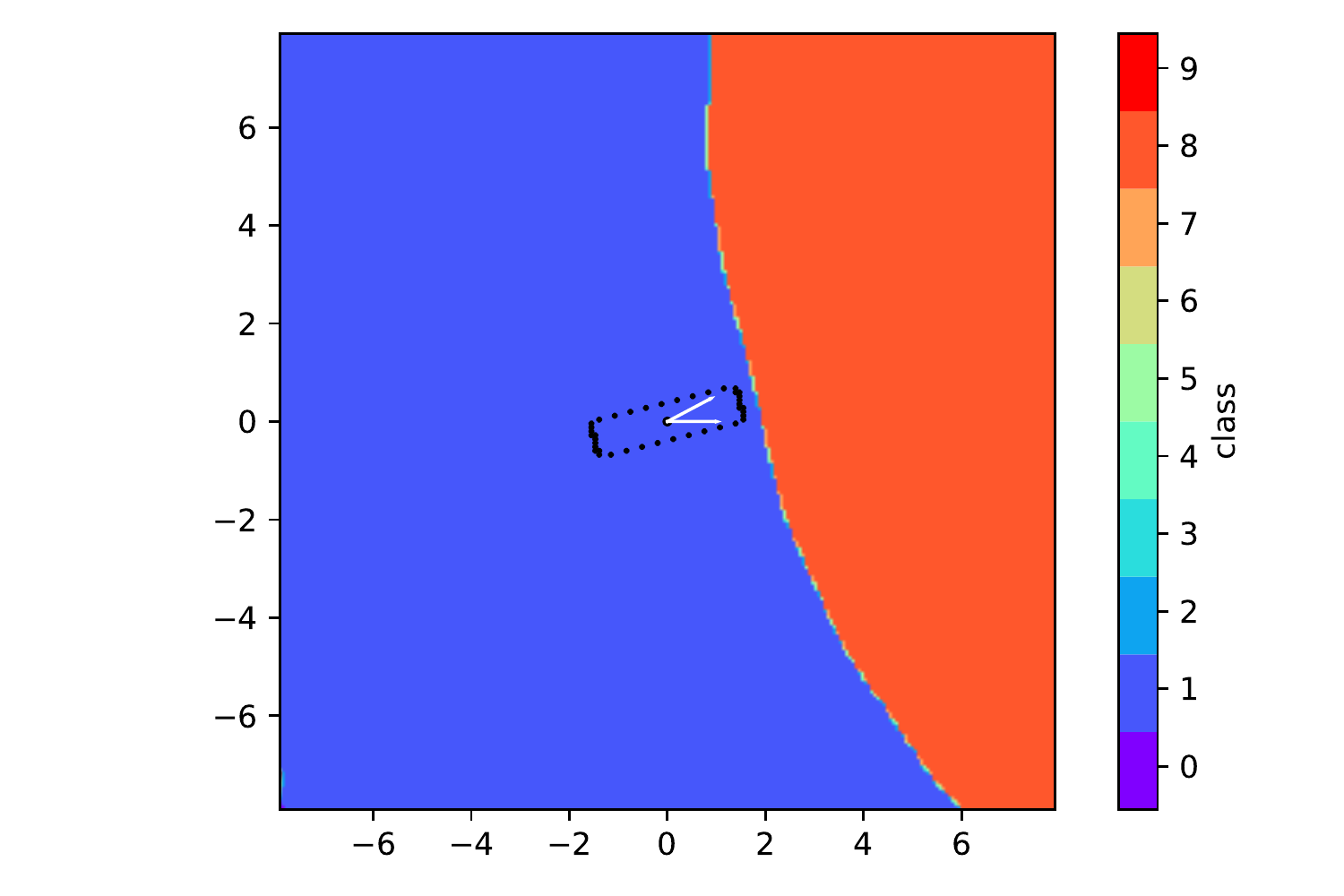}
    \includegraphics[width=0.19\linewidth]{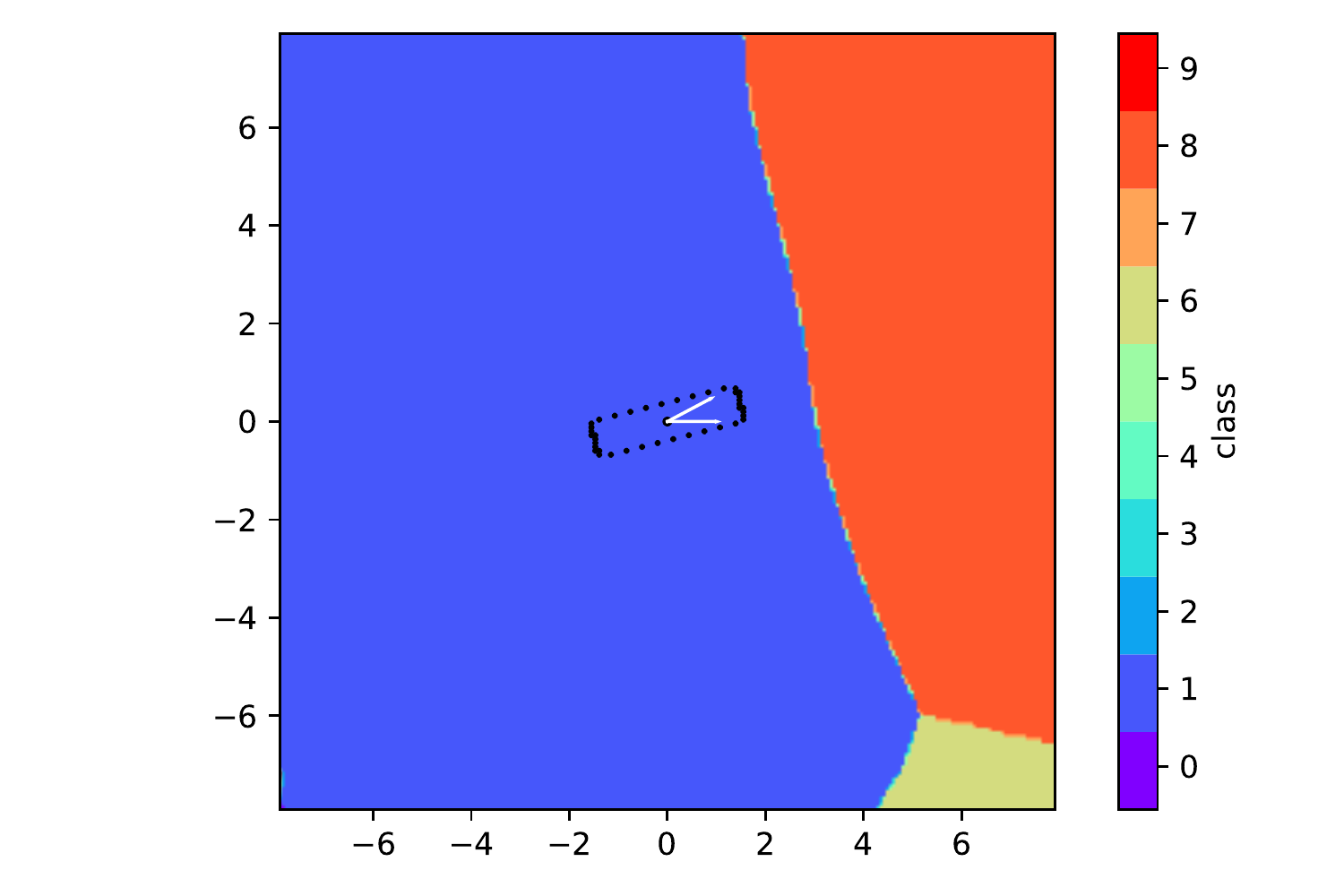}
    \includegraphics[width=0.19\linewidth]{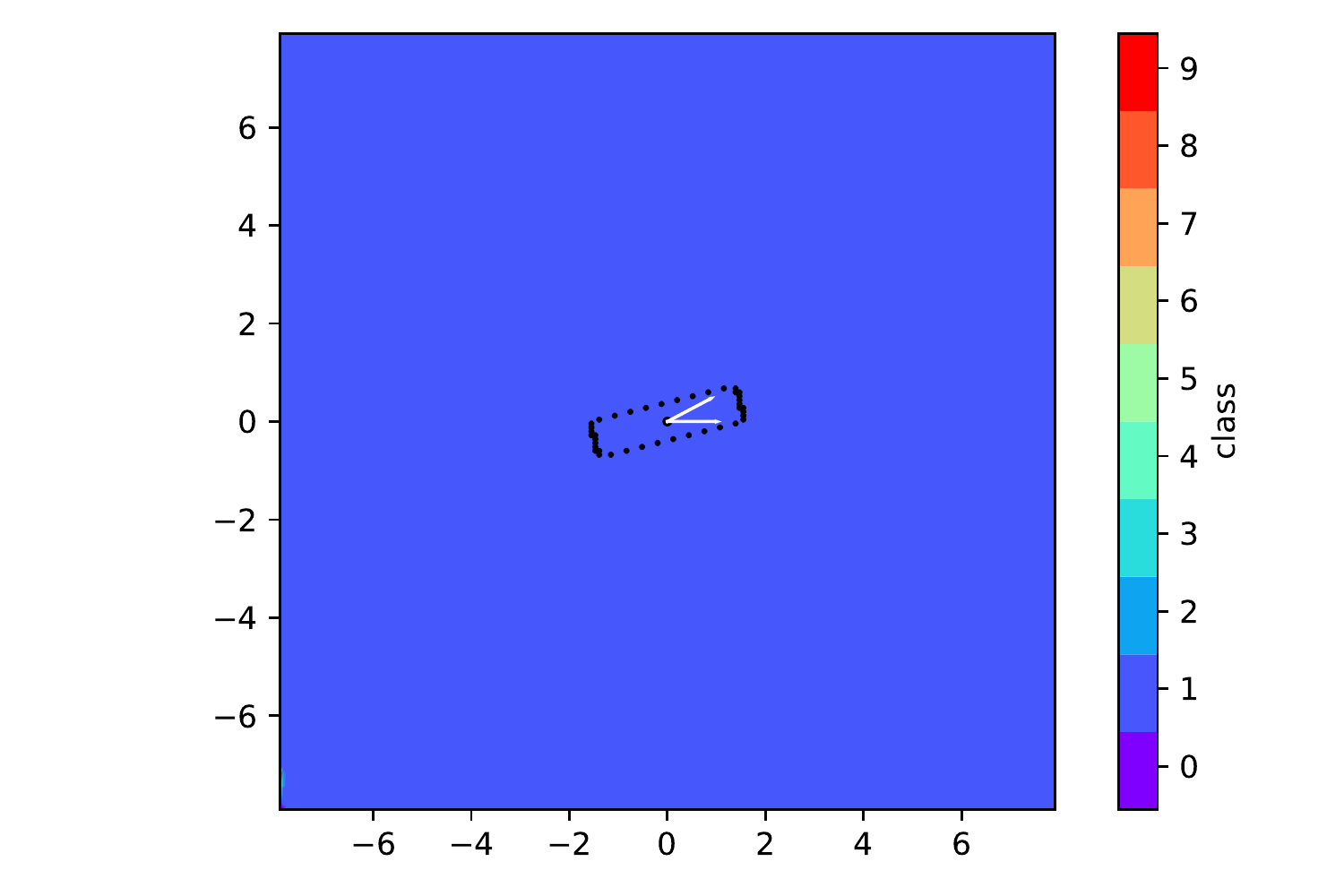}
    \includegraphics[width=0.19\linewidth]{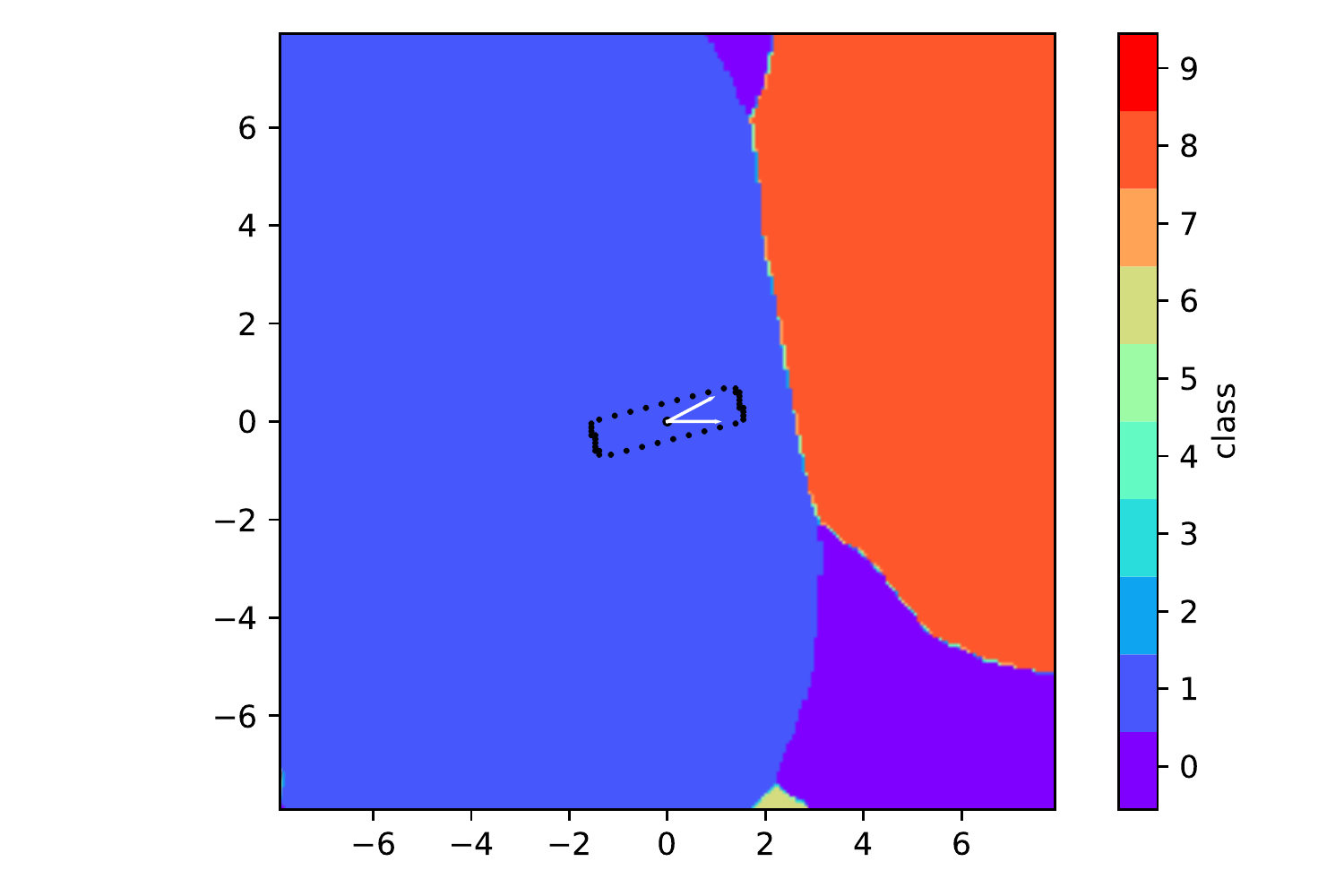}
    \includegraphics[width=0.19\linewidth]{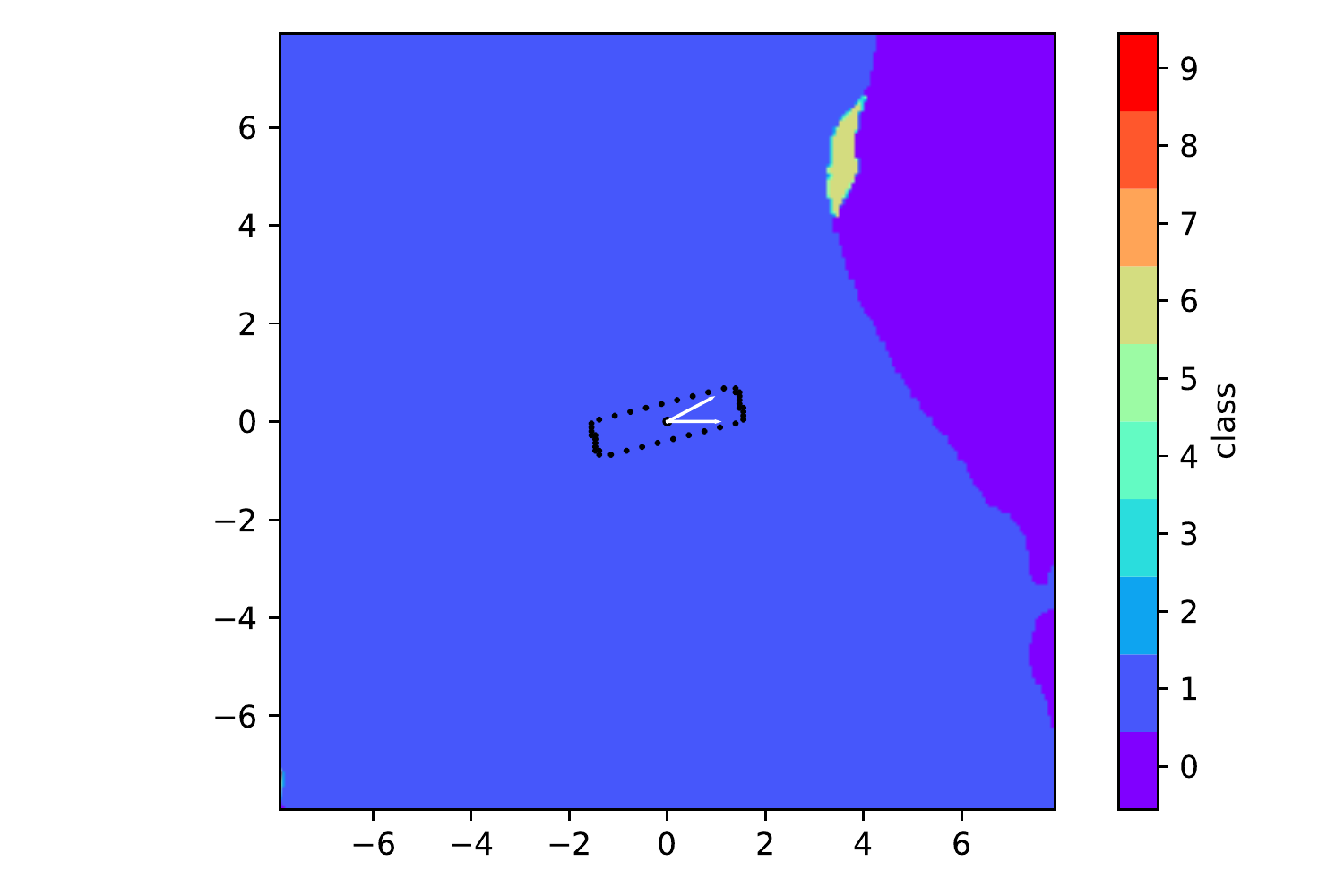}
    \includegraphics[width=0.19\linewidth]{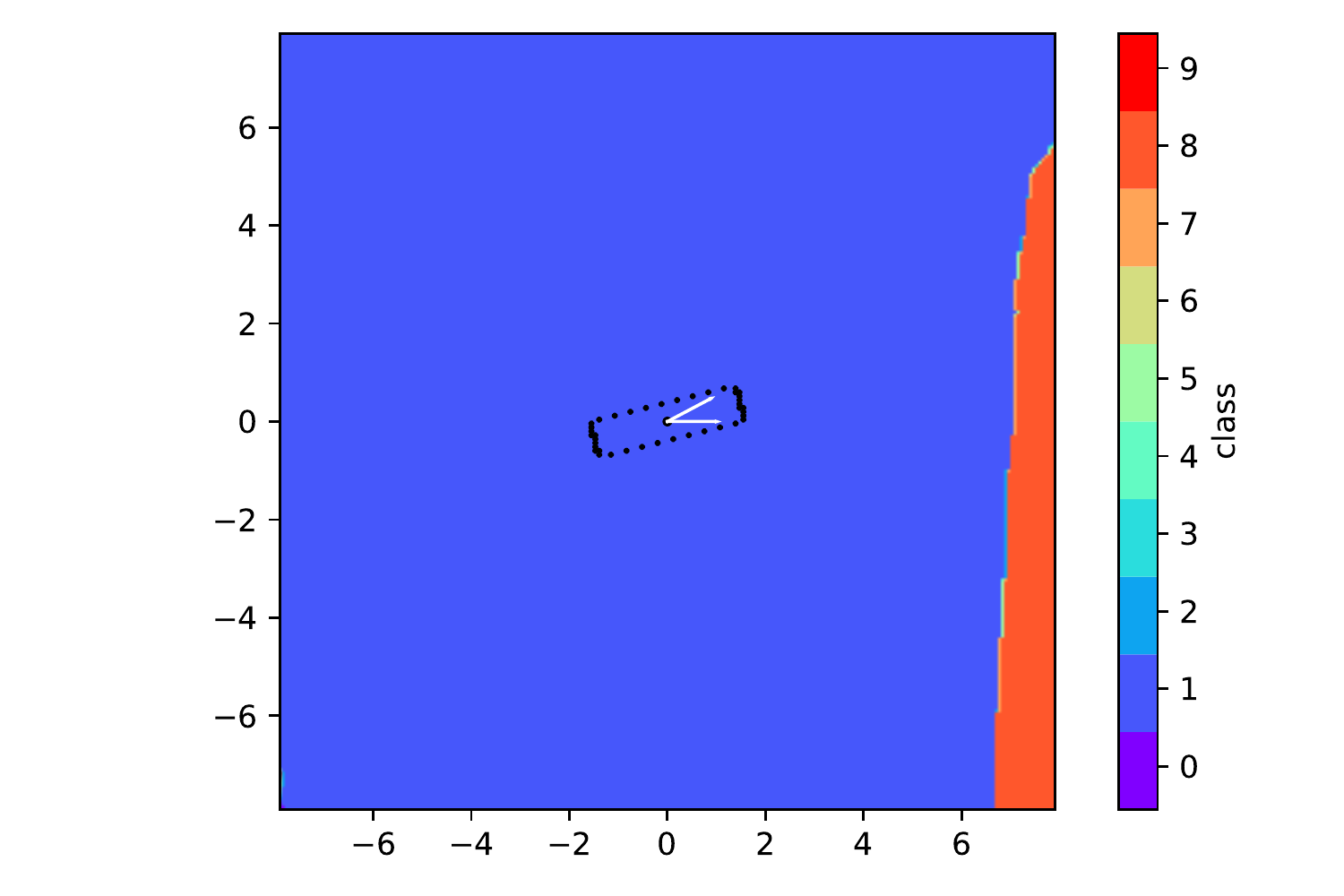}
    \includegraphics[width=0.19\linewidth]{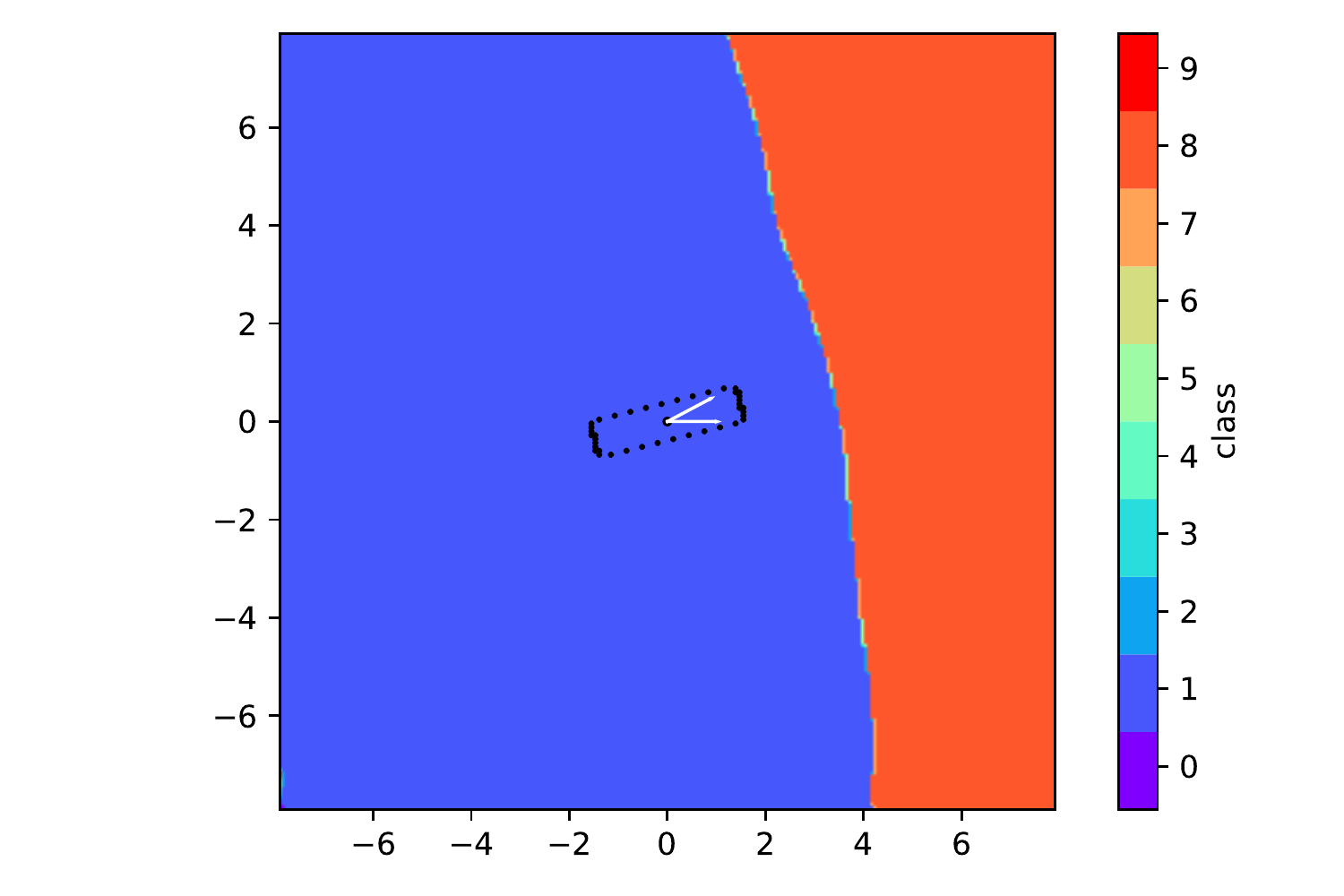}
    \includegraphics[width=0.19\linewidth]{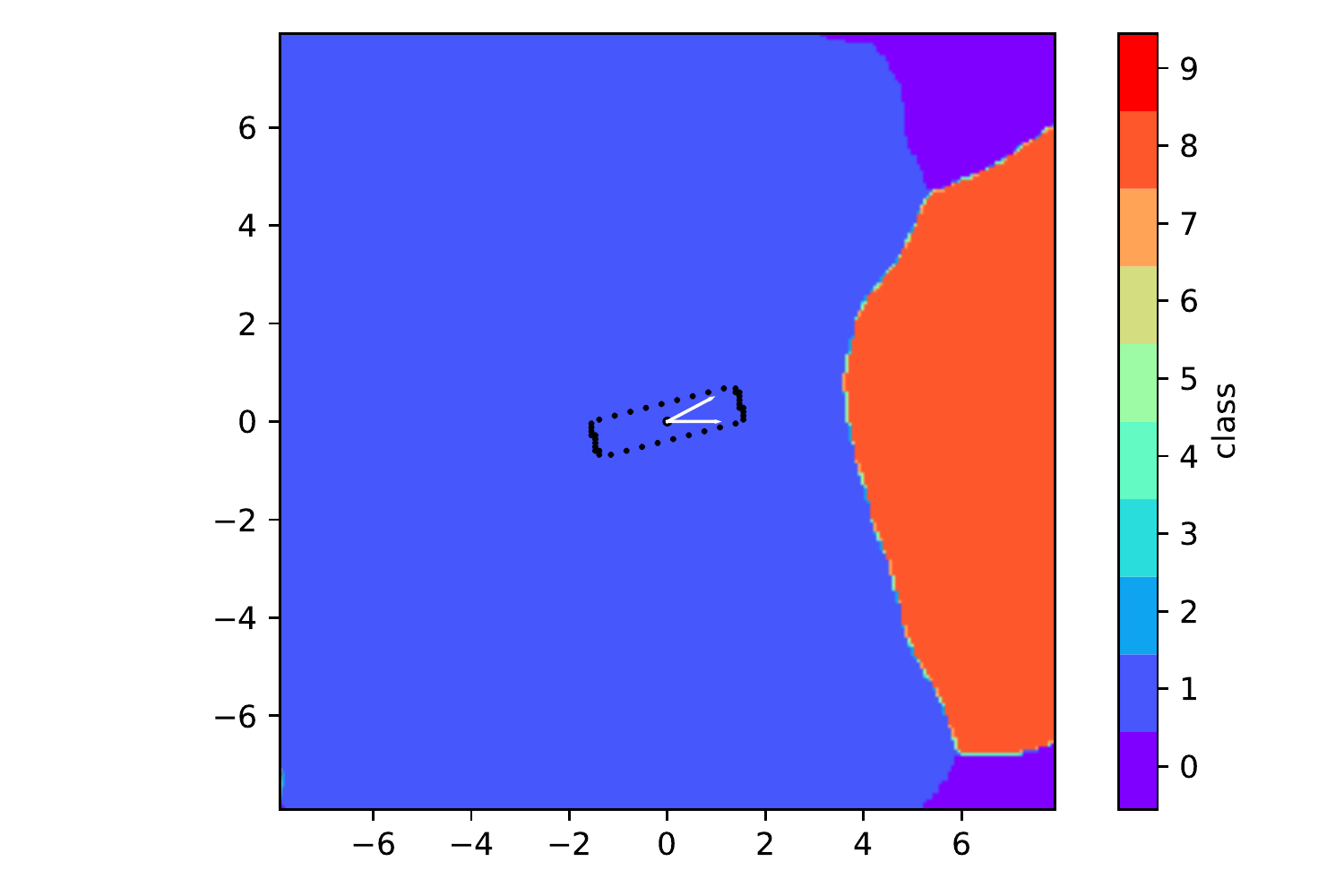}
    \includegraphics[width=0.19\linewidth]{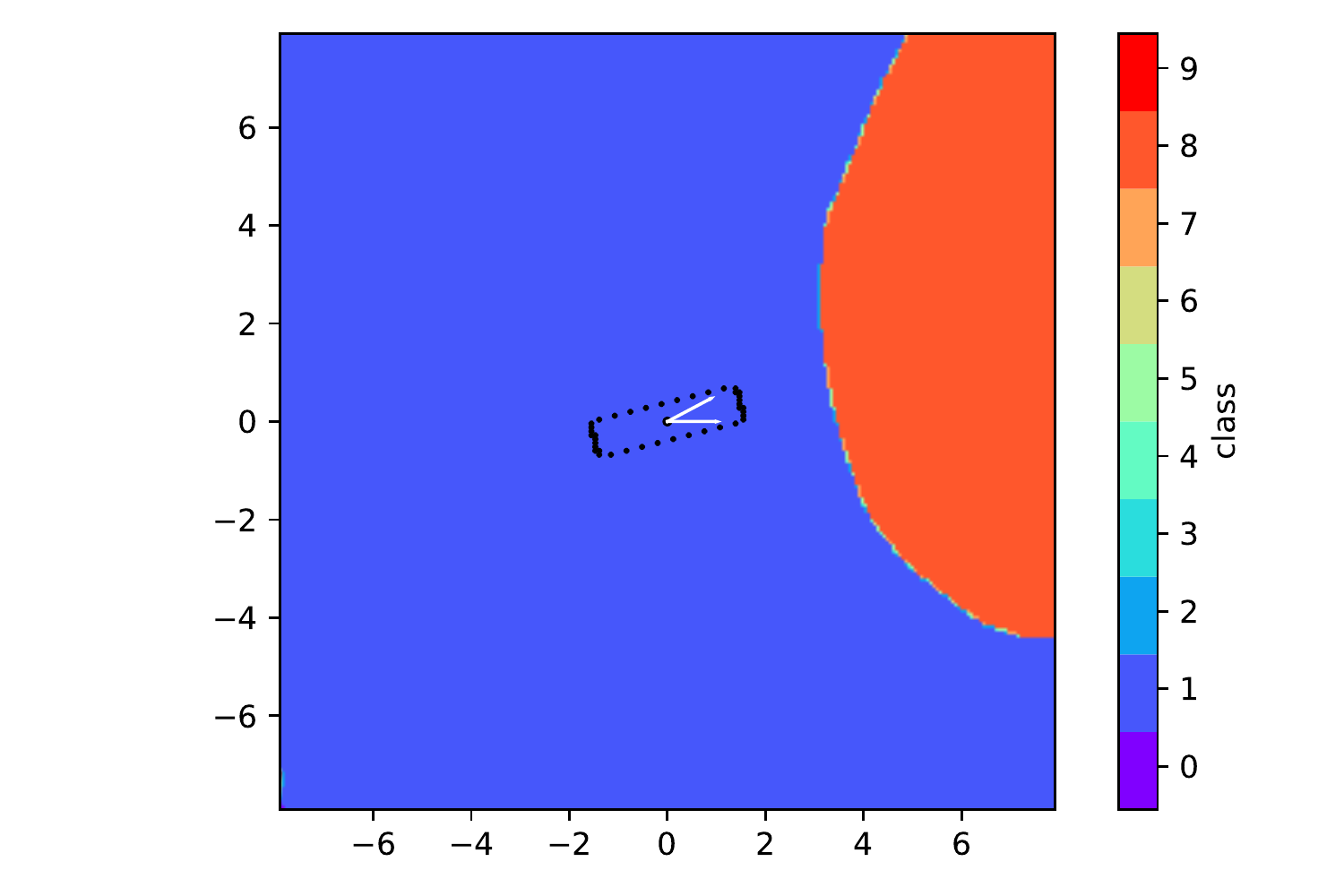}
    \includegraphics[width=0.19\linewidth]{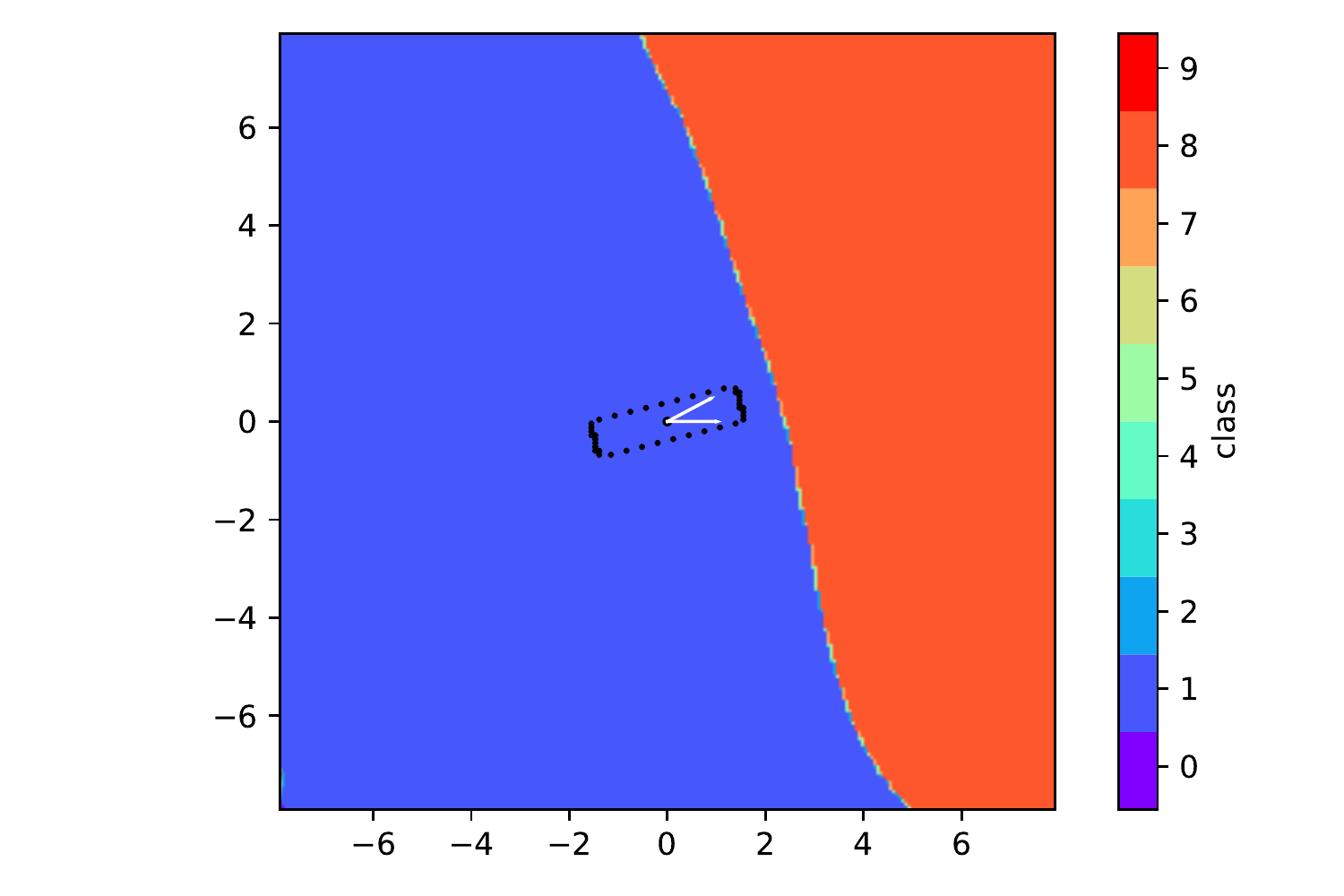}
    \caption{
    \textbf{Decision boundary of sample models projected on the 2D plane spanned by two EOT-PGD directions.}
    The randomized neural networks are trained with GradDiv (top two rows) and without GradDiv(bottom two rows), respectively.
    When trained with GradDiv, the sample models has more diverse decision boundaries.
    We indicate the two EOT directions with white arrows and the $\ell_\infty$-bounded box with a parallelogram.
    }
    \label{suppfig:vis}
\end{figure}

Figure \ref{suppfig:vis} shows that GradDiv diversifies not only the gradient distribution, but also the decision boundary of the sample models.
For the model trained with GradDiv, we used GradDiv-DPP with $\lambda=1$ on the CIFAR-10 dataset.

\subsection*{Sensitivity Analysis on $\lambda$}
\begin{figure}[]
    \centering
    \includegraphics[width=0.5\linewidth]{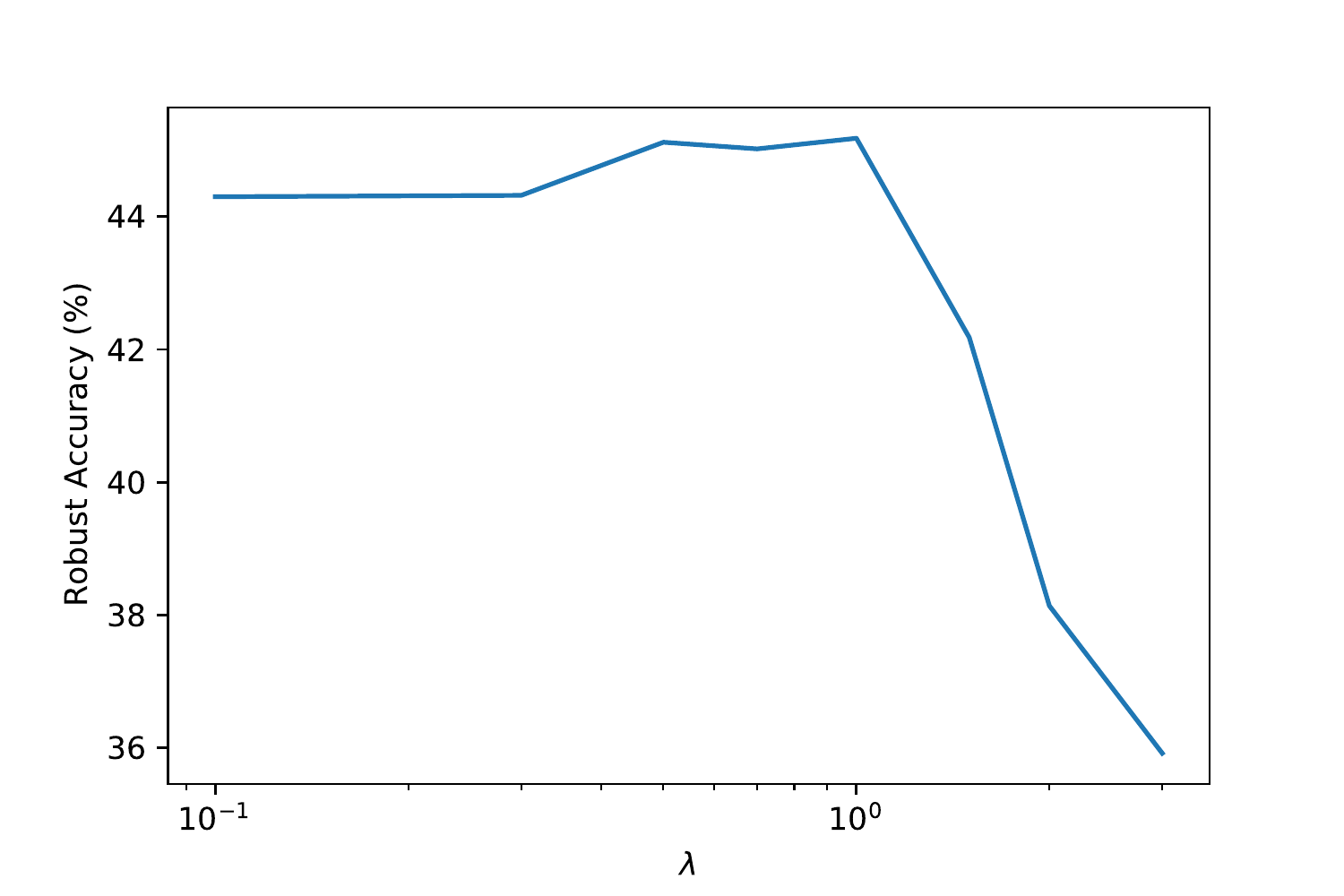}
    \caption{\textbf{Sensitivity of the robustness of GradDiv to the regularization weight $\lambda$.}
    }
    \label{suppfig:sensitivity}
\end{figure}

Figure \ref{suppfig:sensitivity} shows the sensitivity of the robustness of GradDiv to the regularization weight $\lambda$.
We used GradDiv-DPP on the CIFAR-10 dataset and evaluated with the EOT-PGD attack.

\subsection*{Sensitivity Analysis on Attack Iteration $m$}
\begin{figure}[]
    \centering
    \includegraphics[width=0.5\linewidth]{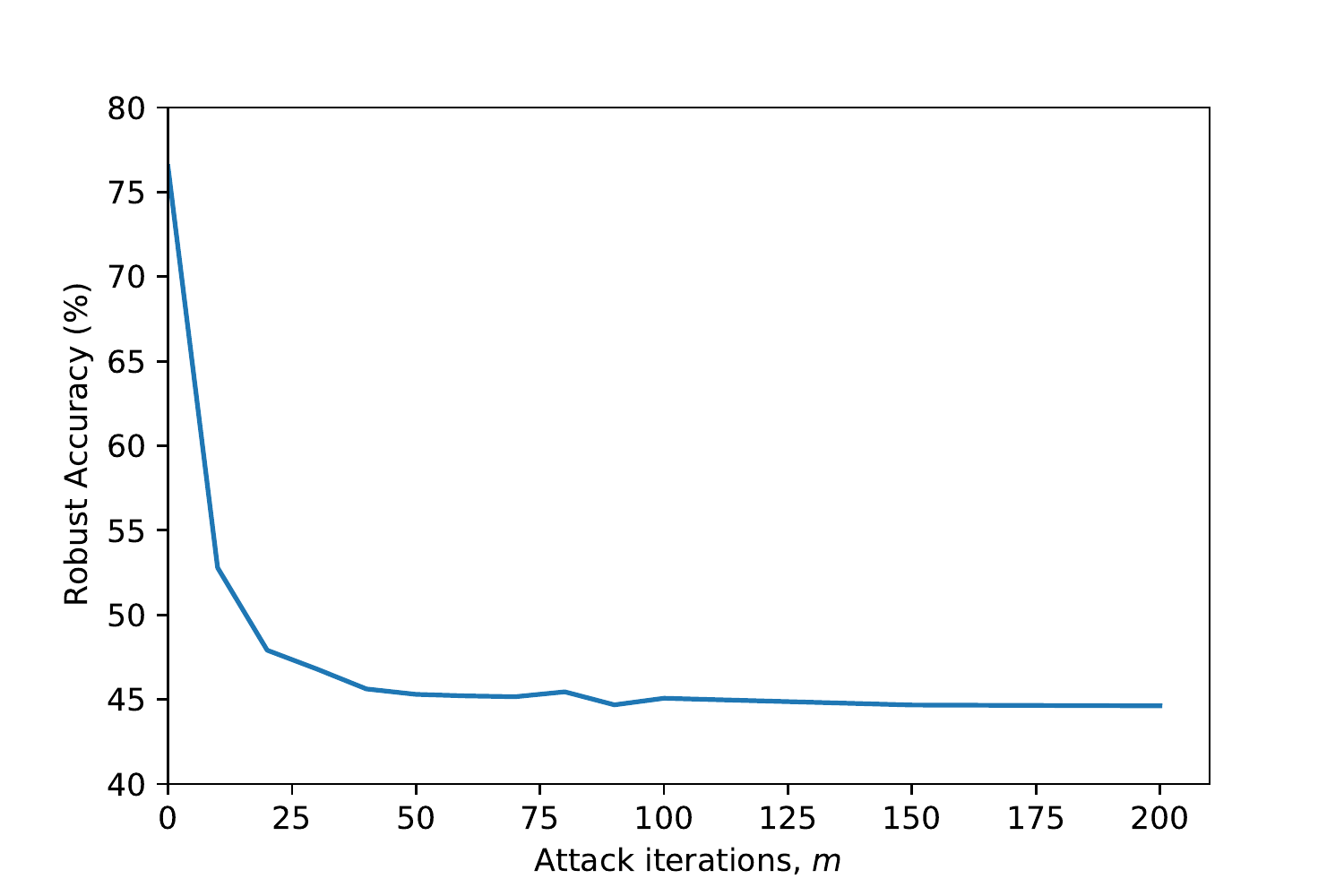}
    \caption{\textbf{Sensitivity of the robustness of GradDiv to the attack iteration $m$.}
    }
    \label{suppfig:attack_iters}
\end{figure}

Figure \ref{suppfig:attack_iters} shows the sensitivity of the robustness of GradDiv to the attack iterations $m\in[0,200]$.
The attack iteration $m=50$ used in Table \ref{table:acc_cifar10} is enough to evaluate the robustness.
We note the minimum robust accuracy is 44.62\% which is only 0.33\%p lower than the result with $m=50$.
We used GradDiv-DPP on the CIFAR-10 dataset and evaluated with the EOT-PGD attack.

